\documentclass{article}
\usepackage{iclr2027_conference,times}
\iclrfinalcopy  % uncomment for camera-ready
\usepackage{amsmath,amssymb,amsthm,mathtools,bm}
\usepackage{graphicx}
\usepackage{booktabs}
\usepackage{enumitem}
\usepackage{hyperref}
\usepackage{url}
\usepackage{wrapfig}

\newtheorem{theorem}{Theorem}[section]
\newtheorem{proposition}[theorem]{Proposition}
\newtheorem{corollary}[theorem]{Corollary}
\newtheorem{lemma}[theorem]{Lemma}
\theoremstyle{definition}

\newtheorem{definition}[theorem]{Definition}
\newtheorem{example}[theorem]{Example}
\theoremstyle{remark}
\newtheorem{remark}[theorem]{Remark}

\newcommand{\R}{\mathbb{R}}
\newcommand{\C}{\mathbb{C}}
\newcommand{\Z}{\mathbb{Z}}
\newcommand{\Ghat}{\widehat{G}}
\newcommand{\Chat}{\widehat{C}}
\newcommand{\Mstar}{M^{\star}}
\newcommand{\Wbar}{\overline{W}}
\newcommand{\tr}{\operatorname{tr}}

\newcommand{\id}{\mathrm{I}}
\newcommand{\inner}[2]{\langle #1, #2\rangle}
\DeclarePairedDelimiter\norm{\lVert}{\rVert}

\title{Group-Invariant Statistics Determine Embedding Geometry:\\
Harmonic Analysis of Representations from Bach to the Night Sky}
\author{Liam Storan\thanks{Correspondence: lstoran@stanford.edu} \\
Stanford University
\And
Andreas Tolias \\
Stanford University
\And
Nina Miolane \\
University of California, Santa Barbara
}

\begin{document}
\maketitle
\lhead{Preprint. Under review.}
\begin{abstract}
The representations that language models learn for concepts such as
months, weekdays, and places display consistent geometric structure:
circles and saddle-shaped ``Pringle''
manifolds. Recent work traced these structures to \emph{translation
symmetry} in word co-occurrence statistics, deriving the observed
Fourier geometry when co-occurrence depends only on distance on an
abelian lattice of concepts. We demonstrate that more general notions of symmetry lead to equally structured predictions. Considering symmetries defined by arbitrary finite groups, compact groups, and homogeneous
spaces, we prove that whenever the co-occurrence statistics of a word family
are invariant under a group $G$, the learned word embeddings consist of matrix elements of the irreducible representations (irreps) of $G$. Circles and Pringles arise when $G$ is cyclic, in which case
the irreps are Fourier modes. We verify the irrep structure in three experimental settings.
(i)~The cyclic group $\mathbb{Z}_{12}$: for the months of the year we
recover the known circular geometry. (ii)~A dihedral
group acting on the major and minor triads: we unify two
classical observations---that transposition and chord inversion form a group ($T/I$)
acting on chords (music theory), which \textit{implies} that the well-known ``circle of fifths'' emerges
in learned chord embeddings (machine learning). 
%Namely, chord co-occurrence in Bach's
%chorales is largely invariant under $T/I$ which \emph{forces} the observed geometry. 
(iii)~We explain and reproduce a recently discovered spherical representation of celestial objects in large language models (LLMs) as a spherical-harmonic embedding derived from our theory. 
%The same structure is present in text embeddings from a corpus of star guides. 
Our results demonstrate that the
geometry of learned representations is often a consequence of the statistical symmetry of underlying data. 
\end{abstract}

\section{Introduction}
\label{sec:intro}

Large language models (LLMs) organize concepts geometrically in their activations. Months of the year lie on circles \citep{engels2024not}, latitude and longitude of cities can be linearly decoded from LLM activations \citep{gurnee2024language}, and celestial objects reside on a sphere inside a LLM's residual stream \citep{berdnikov2026sky}. A compelling explanation for the ubiquitous circular geometry was given by \citet{karkada2026symmetry}: when the co-occurrence
of words in text is only a function of their distance on some ``cyclic latent lattice'', the
optimal embedding under a quadratic word-embedding model
\citep{karkada2025quadratic} consists of Fourier modes which in turn generates circles. For example, they found that the measured co-occurrence of January and February is about the same as November and December in general text, which explains why language models embed the 12 months on a circle.

% so periodic
% statistics \emph{force} circles, with radii given by the Fourier
% coefficients of the statistics.

\begin{figure}[t]
\centering
\includegraphics[width=\textwidth]{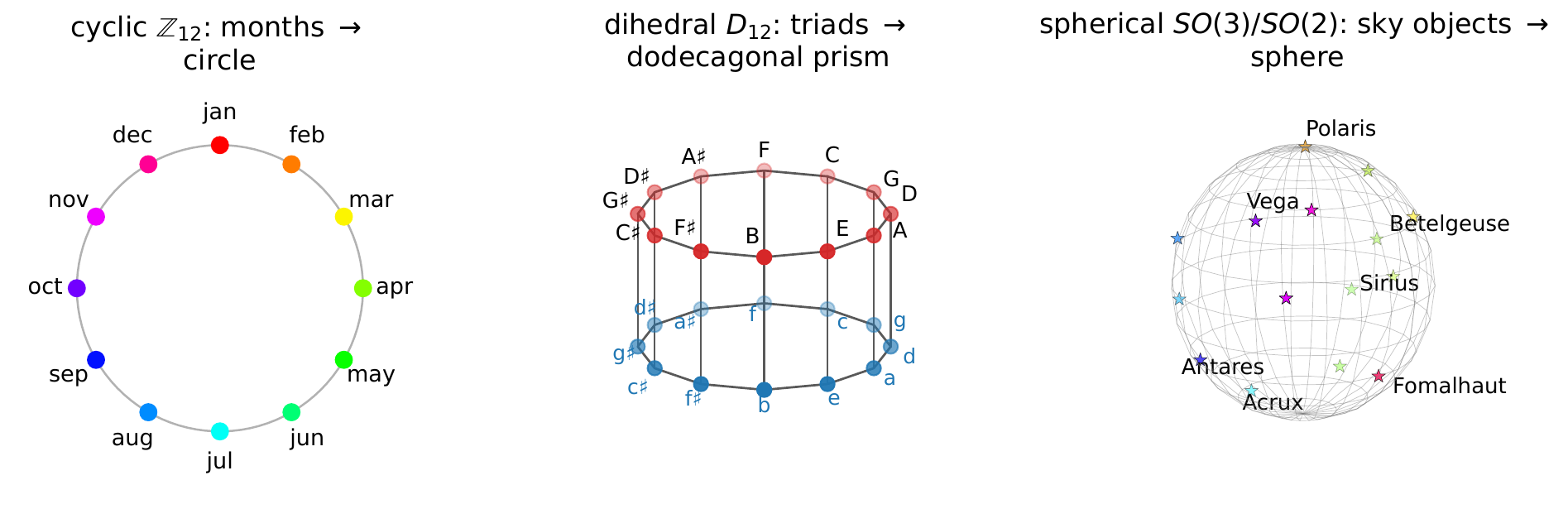}
\caption{\textbf{Months, musical chords, and celestial objects exhibit distinct latent geometries}, dictated by the symmetry of their word co-occurrence statistics in text corpora.
% \emph{Left}: months lie on a latent circle. \emph{Middle}: musical chords in Bach lie on a latent dodecagonal prism. \emph{Right}: celestial objects lie on a latent sphere.} 
\emph{Left}: month statistics are approximately invariant under cyclic shift
($\Z_{12}$), which places them on a circle. \emph{Middle}: musical chord statistics are approximately invariant under
transposition and inversion ($D_{12}$), which places them on a dodecagonal prism. \emph{Right}: celestial object statistics
are approximately invariant under 3D rotations ($SO(3)$), which 
%($S^2$), placing celestial objects 
places them on a sphere.} 
\label{fig:geometries}
\end{figure}

\paragraph{Contributions.}
Cyclic symmetry is just one example of structure that can exist in data (Figure~\ref{fig:geometries}). Here, we ask and answer the general question:
\emph{what does an arbitrary symmetry of co-occurrence statistics imply
for embedding geometry?} The answer is a complete picture of the components used to construct word embedding geometry, assembled from group harmonic analysis. Specifically, our contributions are:
\begin{itemize}[leftmargin=*]
\item \textbf{A general theory of word embedding geometry under group symmetry}: if the words form a torsor for a group
$G$ (\textit{i.e.}, if we can form a bijection between the words and group elements, Theorem~\ref{thm:main-finite}) or a homogeneous space $G/K$ (Theorem~\ref{thm:main-GK}), the embedding's entries are matrix elements of the irreps of $G$ that can be expressed analytically.
\item \textbf{A non-Abelian validation of the theory} \textit{that unifies existing, but disconnected, observations from music theory}: 
we use our theory to reconcile two classical observations about music:
the existence of transposition symmetry ($T/I$) in music theory~\citep{crans2009musical}, and 
the so-called circle of fifths that emerges in learned chord embeddings~\citep{lahnala2021chord}. 
%Our theory supplies the missing bridge. 
Specifically, we show that
Bach's chorales have approximately $T/I$-invariant chord statistics
(invariance captures $R^2 = 0.822$ of the co-occurrence structure), and
that this statistical symmetry \emph{forces} the observed geometry. The empirical circle of fifths matches our theoretical predictions at a per-chord cosine of $0.98$.  We observe the same structure in ChordBERT, a masked language model for symbolic chord sequences.

% Embeddings fit on the chorales realize the predicted
% two-dimensional irrep planes, \emph{led by the circle of fifths}, whose
% full $24$-point configuration matches the prediction computed from the
% symmetrized statistics at per-chord cosine $0.96$
% (Section~\ref{sec:music}). The circle of fifths, on this account, is
% not an emergent curiosity but the leading irrep of a symmetry the data demonstrably carries.

\item \textbf{A continuous validation of the theory} \textit{that challenges existing results on LLM activations}: We reanalyze the sky-sphere
LLM activations of \citet{berdnikov2026sky} and show that the representation of celestial objects is
not a sphere, as the authors initially reported, but is actually \emph{spherical-harmonics}.
Specifically the eigenmodes of the LLM activations are degree-ordered spherical-harmonic blocks ($\ell{=}1$, $\ell{=}2$,
$\ell{=}3$), in five of the seven models examined. We reproduce these harmonic results using four public-domain
astronomy books ($131$k tokens, selected by a ``star guide'' criterion) and examining classical word embeddings. We show that co-occurrence statistics place the objects on the true
celestial sphere at a statistically significant per-object cosine $0.55$ (median $0.68$), which is within the range of accuracy that an idealized spherical model predicts for a sample of
this size. The same symmetry fingerprint, present in a $131$k-token
field guide and in a $123$B-parameter LLM's activations, sharpens in
the model, as expected if training supplies vastly more such text.
\end{itemize}

Together, these results show that word embeddings admit analytical expressions determined by the symmetry of their co-occurrence statistics in quadratic embedding models, and suggest that LLM activations might be described by the same rule.

\section{Related work}
\label{sec:related}

We review work linking word statistics to their embeddings, focusing on symmetry-driven statistics, and survey the role of group theory in interpretability research.

\textbf{From word statistics to their embeddings.} Early work has linked the structure of word embeddings to their co-occurrence statistics, but did not connect that structure to any symmetry of these statistics. Skip-gram with
negative sampling implicitly factorizes a shifted Pointwise Mutual Information (PMI) matrix
\citep{levy2014neural}. GloVe (Global Vectors for Word Representation) fits log co-occurrence directly
\citep{pennington2014glove}. \citet{karkada2025quadratic} derive a
quadratic word-embedding model in which trained embeddings realize a
low-rank factorization of the normalized statistics $\Mstar$
(Eq.~\ref{eq:mstar}), the premise we adopt (Assumption~\ref{ass:premise}).

\textbf{Word statistics described by a symmetric group and their embeddings.}
\citet{karkada2026symmetry} prove that translation-invariant statistics on
$\Z_L^D$ yield sinusoidal embeddings and validate this for periodic
concepts; our Appendix~\ref{app:special_finite} recovers their Proposition~1 as
the cyclic case. \citet{engels2024not} found the day/month circles
empirically. \citet{gurnee2024language} linearly decoded measurements of space and time from activations.
\citet{berdnikov2026sky} discovered the sky sphere, asking explicitly whether the
statistics-based paradigm explains it, which our
Section~\ref{sec:sky} addresses. The linear representation hypothesis and
the geometry of categorical concepts \citep{park2024linear,
park2024geometry} concern complementary, non-manifold structure.

\textbf{Group theory in interpretability.} Beyond word embeddings, group theory has been successfully used to describe representations emerging across neural network architectures. Networks trained on
group operations learn irrep-based algorithms: Fourier features for
modular addition \citep{nanda2023progress} and general irreducible
representations for finite groups \citep{chughtai2023toy,marchetti2026sequential}. Those results
concern models trained \emph{on a group operation}. By contrast, our work concerns models
trained on \emph{natural corpora whose statistics happen to be
group-invariant}, with the group acting on concepts rather than being the
task. 

% The mathematical toolkit---Peter--Weyl, Fourier analysis on groups
% \citep{diaconis1988group}, spherical harmonics and Funk--Hecke
% \citep{dai2013approximation}, Gelfand pairs
% \citep{ceccherini2008harmonic}---is classical.

% \textbf{Music.} Chord embeddings are an established object:
% chord2vec adapted word2vec to chord sequences
% \citep{madjiheurem2016chord2vec}, and \citet{lahnala2021chord}, training
% word2vec on ${\sim}89$k popular songs, found the circle of fifths emerge
% empirically in the embeddings' principal components. The
% dihedral $T/I$ group is a standard object of mathematical music theory
% \citep{lewin1987generalized,crans2009musical}. 

% Our contribution is to
% show that corpus statistics select one of these groups, that embeddings
% inherit its harmonic analysis---and thereby to give the circle-of-fifths
% findings a theoretical origin.

\section{Background on Group Theory}
\label{sec:background}

We briefly review the concepts from group theory that our results are built from. We refer to \citet{diaconis1988group,folland2016course} for details.

\paragraph{Groups.}
A \emph{group} is a set $G$ equipped with an associative binary operation $(g,h)\mapsto gh$, an identity element $1\in G$ with $1g=g1=g$, and an inverse $g^{-1}\in G$ for each $g\in G$ with $g^{-1}g=gg^{-1}=1$.
A group is \emph{Abelian} if $gh=hg$ for all $g,h\in G$, and non-Abelian otherwise. The cyclic group $\mathbb{Z}_n$ (integers modulo $n$ under addition) is Abelian. The dihedral group $D_n$ of rotations and reflections of a regular $n$-gon is not, since rotating and reflecting do not generally commute. 
$\mathbb{Z}_n$ and $D_n$ are finite, but groups can also be continuous, i.e., the group $SO(3)$ of all possible 3D rotations. We will use $\mathbb{Z}_{12}$, $D_{12}$ and $SO(3)$ to characterize the word embeddings of the months of the year, musical chords, and celestial objects, respectively (Figure~\ref{fig:geometries}).

\paragraph{Group actions.}

A (left) \emph{action} of $G$ on a set $L$ is a map $(g,x)\mapsto g\cdot x$ with $1\cdot x=x$ and $g\cdot(h\cdot x)=(gh)\cdot x$ $\forall g\in G, \forall x \in L$. In our theory, $L$ is a set of words: for example, the months of the year, on which the cyclic group $G=C_{12}$ acts. The action is \emph{transitive} if every $x\in L$ can be moved to every $y\in L$ by some $g\in G$. The action is \emph{free} if $g\cdot x = x$ implies $g=1$, i.e, only the identity can fix a $x \in L$. When the action is both free and transitive, $L$ is a \emph{$G$-torsor}: choosing any base point $x_0\in L$ yields a bijection $G\to L$, $g\mapsto g\cdot x_0$. For example, the action of $G=C_{12}$ on the months is free and transitive, such that the months themselves can be thought of as elements of $C_{12}$ after choosing a base point (Figure~\ref{fig:geometries} left). When the action is transitive but not free, each point $x$ has a nontrivial \emph{stabilizer} $K=\{g\in G : g\cdot x = x\}$, and $L$ is identified with the \emph{homogeneous space} $G/K$ of left cosets $gK$. The sphere $L=S^2 = SO(3)/SO(2)$ of the celestial objects is our example of a homogeneous space (Figure~\ref{fig:geometries} right): $SO(3)$ acts transitively on $S^2$, the stabilizer of a point is the group $SO(2)$ of rotations about the axis through the point. 

%A torsor is the special case $K=\{1\}$.

\paragraph{Irreducible representations.} Our theory shows that word embeddings can be computed analytically using irreducible representations of group, which we define now.
% Elements of a group can be realized concretely as invertible matrices, with composition becoming matrix multiplication.
%
A \emph{unitary representation} of $G$ of dimension $d_\rho$ is a map $\rho: G\to U(d_\rho)$ into the unitary group such that $\rho(gh)=\rho(g)\rho(h)$ for all $g,h\in G$. It is \emph{irreducible} (an \emph{irrep}) if no proper nonzero subspace of $\mathbb{C}^{d_\rho}$ is mapped to itself by every $\rho(g)$.
Any unitary representation decomposes into a direct sum of irreps. A finite group has finitely many irreps up to unitary equivalence. All irreps of Abelian groups (e.g., $C_{12}$) are one-dimensional: their $\rho(g)$ are complex scalars for any irrep $\rho$ and group element $g$. Higher dimensional irreps may exist for non-Abelian groups (e.g., $D_{12}$ and $SO(3)$): their $\rho(g)$ are generally complex matrices of size $d_\rho \times d_\rho$. Our theory will often refer to the entries of those matrices. Lastly, irreps are used to define the \textit{group Fourier transform} of any function $f$ over a finite group $G$ as $\hat{f}(\rho)=\sum_{g}f(g)\rho(g)^\dagger$. The sum becomes an integral if $G$ is compact.%We recover the classical DFT when $G=\mathbb{Z}_n$. 

% \paragraph{Group Fourier Transform.} A function on a finite group $G$ has a \emph{group Fourier transform} $\hat{f}(\rho)=\sum_{g}f(g)\rho(g)^\dagger$. The sum becomes an integral if $G$ is compact. We recover the classical DFT when $G=\mathbb{Z}_n$. 

% On a homogeneous space $G/K$ only the columns of $\rho$ that are fixed by $K$ survive, and for $S^2$ these are the spherical harmonics $Y_\ell^m$, $m\in\{-\ell,\dots,\ell\}$, one $(2\ell+1)$-dimensional block per degree. Irreps are complex in general; each has a \emph{type} (real, complex, or quaternionic) given by its Frobenius--Schur indicator, which determines how its matrix elements combine into real-valued functions. The irreps of $\mathbb{Z}_n$ are of complex type, and pairing the modes $k$ and $n-k$ produces the real cosines and sines; those of $D_{12}$ and $SO(3)$ are of real type and require no such pairing.

\section{Theory: from invariant statistics to harmonic geometry}
\label{sec:theory}

We state the framework and main results. Complete statements and proofs are in
Appendices~\ref{app:premise}--\ref{app:homog}.

\paragraph{Word embeddings.} For words $x,y$ in a $n$-word family, we write $P_x, P_y$ their \textit{unigram probabilities} --how often word $x$, resp. $y$, appears in text -- and $P_{xy}$ their \textit{co-occurrence
probability} --how often $x, y$ appear together (formal definitions in Appendix \ref{app:premise}). We define the $n \times n$ \textit{normalized co-occurrence matrix} element-wise by
\begin{equation}
\label{eq:mstar}
\Mstar_{xy} = \frac{P_{xy} - P_xP_y}{\tfrac12(P_{xy} + P_xP_y)},
\end{equation}
and $P \Mstar P$ its \textit{doubly centered version}, where $P = \id - \tfrac{1}{n}\bm{1}\bm{1}^{\top}$ is the
centering projector.

Recent work shows that the matrix $P \Mstar P$ specifies which word embeddings will be learned from text corpora. Specifically, \citet{karkada2025quadratic} established that the $r$ leading PCA scores of a set of $n$ \texttt{word2vec}-style embeddings in $d$ dimensions $\Wbar \in \R^{n \times d}$ can be approximated with the positive eigenpairs $\{\big(\lambda_i, v_i)\}_{i\leq r}$ of $P\Mstar P$, taken in decreasing order. Namely, up to an orthogonal transform and realification, the projection of $\Wbar_x \in \mathbb{R}^d$ along the first $r$ principal components of $\Wbar$ is approximately $\Big(\sqrt{\lambda_i}v_i(x) \Big)_{i\leq r}$. We use this characterization of empirical word embeddings throughout our paper, as also done in \citet{karkada2026symmetry}.

\paragraph{Assumption: Statistical invariance.} Our starting point is the assumption that a given $n$-word family carries
a symmetry, as in the real-world examples of Figure~\ref{fig:geometries}. Specifically, we assume that a symmetry group $G$ acts on a word set $L$ and that the co-occurrence matrix verifies:
\begin{equation}\label{ass:stat-invariance}
    \Mstar_{g\cdot x,\,g\cdot y} = \Mstar_{x,y}, \quad \forall g \in G,\; \forall x,y \in L.
\end{equation}
We say that the word statistics are \emph{invariant} to the group $G$. If the
words $L$ form a \emph{$G$-torsor}, fixing
any word as the group identity\footnote{The base word choice is inconsequential as shown in
Lemma~\ref{lem:base_point}.} creates a bijection between the word set and the group: $L\simeq G$. Abusing notation, Assumption~\ref{ass:stat-invariance} implies that
$\Mstar_{g,h} = C(g^{-1}h)$ for a kernel $C : G \to \R$ with
$C(g^{-1}) = C(g)$ $\forall g,h \in G$, the latter property forced by the symmetry of $\Mstar$. If instead the action is transitive with stabilizer $K$,
the word set is the homogeneous space $L\simeq G/K$ and $C$ is bi-$K$-invariant.

\paragraph{Theory: Structure of word embeddings} Using the characterization of word embeddings and the statistical invariance assumption above, we derive two theorems giving the analytical expression of word embeddings under group symmetry. Their formal versions are in Thm.~\ref{thm:finite-main} and Thm.~\ref{thm:main_homog_space}.

\newcommand{\pc}{\mathrm{pc}}

\begin{theorem}[Finite group, informal]
\label{thm:main-finite}
Let $G$ be a finite group of order $n$, and let the word set $L$ be a
$G$-torsor satisfying Assumption~\ref{ass:stat-invariance}. For each
nontrivial real irrep $\rho$ of $G$, let $\Chat(\rho)$ be the group Fourier coefficient at $\rho$ of the kernel
$C$ associated with $\Mstar$, with eigenvalues
$\lambda^{(1)}_\rho,\dots,\lambda^{(d_\rho)}_\rho$. 
% Set
% \begin{equation*}
%   \mathcal{I}
%   = \bigl\{(\rho,s,a) \;:\; \rho,\
%      \lambda^{(s)}_\rho>0,\ 1\leq a\leq d_\rho \bigr\},
%   \qquad r = |\mathcal{I}|,
% \end{equation*}
% enumerated in decreasing order of $\lambda^{(s)}_\rho$. 
Then, denoting $\pc_r$ the top-$r$ score map $\pc_r\colon\mathbb{R}^{d}\to\mathbb{R}^{r}$, the word embedding $W_g$ of any $g \in G$ verifies:
\begin{equation}
  \pc_r(W_g)
  = \Biggl[\sqrt{\frac{\lambda^{(s)}_\rho\, d_\rho}{n}}\;\rho(g)_{as}
    \Biggr]_{(\rho,s,a)},
\end{equation}
for some choices of $(\rho,s,a)$ sorted in decreasing order of $\lambda^{(s)}_\rho > 0$, and up to orthogonal transform and realificaction.
\end{theorem}

A version of Theorem \ref{thm:main-finite} also holds for compact groups, see Thm. \ref{thm:compact-main}. These theorems demonstrate that word embeddings of a $G$-torsor word set appear like the matrix elements of the irreps of $G$. The text source determines, through its kernel $C$, which columns of the irreps will be most pronounced (largest $\lambda_{\rho}^{(s)})$. Furthermore, if the dimension prescribed to the word embeddings is too small, irrep elements with sufficiently small $\lambda_{\rho}^{(s)}$ are truncated. 

Next, we consider homogeneous spaces. We save the general discussion for the Appendix and focus on the particular case of the sphere $S^2$ for clarity, which is the case relevant for our experiments. 

\begin{theorem}[Homogeneous space $S^2$]
\label{thm:main-GK}
Let $L$ be a finite word set in which each word $x$ is assigned a distinct
point $p_x \in S^2$, and satisfying Assumption~\ref{ass:stat-invariance}. For each degree $\ell \geq 1$, let
$c_\ell = \frac{1}{2}\int_{-1}^{1} C(t)\,P_\ell(t)\,dt$
for $P_\ell$ the degree-$\ell$ Legendre polynomial and $C$ a kernel corresponding to $\Mstar$. Then, denoting $\pc_r$
the top-$r$ score map $\pc_r\colon\mathbb{R}^{d}\to\mathbb{R}^{r}$, the word
embedding $W_x$ of any $x \in L$ satisfies:
\begin{equation}
  \pc_r(W_x)
  = \Bigl[\sqrt{c_\ell}\; Y^{m}_{\ell}(p_x)\Bigr]_{(\ell,m)},
\end{equation}
for some choice of $(\ell,m)$, sorted in decreasing order of $c_\ell > 0$,
where $Y^{m}_{\ell}$ denotes a real spherical harmonic of degree $\ell$, and  and up to orthogonal transform and realificaction. Since there are $2\ell+1$ spherical harmonics of degree $\ell$, the embedding contains subspaces of dimension $2\ell +1$.  
 % In this case, up to a rotation, the leading modes of a word $x$'s embedding takes the form $\Big(\sqrt{c_{\ell}}\; Y^{m}_{\ell}(x)\Big)_{\ell \ge 1,\, |m| \le \ell}$, with $(\ell, m)$ sorted in decreasing order of $c_{\ell} >0$. $Y^{m}_{\ell}(x)$ is a spherical harmonic evaluated at the spherical position corresponding to $x$.  Each $c_{\ell}$ is a corpus derived constant. Formally, $c_{\ell} = \frac{1}{2} \int_{-1}^{1} C(t)P_{\ell}(t)dt$ where $P_{\ell}(t)$ is the $l^{th}$ Legendre polynomial. Note that they are $2\ell+1$ spherical harmonics of degree $\ell$, so the embedding will contain subspaces of dimension $2\ell +1$.  
\end{theorem}

\paragraph{Realification.} Embeddings are real, while the
eigenvectors above are in general complex. Hence, we must convert them to real coordinates, using a methodology that depends on the type of irrep (Proposition~\ref{prop:FS}). In our experiments, irreps can be realified (if needed) in a simple, formulaic way.

\section{The cyclic case: months and the pringle}
\label{sec:months}

\begin{figure}[t]
\centering
\includegraphics[width=1.0\textwidth]{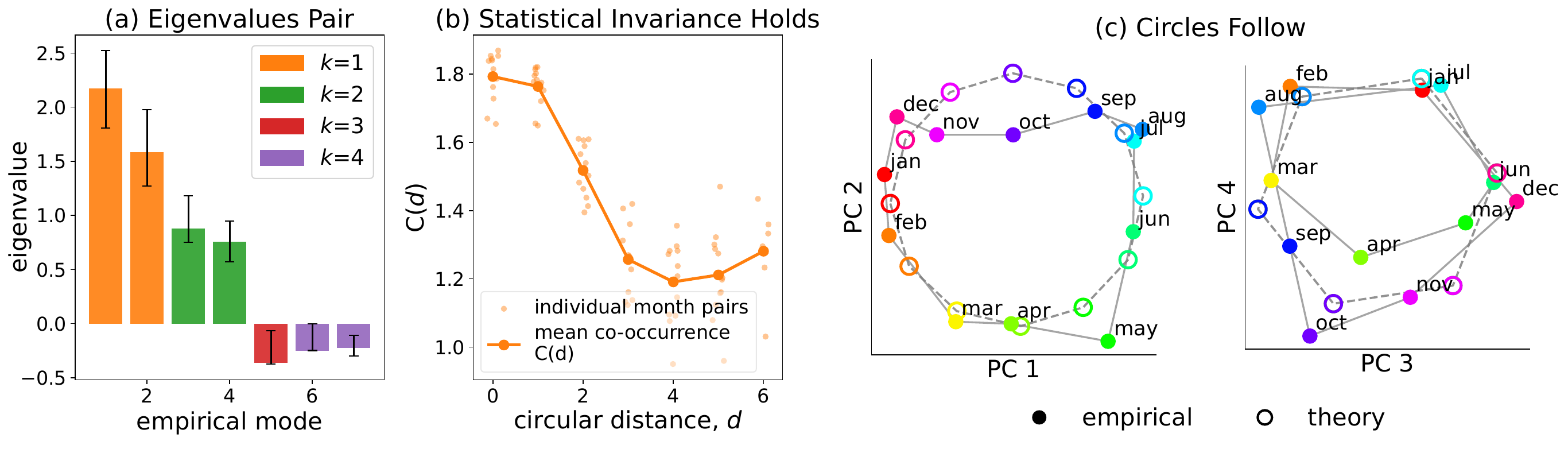}
\caption{\textbf{Month statistics are $\mathbb{Z}_{12}$-invariant $\rightarrow$ embed as circles (known result, re-derived).}
$\Mstar$ is formed using month co-occurrences in the corpus \texttt{text8}, using month-sense disambiguation for ``may.'' The approximate $\mathbb{Z}_{12}$-invariance of $\Mstar$ causes circles to emerge in the months' word embeddings.  See App.~\ref{app:months-details} for all details.  
(a)~Spectrum of $P\Mstar P$ (corpus bootstrap error bars, $2.5\%-97.5\%$), wavenumber assigned by dominant Fourier mode.  The first 4 eigenvalues split into approximate pairs with the same wavenumber, a hint that $\mathbb{Z}_{12}$-invariance could be present in the data.
(b) Averaged kernel $C(d)$ of the individual month pair co-occurrences.  The data follow the kernel rather than just averaging to it (average predicts raw data with $R^2 = 0.87)$, asserting an approximate invariance to $Z_{12}$.
(c)~The two circular planes.  Left: $k=1$ plane with median $31^\circ$/month.  Right: $k=2$ plane with median $66^\circ$/month.}
% (a)~Equivalence class-averaged kernel $c(d)$ of the pairwise month co-occurrences in the corpus \texttt{text8}, using month-sense disambiguation for ``may'' (App.~\ref{app:months-details}). Small dots are the individual month
% pairs entering each class: the data follow the kernel rather than
% just averaging to it (class-averaging $R^2 = 0.87$).
% (b)~Spectrum of the centered $\Mstar$ colored by wavenumber
% (corpus bootstrap error bars App.~\ref{app:months-details}): wavenumber is assigned by which Fourier mode captures the most energy of the eigenvector (App.~\ref{app:months-details}).  The predicted multiplicity-$2$ pairs appear as
% same-color adjacent bars.  (c)~The two irrep planes, empirical
% vs. theory: $30^\circ$/month in the $k{=}1$ plane
% (left, median step $31^\circ$), two laps at $60^\circ$/month in $k{=}2$
% (right, median $66^\circ$).}
\label{fig:months}
\end{figure}

We test the theory on a known case: the cyclic group $G = \Z_{12}$ acting on the twelve months. We utilize the \texttt{text8}
corpus (Wikipedia dump, 17M tokens, \citep{mahoney2011text8}). We recover the same results as \citep{karkada2026symmetry}, but using group theory instead of their Fourier analysis.

\paragraph{Predictions} All irreps of $\Z_{12}$ are 1-dimensional complex Fourier modes,
indexed by a wavenumber $k \in \{0, \ldots, 11  \}$.  
%After a necessary
%realification, and the theory's
%prediction takes a completely explicit form: 
The theory predicts that
the embedding of month
$m \in \{0, \dots, 11\}$ belongs to a direct product of circles of the form $\big(\cos(2\pi km/12),\sin(2\pi km/12) \big)$ with nonzero $k$. The radii of these circles is obtained through the discrete Fourier transform of the kernel $C(d)$, the average co-occurrence of months $d \in \{0,\ldots,6\}$ months apart. Specifically, $\sqrt{\widehat{C}(d)}$ is the radius of the circle with $k=d.$ These radii are the corpus supplied scalars for ranking the irreps.

% High frequencies tend to be less prevalent compared to low frequencies in natural text or images, perhaps why we see the circles corresponding to $k=1$ and $2$ as the leading principal components.

% From Figure \ref{fig:months} A, we see that months closer in the year are more likely to be mentioned together, likely why the $k=1$ and $k=2$ circles are the most prevalent in Figure \ref{fig:months} B.

\paragraph{Results: The embeddings realize the cyclic irreps} As we see in Figure \ref{fig:months} B, the averaged kernel is faithful to the individual months' co-occurrences. Although the statistics will never be perfectly invariant, they are close enough that our theoretical circular predictions line up nicely with the actual embeddings in Figure \ref{fig:months} C. 
The first 4 empirical PCs of the embeddings match the theoretical predictions at a per-month cosine of 0.96 (median 0.97), under one aligning orthogonal transform per plane, compared to a label-shuffle null of mean 0.35.
The classic Pringle lives in the first 3 dimensions of these embeddings, a sinusoidal on top of a circular base.

\section{The nonabelian case: dihedral chord geometry in Bach}
\label{sec:music}

We test the theory on a novel case: the dihedral group $G = D_{12}$ acting on the $24$ major and minor triads, which form a $D_{12}$-torsor under the transposition--inversion group 
$T/I \cong D_{12}$. We utilize the chord sequences from the $371$ Bach chorales, \texttt{music21} \cite{cuthbert2010music21}:
$24$k chord tokens, full sonority vocabulary of $92$ types with the $24$
triads as the analyzed subsystem.  See App.~\ref{app:music-details} for details.

\begin{figure}[t]
\centering
\includegraphics[width=\textwidth]{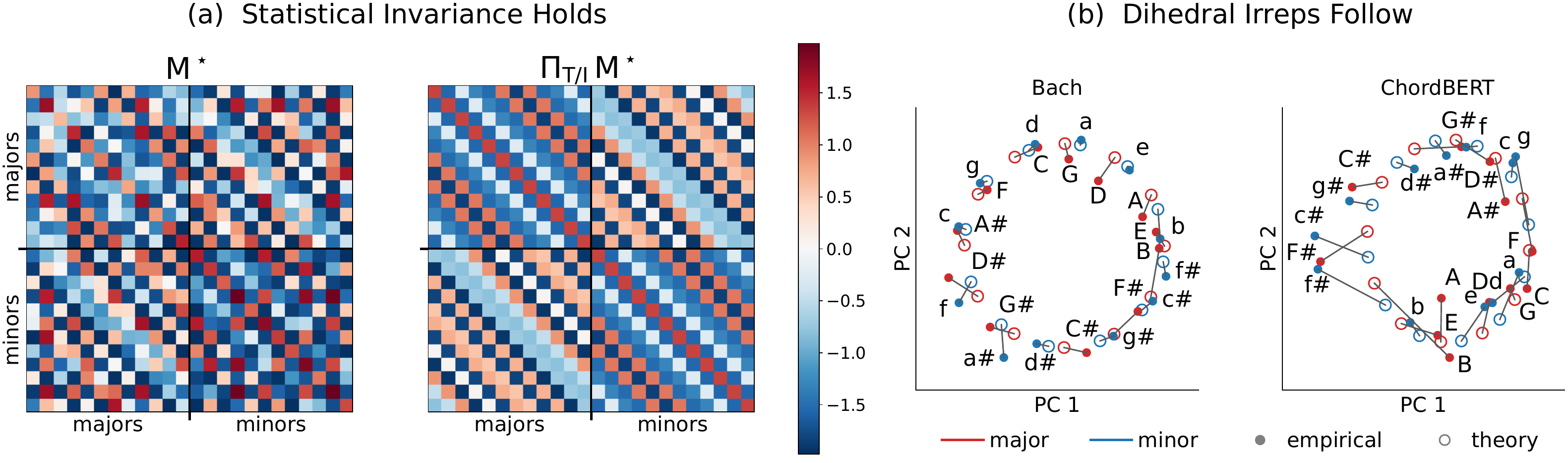}
\caption{\textbf{Chord statistics are $T/I$-invariant $\rightarrow$ embed as dihedral irreps.} See Appendix \ref{app:music-details} for all details. (a)
Left: Empirical $\Mstar$ over the $24$ triads from $371$ Bach chorales.  Majors and minors are listed from $C$ to $B$ in chromatic order.  Right: $\Pi_{T/I} \Mstar$ is the projection of $\Mstar$ onto matrices that are $T/I$-invariant.  Importantly, $\Pi_{T/I} \Mstar$ captures much of the structure of $\Mstar$ ($R^2 = 0.822$), demonstrating an approximate $T/I$-invariance in $\Mstar$. (b) Uppercase and lowercase refer to major and minor chords respectively.  An orthogonal transform is used to align theoretical predictions with empirical embeddings in each case. Left: due to the statistical invariance in (a), a dihedral irrep, in the form of interleaved circles of majors and minors, emerges in the Bach chord embeddings.  Our predictions obtain a per-chord cosine similarity of $0.98$ with the real embeddings.  Right: Our empirical embeddings are the average activation of each triad in the layer 3 (selected for high isotypic energy, Figure \ref{fig:irrep-content_BERT}) residual stream of ChordBERT across all 371 chorales.  The same dihedral structure emerges.  We use the Gram matrix of these centered embeddings (normalized by the root mean square of its diagonal) as a proxy for $\Mstar$.  This gives us theoretical predictions with a per-chord cosine sim. of $0.91$.  
% plane 1 (left) and plane 2 (right).  We form predictions for the first 4 dimensions and solve for the best orthogonal transform to align them with the empirical embeddings.  The cosine similarity between our theoretical predictions and the embeddings are 0.99/0.94 respectively for each plane.}
}
\label{fig:main_music}
\end{figure}

\paragraph{Predictions.} The $24$ major and minor triads form a $D_{12}$-torsor under the transposition-inversion group $T/I \cong D_{12}$.  Hence, Theorem~\ref{thm:main-finite} applies. $D_{12}$ has four $1$-dimensional embeddings and five $2$-dimensional irreps ($E_1\ldots, E_5)$. From foresight, none of the $1$-dimensional irreps are strongly present in the empirical embeddings, so we will neglect them.  We predict that the embeddings of the majors $m \in \{0,\dots, 11\}$ are a direct product of circles of the form $\big(\cos(2\pi hm/12 + \phi), \sin(2\pi hm/12 + \phi) \big)$, while the embeddings of the minors $n \in \{0, \dots, 11\}$ are instead a direct product of circles of the form $\big(\cos(2\pi hn/12 - \phi), \sin(2\pi hn/12 - \phi) \big)$, existing in the same set of planes, where $\phi$ is some constant (dependent on the corpus) and $h \in \{1,\ldots ,5\}$.  Furthermore, we predict that these interleaved circles of majors and minors share the same radius, a corpus determined value computed as in Thm.~\ref{thm:main-finite}. 

% \paragraph{Predictions.} The $24$ major and minor triads form a $D_{12}$-torsor under the transposition--inversion group 
% $T/I \cong D_{12}$. Hence,
% Theorem~\ref{thm:main-finite} applies. $D_{12}$ has four $1$-dimensional and five
% $2$-dimensional irreps $E_1,\dots,E_5$ ($E_k$ corresponding to rotation by
% $30k^\circ$), all of real type. The theory predicts: up to
% three $1$-dim. modes and, for each $k$, \emph{two} $2$-dimensional planes
% (the split eigenvalues of the $2\times2$ Fourier block
% $\Chat(\rho_k)$ fuse to single $4$-dimensional blocks only if $C$ were a class function
% (Corollary~\ref{cor:dihedral}). Within each plane, chords sit on a circle
% traversed at $30k^\circ$ per semitone of transposition, majors and minors
% interleaved by the inversion. The $24$ triads also carry a second symmetry: the neo-Riemannian PLR
% group of transformational theory. The two act at different levels ---$T/I$ is induced from an action on the $12$ pitch classes, so each
% element moves every chord by one uniform rule, while $P$, $L$, $R$ are
% \emph{contextual} operations, each sending a
% triad to its own parallel, leading-tone, or relative neighbor.

\paragraph{Statistical assumption is verified: Chord co-occurrence statistics exhibit dihedral symmetry}
We form $\Mstar$ with a $3$-token window from our collected $24$k chord tokens.  We can find the projection of $\Mstar$ onto matrices that are invariant with respect to some group $G$, $\Pi_{G}\Mstar$, by the element-wise computation $\Pi_{G}\Mstar_{x,y} = \tfrac1{|G|}\sum_{g\in G} \Mstar_{gx,gy}$, i.e., replacing entries that would be equivalent under invariance by their average.  Transposition invariance only, $\Pi_{\mathbb{Z}_{12}}\Mstar_{x,y}$ (26 independent parameters), explains $R^2 = 0.830$ of $\Mstar$.  Full $T/I$ invariance, $\Pi_{T/I}\Mstar$ (19 independent parameters), still explains $R^2 = 0.822$ with 7 less parameters.  It would be impossible for $T/I$ invariance to outperform $\mathbb{Z}_{12}$ as it strictly adds restrictions on which elements must be equal, yet it has near identical performance.  Furthermore, it is not the case that $T/I$ and $\mathbb{Z}_{12}$ score highly due to some trivial reason. A chord-relabeling null gives $p = 5\times10^{-4}$: none of
$2000$ random relabelings yields a
$T/I$-invariance $R^2$ approaching the observed $0.822$.  

% \paragraph{Statistical assumption is verified: Chord co-occurrence statistics exhibit dihedral symmetry}
% We extract $24$k chord tokens from the $371$ Bach chorales (\texttt{music21})
%  and form
% $\Mstar$ with a $3$-token window to verify Assumption~\ref{ass:invariance}. Projecting onto invariant subspaces:
% transposition-only invariance ($\Z_{12}$, $48$ parameters) explains
% $R^2 = 0.830$ of the centered structure; full $D_{12}$ ($24$ parameters)
% explains $0.822$---nearly all of it, at half the parameters; the
% class-function refinement ($9$ parameters) collapses to $0.435$, and
% PLR to $0.442$. The statistics are thus
% left-$D_{12}$-invariant but decisively \emph{not} bi-invariant: the
% empirically operative case is the generic one of
% Theorem~\ref{thm:main-finite}, predicting split planes.  A chord-relabeling permutation null gives $p = 5\times10^{-4}$: none of
% $2000$ random relabelings of the $24$ triads yields a
% $D_{12}$-invariant $R^2$ approaching the observed $0.822$
% (Figure~\ref{fig:heatmap}).

\paragraph{Result I: The embeddings realize the dihedral irreps} We factorize $\Mstar$ and analyze its leading plane.  Our prediction of interleaved circles of major and minors emerges (Figure \ref{fig:main_music} B, left).  The plane's isotypic energy (loosely speaking, value from 0 to 1 on how much a \emph{subspace} corresponds to a given irrep), is very high at 0.958 ($E_5)$, far above a chance level of $d_{\rho}^2/24 = 4/24 \approx 0.17$.  (App.~Fig.~\ref{fig:irrep-content}).  Most importantly, our theoretical predictions for the 24 chords give a \emph{per-chord} cosine of $0.98$.  In fact, our theoretical predictions for the first 3 planes of embeddings (6 dimensions) are all high at $0.98/0.94/0.80$, respectively (Appendx \ref{app:music-details}).

\paragraph{Result II: The circle of fifths}
In particular, the leading plane is \emph{the circle of fifths}, a classic organizing structure of tonal harmony, realized as interleaved circles at the corpus dictated radius.  The circle of fifths
has been found \emph{empirically} in chord embeddings before, on a far
larger and entirely different music source of ${\sim}89$k popular songs
\citep{lahnala2021chord}; the present result \emph{derives} it. It is
a column of the 2-dimensional $E_5$ irrep of $D_{12}$. 

\paragraph{Result III: Same structure in chord representation model}  Obtaining dihedral irreps in our chord embeddings is theoretically guaranteed once statistical invariance is established.  However, in ChordBERT \cite{chordbert}, a 4.71M parameter transformer, the circle of fifths also emerges (Figure \ref{fig:main_music} B, right).  The structures we predict for word embeddings potentially have wide spread appearances across many architectures; if a structure is useful in one setting, it is likely helpful elsewhere.

% \paragraph{Result II: The circle of fifths}
% Figure~\ref{fig:music-overlay} overlays the trained embedding against the
% prediction computed from the symmetrized statistics:
% all $24$ chords, three planes, aligned by a \emph{single global
% rotation}. Per-chord cosine: $0.99 / 0.96 / 0.88$ by plane, $0.965$
% overall. The leading plane is \emph{the circle of
% fifths}---the organizing structure of tonal harmony---realized as two
% interleaved circles (majors and minors offset by the reflection) at the
% corpus-dictated radius; the $E_2$ and $E_3$ planes follow at their
% predicted rates and radii; none of this was fit. The circle of fifths
% has been found \emph{empirically} in chord embeddings before, on a far
% larger and entirely different corpus of ${\sim}89$k popular songs
% \citep{lahnala2021chord}; the present result \emph{derives} it: it is
% the $E_5$ matrix-coefficient plane of $D_{12}$, dominant because the
% chorale kernel's largest Fourier block sits at $k = 5$---a
% corpus-measurable fact, not an axiom.

\paragraph{Competing symmetry models and data efficiency}
Although the chord statistics appears to follow a dihedral ($T/I$) structure, other conjectures for the underlying symmetry could perform even better.
We compare seven other natural hypotheses about the structure of the co-occurrence matrix in \ref{app:music-details}.  Some notable examples include $\mathbb{Z}_{12}$ (transposition-only invariance as before, 26 independent parameters), and the raw empirical $\Mstar$ itself (300 independent parameters). The $T/I$ symmetry has (at least) comparable predictive performance on held-out chorales to each of other models, even those with more parameters. Furthermore, the $T/I$ symmetry is the most efficient hypothesis at gaining predictive power when fit on a fewer number of chorales.  Both of these observations point to $T/I$ being the most stringent form of symmetry the data reliably follows.  

% We compare seven hypotheses about which pairs of chords are forced to
% share a co-occurrence value, from coarse ($9$ parameters) to the
% \emph{unconstrained} model---one free parameter per unordered chord
% pair ($576$), i.e., the empirical co-occurrence matrix itself, fit with
% no structural assumption (definitions in Table~\ref{tab:models}).
% Predicting held-out embeddings (chorale-disjoint halves), $D_{12}$
% scores per-chord cosine $0.942$: statistically tied with the
% unconstrained model ($0.945$) and with its own refinement
% $\mathbb{Z}_{12}$ (transposition-only invariance, $48$ parameters,
% $0.943$), and dominating every structurally \emph{different}
% alternative (App.~Fig.~\ref{fig:model-comparison}a). The tie with
% $\mathbb{Z}_{12}$ is itself a reflection of the symmetry: dropping
% inversion doubles the parameters without improving prediction, so the
% extra freedom buys nothing---the data supports the full dihedral
% invariance. The learning curve makes the mechanism explicit: $D_{12}$
% reaches $0.935$ from \emph{twelve} chorales, while the unconstrained
% model needs the full corpus to catch up---symmetry is a data
% amplifier, trading bias for a $15\times$ variance reduction
% (App.~Fig.~\ref{fig:model-comparison}b). 

% Notably, quantities fixed by
% group structure (plane identities, angles) are flat across a
% $16\times$ window sweep, while the corpus-derived scalars
% (eigenvalues) drift (App.~Fig.~\ref{fig:window-sweep}).

\section{The continuous case: the sky sphere as a harmonic embedding}
\label{sec:sky}

\begin{figure}[t]
\centering
\includegraphics[width=\textwidth]{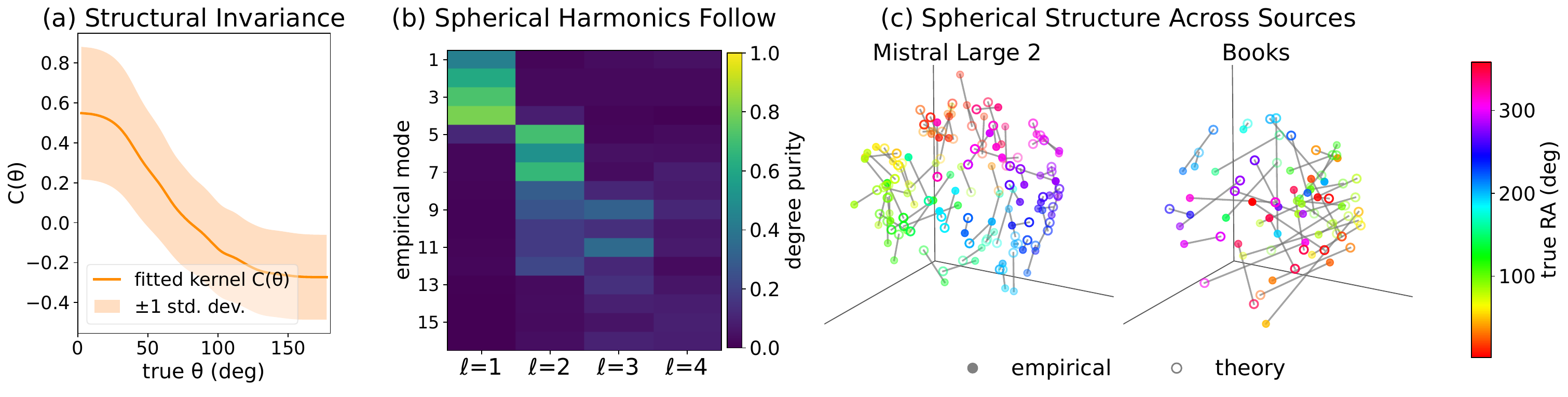}
\caption{\textbf{LLM's structure is function of true celestial angles $\rightarrow$ spherical harmonics emerge, same structure also in text.} Mistral Large 2 ($123$B) residual stream for $188$ sky objects
\citep{berdnikov2026sky} at layer 72 (where positions $\in \mathbb{R}^3$ on the celestial sphere can be best linearly decoded).  See Appendix \ref{app:sky-details} for all details.
(a) Weighted Nadaraya--Watson
regression kernel predicts Gram matrix of LLM activations over objects from actual angular separation. Standard deviation follows the curves shape, suggesting structure is largely dependent on the true angle, our proxy for statistical invariance of $\Mstar$.  
(b) Spherical harmonic purity of the model's PCs: blocks of spherical harmonics appear. $\ell =0$ mode is removed by centering.
(c) Comparison with $\ell =1$ harmonics (sphere) evaluated at true celestial positions after a fitting rotation (and global scaling for the LLM).  Color bar shows right ascension Left: LLM residual steam with a randomly selected 75/188 objects shown. Best-aligned $3$-dimensional 
subspace is found within the first four empirical modes (same search repeated for permutation null).  Per-object cosine sim. of $0.92$ (median $0.97$) against an object permutation null of $0.12$ ($95\%$ $0.17$). Right: Top 3 PCs of book corpus's centered $\Mstar$.  Per-object cosine sim. of $0.55$ (median $0.68$) against a null $0.22$ ($95\%$ $0.32$).}

% (a)~Decode $R^2$: Linear decodability of true celestial positions from residual stream (peaks at layer 72). Gram zonality $R^2$: Measurement of how much activation Gram matrix is a function of only true angle between celestial objects (b)~The layer-$72$ similarity kernel formed from the Gram matrix is largely zonal: the true angular separation alone explains $R^2 = 0.49$ (object location-permutation null gives $0.002$). (c)~Spectrum and degree/spherical harmonic
% purity of the PCs at layer 72, rows shared: the $\lambda$ plot (bars colored
% by each mode's winning degree, $\ell=1$ is dipole, $\ell=2$ is quadrupole) and heatmap both demonstrate the emergence of blocks of spherical harmonics.
% (the $\ell{=}0$ column is removed by centering).  Details are in Appendix \ref{app:sky-details}.}
\label{fig:main_sky}
\end{figure}

\citet{berdnikov2026sky} discovered that large language models, prompted
about \emph{celestial proximity}, represent $188$ stars and constellations on a
decodable sphere in the residual stream, and posed the question of this geometry's origin. Our theory answers this question.

\paragraph{Predictions} The sphere is the homogeneous
space $S^2 = SO(3)/SO(2)$. Although an LLM's residual stream is not strictly in our theory, which considers word embeddings, we use
Theorem~\ref{thm:main-GK} to predict what to expect in its structure. If the ``effective'' statistics are invariant (co-occurrence of celestial objects is a function only of angular separation), each dimension of the representation takes the form of a spherical harmonic.  

% truncated harmonic embedding
% with degree blocks of multiplicity $2\ell{+}1$ and amplitudes
% $\sqrt{\lambda_\ell}$, $\lambda_\ell = 2\pi\!\int_{-1}^{1 } c(t)\,P_\ell(t)$dt, where $t$ is the cosine similarity, $P_\ell$ is the Legendre polynomial of degree $\ell$, and $c(t)$ is the co-occurrence.
The dimensions corresponding to $\ell = 1$, first order spherical harmonics, are precisely a sphere when plotted, as has been seen in \citet{berdnikov2026sky}.  However, our theory further predicts the presence of spherical harmonics of higher orders (e.g. $\ell=2,3,\ldots$).
% We refer to the $\ell=1$ and $\ell=2$ spherical harmonics as their alternative names, ``dipole'' and ``quadrupole'', throughout this section for convenience.

A small technical detail: our theory assumes a uniform sampling on $S^2$, which celestial objects do not exactly obey.  To prevent any of our estimators being strongly influenced by over-sampled regions of the sphere, we employ inverse density weighting (see Appendix \ref{app:sky-details}).
\paragraph{(Effective) statistical assumption is verified: Gram matrix of activations is function of true angular separation.}
Figure~\ref{fig:main_sky} analyzes the released activations for Mistral Large 2, $X \in \mathbb{R}^{188 \times 16}$ (mean across prompts for each object, top 16 PCs, App.~\ref{app:sky-details}). As a proxy for $\Mstar$, we consider the doubly centered, Gram matrix $PXX^TP \in \mathbb{R}^{188 \times 188}$ (normalized by the root mean square of its diagonal).  In place of statistical invariance, we ask for \emph{structural invariance}: off diagonal-elements of this matrix are a function of only the true angular separation between the corresponding celestial objects.  In Figure~\ref{fig:main_sky} A, we see this approximately holds.  The standard deviation of a regressed kernel, predicting the upper triangle of the Gram matrix from objects' true angular separation, hugs the curve's shape.  As further evidence, an averaging of all the upper triangular elements, grouped in bins of $10^{\circ}$, gives an $R^2 = 0.49$ for predicting the original matrix (against a object permutation null of $0.002$).

% Figure~\ref{fig:ma} analyzes their released
% activations for Mistral Large 2 (mean across prompts for each object, density-weighted
% estimators throughout, App.~\ref{app:sky-details}). At layer 72, where true celestial positions are best linearly decoded,
% the Gram matrix (cosine sim. between activations for different objects) is $49\%$ a function of true angular separation alone.

\paragraph{Results: The embeddings realize the spherical harmonics} 
% The embeddings' spherical harmonic spectrum is steeply decaying and effectively
% band-limited ($\lambda_{1..4} = 1.70, 0.37, 0.04, 0.004$). 

As seen in Figure \ref{fig:main_sky} B the
eigenmodes form \emph{degree-ordered harmonic blocks}: modes $1$--$4$ are
$\ell{=}1$, $5$--$8$ are $\ell{=}2$, and then $\ell{=}3$ emerges. A completely random direction would have a purity on the order of $10^{-2}$ so the values in this heatmap are far above chance.  Five of seven
models in \citet{berdnikov2026sky} demonstrate the same spherical harmonic ordering (Table~\ref{tab:sky-models}).  Notably, the $\ell{=}1$ energy occupies four modes rather
than three because PC $1$ is largely occupied by a star/constellation type
direction, as noted by \citet{berdnikov2026sky}.

% The degree $\ell=1$ purity per PC is $0.43, 0.65, 0.72, 0.81$, i.e, the
% sphere is rotated inside the top-$4$ package by a single superposed
% semantic direction, which motivates locating blocks by canonical angles
% rather than fixed PC slices.
% ($D = 0.22$--$0.45$: Olcott
% $1907$, $1912$; Serviss $1888$; Bonwick $1866$; $131$k tokens in total)

\paragraph{Results: The same structure appears in an independent corpus.}
The harmonic structure should be visible in text embeddings themselves, specifically in the \emph{genre}
whose discourse is organized by sky position: star observing
guides.  However, many astronomy books are organized by topics such as mythology or famous constellations rather than by complete ``tours of the sky''.  Hence, from 24 open-source general astronomy books, we utilize the metric
$D = \operatorname{corr}(\Mstar_{ij}, \cos\theta_{ij})$ with a one-sided location-permutation $p < 0.05$ (discussed in
App.~\ref{app:sky-details}) to select our corpus.  Although passing this criterion would certainly be required for developing spherical harmonics in the embeddings, linear correlation with the true cosines does not by itself ensure the structure will be present. Four guides pass ($D$ from $0.22$ to $0.45$, $131$k tokens in total).  Volume is not enough to score highly on this metric, \emph{genre} is critical: \texttt{text8} ($17$M tokens, $130\times$ the guides)
gives $D = 0.01$ ($p = 0.48$). \texttt{text8}'s counts are dominated by non-sky versions of the terminology.
(Table~\ref{tab:astrobooks}; App.~\ref{app:sky-details}). Notably, the guides'$\Mstar$ over $39$ well-counted objects yields a
kernel of similar form as the model's activation kernel (correlation $0.95$ with Figure \ref{fig:main_sky} A). 

Figure~\ref{fig:main_sky} C compares the LLM activations in layer 72 (where celestial positions are best linearly decoded) and the book corpus's embeddings with degree-1 spherical harmonics evaluated at the celestial object's true positions.  For the books, we view the first 3 PCs of its embeddings, while for the model, we look at the most $\ell=1$ aligned $3$D-subspace in its first 4 PCs, hypothesizing that the structure could be \emph{smeared} across dimensions. 
% Note that the dipole structure in the books is approximately invariant to fluctuations in our choice of gate.  Using the top 3, 4, or 5 books all yield very similar results. 
In both settings the spherical structure emerges, although in a much sharper fashion for the model.  We perform the same analysis for degree-2 harmonics.  Locating the most $\ell=2$ aligned subspace in PCs 5-12 of the model (based on Figure \ref{fig:main_sky} B), we obtain a per-object cosine similarity of $0.73$ (median $0.82$) compared to a object permutation null of $0.17$ ($95\%$:
$0.21$).  Similarly, looking at PCs 4-8 of the books' centered $\Mstar$, we see a per-object cosine sim. of $0.39$ (median $0.44)$ against a null of $0.27$ ($95\%$ $0.35)$.  In either case, the $\ell =2$ structure is weaker than that of $\ell =1$, but it is certainly still present.

\begin{wrapfigure}[26]{r}{0.55\textwidth}
  \centering
  \vspace{-12pt}   % pull up toward the paragraph top
  \includegraphics[width=0.53\textwidth]{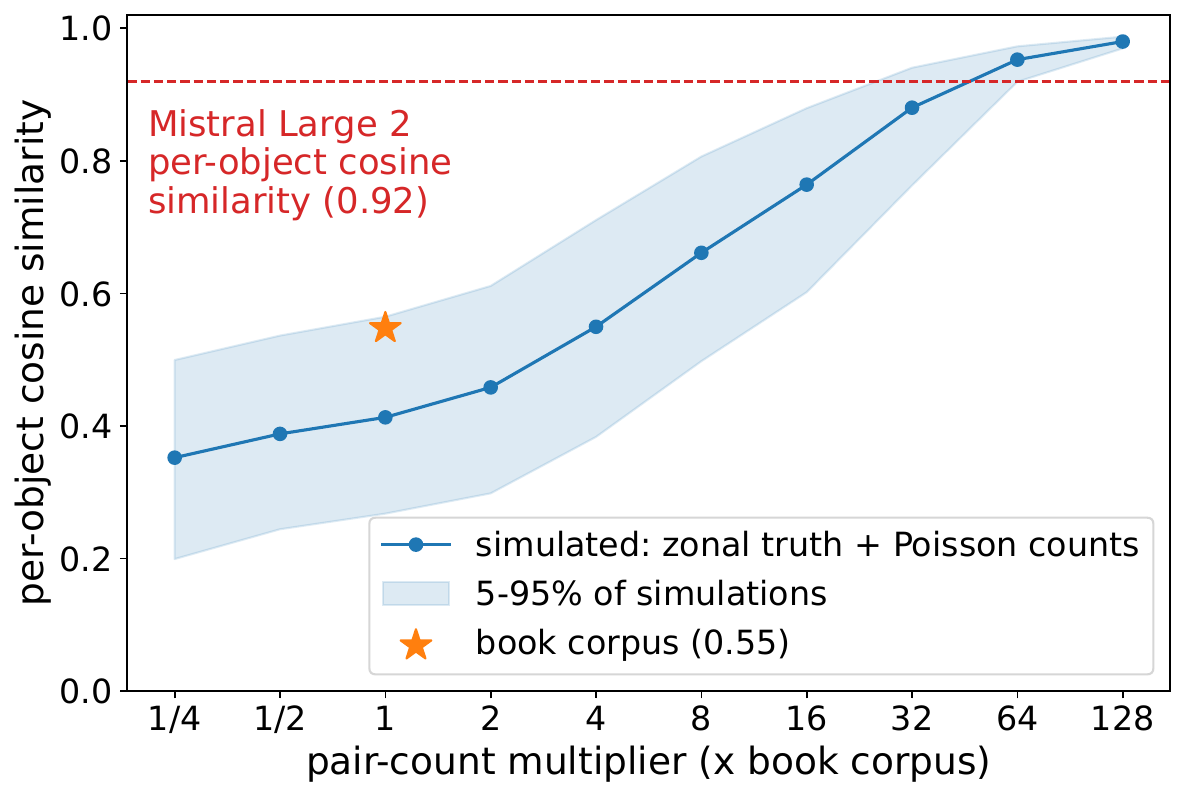}
  \caption{\textbf{Weaker harmonic structure in book corpus could be a sample size effect.} We
take the zonal kernel fit to the books as ground truth and simulate new synthetic $M^{\star}$'s, using Poisson counts for off-diagonal elements at various
multiples of the corpus's size. (full methodology in App.~\ref{app:sky-details}).  
We measure how well the simulated data aligns with dipole spherical harmonics, using the same analysis pipeline as in Figure \ref{fig:main_sky}.
Per-object cos. sim. (mean with
5--95\% band) captures the observed corpus (star) at $1\times$ and
crosses the LLM's performance ($0.92$, dashed) at ${\sim}32$--$64\times$ the books' counts: an \emph{effective corpus
size} for the model.}

  \label{fig:size-effect}
  \vspace{-10pt}   % reclaim space below
\end{wrapfigure}

\paragraph{Book corpus's weaker harmonic structure results from less data:}
It may at first seem surprising that the books exhibit a lesser strength of spherical harmonic structure than the LLM residual stream.  We argue that the book corpus's performance is typical for its size.  We simulate new data assuming the zonal kernel fit to the books exactly specifies the true underlying statistics, and see how well dipole harmonics tend to align with the synthetic data (details in \ref{app:sky-details}).  Figure \ref{fig:size-effect} demonstrates that our book corpus is actually a typical (if not good) draw at its size ($\approx 93\%$ from 1000 simulations).  At around roughly ${\sim}32$--$64\times$ the corpus size, we would expect the dipole structure in the textual embeddings to be just as present as in the model.  The idea that a $123$B parameter LLM could have consumed this amount of relevant stargazing material is very plausible. Hence, a possible explanation for the difference in structure quality between the model and textual embeddings is simply data volume.

% In the corpus, the raw factorization of $\Mstar$ places the
% objects on the celestial sphere at per-object cosine $0.51$ (median
% $0.69$; shuffle null mean $0.21$/ $95$\% $0.32$ from 300 shuffles): the geometry is genuinely present in $131$k
% tokens of text, at the accuracy its size permits---under the fitted
% kernel as ground truth, Poisson pair-count simulations reproduce the
% observed agreement at the corpus's actual count level
% (Appendix~\ref{app:sky-details}, Figure~\ref{fig:size-effect}), so the
% corpus recovers the sphere about as well as a typical finite sample of
% its size could. In the model, the same comparison is far sharper
% (mean cosine $0.93$, median $0.98$; shuffle null mean $0.12$/$95$\%$0.17$ from 300 shuffles), with the
% quadrupole block also expressed (mean cosine ?, median; shuffle null mean ? / $95$ \% from 300 shuffles), App.~\ref{app:sky-details}) The gate does not manufacture this agreement: these numbers are largely invariant to the selection cut (top-$3$/$4$/$5$ book
% sets essentially agree), and the simulations above show the gated corpus scores as
% a typical draw, not a lucky one. The interpretive summary: the co-occurrence statistics of a small,
% interpretable genre are approximately $SO(3)$-invariant, and our theory
% dictates the geometric consequence---a harmonic embedding. That
% embedding appears from the book corpus, and far more sharply in a large model's activations. We conjecture the gap is data volume:
% the model has absorbed vastly more such relevant text from which it can optimize its form.

% \begin{figure}[t]
% \centering
% \includegraphics[width=0.75\textwidth]{figures/appendix_size_effect_2_nopin.pdf}
% \caption{\textbf{Weaker harmonic structure in book corpus could be a sample size effect.} We
% take the zonal kernel fit to the books as ground truth and simulate new synthetic $M^{\star}$'s, using Poisson counts for off-diagonal elements at various
% multiples of the corpus's size. (full methodology in App.~\ref{app:sky-details}).  
% We measure how well the simulated data aligns with dipole spherical harmonics, using the same analysis pipeline as in Figure \ref{fig:main_sky}.
% Per-object cos. sim. (mean with
% 5--95\% band) captures the observed corpus (star) at $1\times$ and
% crosses the LLM's performance ($0.92$, dashed) at ${\sim}32$--$64\times$ the books' counts: an \emph{effective corpus
% size} for the model's sky statistics}
% \label{fig:size-effect}
% \end{figure}

\section{Discussion}
\label{sec:discussion}

\textbf{A unifying theory of structured embeddings.} A Pringle traced by the months,
the circle of fifths in Bach's chords, and a globe of spherical
harmonics in a language model's activations initially appear unrelated. However, all three are predictions from the same theory. Underlying symmetry in the statistics of data, when present, is a powerful restraint on permissible embeddings.
% \textbf{Context selects the effective statistics.} ``The'' co-occurrence
% structure is not a property of tokens but of a discourse: sky-tour prose
% carries the zonal kernel, while topical astronomy text dilutes it, and
% kernel agreement with the model \emph{peaks} at a small pure-genre corpus
% rather than growing with tokens (App.~\ref{app:sky-details}). The
% model's proximity-prompted representation matches the genre statistics,
% suggesting prompts condition the model onto a sub-corpus in the
% statistical sense. Symmetrization is the mechanism that makes small
% corpora predictive: it estimates only the invariant component---a few
% generalized Fourier coefficients---which is exactly the component that
% survives scaling; hence a single field guide predicts a $123$B model.

% \textbf{Deviations are informative.} Where predictions degrade, the
% theory says why: spectral gaps against the invariance-breaking scale
% (Davis--Kahan) grade the chord planes; superposition with one semantic
% direction rotates the sky dipole; finite pair counts smear the books'
% own factorization by a computable amount. The residual,
% non-invariant statistics (mythological chord-pair and star-pair
% affinities; object fame) are measured, not assumed away, and set natural
% targets for finer structure.

\textbf{Limitations} The theory's scope is set by its
premise: it considers only families of words whose co-occurrence
statistics are, at least approximately, invariant under some group---a
strong structural property that many word families will simply not
have. Months, chords, and celestial objects were chosen because their
statistics demonstrably carry such an invariance. Yet, for most of language such structure might not be immediately clear. For LLMs there is the further caveat that our
theorems concern trained word embeddings, and their application to
activations rests on an empirical premise.

\subsubsection*{Acknowledgments}
We thank Hasan Atakan Bedel, Qiqi Xian, Katrin Franke, and Imran Thobani for their useful comments on the manuscript.  A.T. acknowledges support from the Amaranth Foundation (Enigma Project).  N.M. acknowledges funding from the NSF grant 2313150 and partial support from a Goodfire Research Grant with access to Silico, Goodfire’s interpretability research agent.

\bibliographystyle{iclr2027_conference}
\bibliography{refs}

@misc{chordbert,
      title={Modeling Stylistic Co-evolution in Symbolic Music Heritage Collections}, 
      author={Yulong He and Ivan Smirnov and Yanming Li},
      year={2026},
      eprint={2607.23957},
      archivePrefix={arXiv},
      primaryClass={cs.SD},
      url={https://arxiv.org/abs/2607.23957}, 
}

@book{serre1977linear,
  title     = {Linear Representations of Finite Groups},
  author    = {Serre, Jean-Pierre},
  series    = {Graduate Texts in Mathematics},
  volume    = {42},
  year      = {1977},
  publisher = {Springer-Verlag},
  address   = {New York},
  isbn      = {978-0-387-90190-9}
}

@inproceedings{marchetti2026sequential,
  title={Sequential Group Composition: A Window into the Mechanics of Deep Learning},
  author={Marchetti, Giovanni Luca and Kunin, Daniel and Myers, Adele and Acosta, Francisco and Miolane, Nina},
  booktitle={International Conference on Machine Learning (ICML)},
  year={2026}
}

@article{karkada2026symmetry,
  title={Symmetry in language statistics shapes the geometry of model representations},
  author={Karkada, Dhruva and Korchinski, Daniel J and Nava, Andres and Wyart, Matthieu and Bahri, Yasaman},
  journal={arXiv preprint arXiv:2602.15029},
  year={2026}
}

@inproceedings{karkada2025quadratic,
 author = {Karkada, Dhruva and Simon, James and Bahri, Yasaman and DeWeese, Michael},
 booktitle = {Advances in Neural Information Processing Systems},
 doi = {10.52202/085713-0140},
 pages = {3967--4000},
 publisher = {Curran Associates, Inc.},
 title = {Closed-Form Training Dynamics Reveal Learned Features and Linear Structure in Word2Vec-like Models},
 volume = {38, Main Conference},
 year = {2025}
}

@misc{berdnikov2026sky,
      title={Sky sphere representation in language models}, 
      author={Aleksandr Berdnikov and Yevgeny Liokumovich},
      year={2026},
      eprint={2607.27092},
      archivePrefix={arXiv},
      primaryClass={cs.LG},
      url={https://arxiv.org/abs/2607.27092}, 
}

@inproceedings{engels2024not,
  title={Not all language model features are one-dimensionally linear},
  author={Engels, Josh and Michaud, Eric and Liao, Isaac and Gurnee, Wes and Tegmark, Max},
  booktitle={International Conference on Learning Representations},
  volume={2025},
  pages={84591--84622},
  year={2025}
}

@inproceedings{gurnee2024language,
 author = {Gurnee, Wes and Tegmark, Max },
 booktitle = {International Conference on Learning Representations},
 pages = {2483--2503},
 title = {Language Models Represent Space and Time},
 volume = {2024},
 year = {2024}
}

@inproceedings{levy2014neural,
 author = {Levy, Omer and Goldberg, Yoav},
 booktitle = {Advances in Neural Information Processing Systems},
 pages = {},
 publisher = {Curran Associates, Inc.},
 title = {Neural Word Embedding as Implicit Matrix Factorization},
 volume = {27},
 year = {2014}
}

@inproceedings{pennington2014glove,
    title = "{G}lo{V}e: Global Vectors for Word Representation",
    author = "Pennington, Jeffrey  and
      Socher, Richard  and
      Manning, Christopher",
    booktitle = "Proceedings of the 2014 Conference on Empirical Methods in Natural Language Processing ({EMNLP})",
    month = oct,
    year = "2014",
    address = "Doha, Qatar",
    publisher = "Association for Computational Linguistics",
    doi = "10.3115/v1/D14-1162",
    pages = "1532--1543"
}

@InProceedings{park2024linear,
  title = 	 {The Linear Representation Hypothesis and the Geometry of Large Language Models},
  author =       {Park, Kiho and Choe, Yo Joong and Veitch, Victor},
  booktitle = 	 {Proceedings of the 41st International Conference on Machine Learning},
  pages = 	 {39643--39666},
  year = 	 {2024},
  volume = 	 {235},
  series = 	 {Proceedings of Machine Learning Research},
  publisher =    {PMLR},
}

@inproceedings{park2024geometry,
 author = {Park, Kiho and Choe, Yo Joong and Jiang, Yibo and Veitch, Victor},
 booktitle = {International Conference on Learning Representations},
 pages = {76441--76463},
 title = {The Geometry of Categorical and Hierarchical Concepts in Large Language Models},
 volume = {2025},
 year = {2025}
}

@inproceedings{nanda2023progress,
  title={Progress measures for grokking via mechanistic interpretability},
  author={Nanda, Neel and Chan, Lawrence and Lieberum, Tom and Smith, Jess and Steinhardt, Jacob},
  booktitle={International Conference on Learning Representations},
  year={2023}
}

@inproceedings{chughtai2023toy,
  title={A toy model of universality: Reverse engineering how networks learn group operations},
  author={Chughtai, Bilal and Chan, Lawrence and Nanda, Neel},
  booktitle={International Conference on Machine Learning},
  pages={6243--6267},
  year={2023},
  organization={PMLR}
}

@book{diaconis1988group,
  title={Group representations in probability and statistics},
  author={Diaconis, Persi},
  publisher={Institute of Mathematical Statistics},
  year={1988}
}

@book{folland2016course,
  title={A Course in Abstract Harmonic Analysis, Second Edition},
  author={Folland, G.B.},
  isbn={9781498727136},
  lccn={2015020328},
  series={Textbooks in Mathematics},
  year={2016},
  publisher={Taylor \& Francis Group}
}

@book{dai2013approximation,
  title={Approximation theory and harmonic analysis on spheres and balls},
  author={Dai, Feng and Xu, Yuan},
  publisher={Springer},
  year={2013}
}

@article{crans2009musical,
  title={Musical actions of dihedral groups},
  author={Crans, Alissa S and Fiore, Thomas M and Satyendra, Ramon},
  journal={The American Mathematical Monthly},
  volume={116},
  number={6},
  pages={479--495},
  year={2009},
  publisher={Taylor \& Francis}
}

@inproceedings{cuthbert2010music21,
  author = {Cuthbert, Michael Scott and Ariza, Christopher},
  booktitle = {ISMIR},
  editor = {Downie, J. Stephen and Veltkamp, Remco C.},
  ee = {http://ismir2010.ismir.net/proceedings/ismir2010-108.pdf},
  isbn = {978-90-393-53813},
  pages = {637-642},
  publisher = {International Society for Music Information Retrieval},
  title = {Music21: A Toolkit for Computer-Aided Musicology and Symbolic Music Data.},
  url = {http://dblp.uni-trier.de/db/conf/ismir/ismir2010.html#CuthbertA10},
  year = 2010
}

@misc{mahoney2011text8,
  title        = {Large text compression benchmark ({text8})},
  author       = {Mahoney, Matt},
  howpublished = {\url{http://mattmahoney.net/dc/textdata.html}},
  year         = {2011},
}

@InProceedings{lahnala2021chord,
author="Lahnala, Allison
and Kambhatla, Gauri
and Peng, Jiajun
and Whitehead, Matthew
and Minnehan, Gillian
and Guldan, Eric
and Kummerfeld, Jonathan K.
and {\c{C}}amc{\i}, An{\i}l
and Mihalcea, Rada",
title="Chord Embeddings: Analyzing What They Capture and Their Role for Next Chord Prediction and Artist Attribute Prediction",
booktitle="Artificial Intelligence in Music, Sound, Art and Design",
year="2021",
publisher="Springer International Publishing",
address="Cham",
pages="171--186",
isbn="978-3-030-72914-1"
}

\appendix
\section*{Appendix Overview}

% \citet{karkada2026symmetry}\ show that when co-occurrence statistics are invariant under
% translations of a latent cyclic lattice $\Z_L^D$, the optimal word embeddings are
% Fourier features whose wavenumbers and relative
% amplitudes are dictated by the spectrum of the statistics. We
% develop the general theory of which that result is a special case for the cyclic group.
% The overarching principle is that \emph{the symmetry group of the statistics
% determines the geometry of the
% embedding}.  

Appendices~\ref{app:premise}--\ref{app:homog} contain the complete theory while Appendices~\ref{app:months-details}--\ref{app:sky-details} contain all experimental details.  Specifically, the appendix is organized as follows. 

Appendix~\ref{app:premise} discusses how \texttt{word2vec}-style embeddings can be approximated by an eigendecomposition of a co-occurrence matrix of the words, and states the ``statistical invariance under a group'' assumption necessary for our theory to apply.
Appendix~\ref{app:finite} considers embedding structure when the words are invariant under a finite group (Theorem~\ref{thm:finite-main}), and considers two special cases: the cyclic and dihedral cases, which are relevant to our ``months'' and ``chords'' experiments.
Appendix~\ref{app:compact} generalizes the theory to compact groups. Appendix~\ref{app:homog} further treats homogeneous
spaces $G/K$, considering the special case of Gelfand Pairs and in particular $S^2$, which is relevant to our sky-sphere experiments.

Appendices~\ref{app:months-details},~\ref{app:music-details},~\ref{app:sky-details} give the experimental details for the month, music, and sky-sphere experiments, respectively.  Finally, Appendix~\ref{app:repro} discusses reproducibility, while Appendix~\ref{app:AI_use} gives our AI use statement.

\section{Word Embeddings and Statistical Invariance}
\label{app:premise}

\subsection{From statistics to embeddings}

Let $L$ be a finite set of $n$ words. Following \citet{karkada2026symmetry}, for arbitrary words $x,y \in L$ we consider the \emph{normalized co-occurrence matrix} defined by 
\begin{equation}
\label{eq:mstar-def}
\Mstar_{xy} \;=\; \frac{P_{xy} - P_x P_y}{\tfrac12\,(P_{xy} + P_x P_y)},
\end{equation}
where $P_{xy}$ is the probability that words $x, y$ co-occur within a
context window and $P_x$ is the unigram probability (both defined below). Since
$P_{xy} = P_{yx}$, the matrix $\Mstar$ is symmetric. 

\paragraph{Estimation conventions.} Throughout the paper, $P_{xy}\; \forall x,y \in L$ and $P_x\; \forall x \in L$ are
estimated from a respective corpus $\mathcal{C}$ of $N$ tokens as follows. Fix a function
$f(\delta) \ge 0$ supported on token gaps $\delta = 1, \dots, B$, with total mass
$F := \sum_{\delta=1}^{B} f(\delta)$. We define a weighted count matrix
\[
W_{x,y} = \sum_{i} \delta_{\mathcal{C}[i],x} \sum_{d=1}^B f(d)(\delta_{\mathcal{C}[i+d],y} + \delta_{\mathcal{C}[i-d],y}   )
% W_{xy} \;=\; \sum_{\substack{p < q,\; q - p \le B \\ \text{tokens } x,y \text{ at } p,q}}
% f(q - p),
\]
where $\mathcal{C}[i]$ is the $i^{th}$ token in the corpus.  For multiple corpora, we can form a single $W_{xy}$ through summation. Note that this definition enforces $W_{xy} = W_{yx}$.
Then, with additive smoothing $\alpha \ge 0$, and the number of mentions of word $x$ being $n_x$ we define
\begin{equation}
\label{eq:estimators}
P_{xy} \;=\; \frac{W_{xy} + \alpha}{2FN},
\qquad
P_x \;=\; \frac{n_x}{N},
\end{equation}
so that $2FN$ is the total window mass ($N$ tokens, each opening a
two-sided window of mass $2F$, neglecting edge effects). We can further define an
\emph{opportunity normalized} co-occurrence rate:
\begin{align}
\label{eq:R}
R_{xy} \;=\; \frac{ P_{xy}}{ P_x  P_y}
\;=\; \frac{W_{xy} + \alpha}{n_x n_y \cdot 2F/N}
\;=\; \frac{W_{xy} + \alpha}{\mathbb{E}_0[W_{xy}]},
\end{align}
where $\mathbb{E}_0[W_{xy}] = n_x n_y \cdot 2F/N$ is the expected value of $W_{x,y}$ when mentions are placed independently (neglecting edge effects). We have that
$\Mstar_{xy} = 2(R_{xy}-1)/(R_{xy}+1) = 2\tanh(\tfrac12 \ln R_{xy})$, so $\Mstar$ is strictly increasing in $R_{xy}$ and saturates at $\pm 2$. 
% Pairs that never
% co-occur enter as negative evidence graded by opportunity, at
% $R_{xy} = \alpha/\mathbb{E}_0[W_{xy}]$. 

Per-system choices of
$(B, f, \alpha)$ are stated in the experimental appendices (months:
$B = 16$, $f(\delta) = B{+}1{-}\delta$, $\alpha = 0$; chords: $B = 3$,
flat: $f(\delta) = 1$, $\alpha = 0.5$; sky: $B = 30$, flat: $f(\delta) = 1$, $\alpha = 0.25$).

\paragraph{Word Embeddings}
\label{ass:premise} 

\cite{karkada2025quadratic} shows that the matrix $P \Mstar P$ can be used to generate word embeddings. Specifically, the $r$ leading PCA scores of a set of $n$ \texttt{word2vec}-style embeddings in $d$ dimensions $\Wbar \in \R^{n \times d}$  can be approximated with the positive eigenpairs $\{\big(\lambda_i, v_i)\}_{i\leq r}$ of $P\Mstar P$, taken in decreasing $\lambda_i$ order. Up to an orthogonal transform and realification, the projection of $\Wbar_x \in \mathbb{R}^d$ (for word $x$) along the first $r$ principal components of $\Wbar$ is approximately $\Big(\sqrt{\lambda_i}v_i(x) \Big)_{i\leq r}$.  This is how we generate all empirical word embeddings in this paper.

Furthermore, transforming the generation of word embeddings into an eigendecomposition is incredibly fruitful for theory.  Whenever $\Mstar$ obeys certain symmetries, we can theoretically predict its eigenpairs, which we discuss in the upcoming sections.

\subsection{Statistical Invariance under a Group}
\label{app:torsor}

% We consider our set of words $L$ to support a group 

\begin{definition}[Torsor]
\label{def:torsor}
A (left) $G$-\emph{torsor} is a set $L$ with a (left) action of $G$
that is free and transitive: for every $x, y \in L$ there is
exactly one $g \in G$ with $g\cdot x = y$. A torsor is ``a group that has
forgotten its identity element'': it becomes isomorphic to $G$ once a base
point is chosen, but no choice is canonical.
\end{definition}

\begin{example}
The $24$ major and minor triads form a torsor under the 
transposition--inversion group $T/I \cong D_{12}$ and the $12$ months form a
$\Z_{12}$-torsor. 
\end{example}

\begin{definition}[Statistical invariance under a group]
\label{ass:invariance}
We say that a $G$-torsor word set $L$ is statistically invariant under the group $G$ if its $\Mstar$ obeys 
\[
\Mstar_{g\cdot x,\; g \cdot y} \;=\; \Mstar_{x, y}
\qquad \text{for all } g \in G,\; x,y \in L.
\]
\end{definition}

\begin{lemma}[Kernel]
\label{lem:kernel}
Consider a $G$-torsor word set $L$.  Fix a base point $x_0 \in L$ and identify $L \cong G$
via $g \mapsto g\cdot x_0$. Then, abusing notation, Definition~\ref{ass:invariance} is
equivalent to the existence of some (kernel) $C : G \to \R$ with
\begin{equation}
\label{eq:CG}
\Mstar_{g,h} \;=\; C(g^{-1}h)\;\, \forall g,h \in G,
\end{equation}
and symmetry of $\Mstar$ forces $C(g^{-1}) = C(g)\;\forall g \in G$ (note we change from the main text convention of denoting $1$ as identity to the more standard $e$).
\end{lemma}

\begin{proof}
First, we assume that $\Mstar$ is invariant under the group $G$.  Then we define $C : G \rightarrow \mathbb{R}$ by $C(g) = \Mstar_{e,g} \; \forall g \in G$.  Hence, from statistical invariance,  for any $g,h \in G$, $\Mstar_{g,h} = \Mstar_{e,g^{-1}h} = C(g^{-1}h)$.

Now, we assume that $\Mstar_{g,h} = C(g^{-1}h)\; \forall g,h \in G$.  Hence, for any $a\in G$, we have that $\Mstar_{a\cdot g, a \cdot h} =  C(g^{-1}a^{-1}ah) = C(g^{-1}h) = \Mstar_{g,h}$, which is precisely statistical invariance under $G$.  
Finally, $C(g^{-1}h) = \Mstar_{g,h} = \Mstar_{h,g} = C(h^{-1}g) = C((g^{-1}h)^{-1})\; \forall g,h \in G$, which gives us that $C(g) = C(g^{-1})\; \forall g\in G$.
\end{proof}

\section{Finite groups}
\label{app:finite}

Throughout this section, $G$ is a finite group with $|G| = n$.  Furthermore, we assume we have a $G$-torsor word set $L$ whose $\Mstar$ in statistically invariant under $G$.  Under some arbitrary choice of base point, this allows us to associate a real kernel $C : G \rightarrow \mathbb{R}$ with $\Mstar$ as in Lemma \ref{lem:kernel}.

% Throughout this section $G$ is a finite group, $n = |G|$, and
% $H := \Mstar$ under the identification of Lemma~\ref{lem:gauge}, so
% $H_{g,h} = C(g^{-1}h)$ with $C:G\to\R$, $C(g^{-1}) = C(g)$. 

% We regard $H$
% as an operator on the Hilbert Space $\ell^2(G) = \{f : G \to \C\}$ with inner product
% $\inner{f_1}{f_2} = \sum_g f_1(g)\overline{f_2(g)}$:
% \[
% (Hf)(g) \;=\; \sum_{h \in G} C(g^{-1}h)\, f(h).
% \]

\subsection{Harmonic analysis preliminaries}

Let $\Ghat$ denote a complete set of inequivalent \emph{unitary}
irreducible representations $\rho : G \to U(d_\rho)$, so that
$\sum_{\rho \in \Ghat} d_\rho^2 = n$. We note the Schur orthogonality
relations for matrix coefficients: for $\rho, \sigma \in \Ghat$,
\begin{equation}
\label{eq:schur}
\sum_{g \in G} \rho_{ab}(g)\,\overline{\sigma_{ce}(g)}
\;=\; \frac{n}{d_\rho}\,\delta_{\rho\sigma}\,\delta_{ac}\,\delta_{be},
\end{equation}
and the generalized Fourier transform of $C$,
\begin{equation}
\label{eq:fourier}
\Chat(\rho) \;=\; \sum_{g \in G} C(g)\,\rho(g) \;\in\; \C^{d_\rho \times d_\rho}.
\end{equation}
where $d_\rho$ is the dimension of the irrep $\rho$.
\begin{lemma}
\label{lem:hermitian}
Under $C$ real with $C(g^{-1}) = C(g)$, each $\Chat(\rho)$ is Hermitian.
\end{lemma}

\begin{proof}
$\Chat(\rho)^{\dagger} = \sum_g \overline{C(g)}\rho(g)^{\dagger}
= \sum_g C(g)\rho(g^{-1}) = \sum_g C(g^{-1})\rho(g)
= \sum_g C(g)\rho(g) = \Chat(\rho)$.
\end{proof}

Accordingly each $\Chat(\rho)$ admits an orthonormal eigenbasis
$\{u^{(s)}_{\rho}\}_{s=1}^{d_\rho} \subset \C^{d_\rho}$ with real
eigenvalues,
\[
\Chat(\rho)\, u^{(s)}_{\rho} \;=\; \lambda^{(s)}_{\rho}\, u^{(s)}_{\rho},
\qquad \lambda^{(s)}_{\rho} \in \R .
\]

\subsection{Centering removes exactly the trivial representation}

\begin{lemma}[Centering]
\label{lem:centering}
The constant vector $\bm 1$ is an eigenvector of $\Mstar$ with eigenvalue
$\mu_0 = \sum_g C(g)$.  $\Mstar$ is real symmetric so we may construct an orthogonal eigenbasis for its eigenspace with $\bm 1$ explicitly included. $P\Mstar P$ (with $P = \id - \tfrac1n \bm1\bm1^{\top}$) has these same eigenpairs as $\Mstar$ except that the pair $(\mu_0, \bm1)$ is replaced by $(0, \bm1)$; no other
eigenvector or eigenvalue is altered.
\end{lemma}

\begin{proof}
$\Mstar\bm1 = \big(\sum_g C(g)\big)\bm1$ from equation \ref{eq:CG}, so $\bm 1$ is indeed an eigenvector.  Furthermore, if $\Mstar v = \lambda v$ with
$v \perp \bm1$ then $Pv = v$. Hence, $P\Mstar Pv = P\Mstar v = \lambda Pv = \lambda v$,
while $P\Mstar P\bm1 = 0$. 
\end{proof}

For obtaining our word embeddings from $P\Mstar P$, we can diagonalize $\Mstar$ and discard the constant mode. This constant mode corresponds to the trivial representation of the group, $\rho(g) = 1 \;\forall g\in G$.

\subsection{Embedding Theorem for finite groups}

\begin{theorem}[Diagonalization of invariant statistics; finite case]
\label{thm:finite-main}
Let $G$ be a finite group with $n$ elements.  Assume we have a $G$-torsor (Definition \ref{def:torsor}) word set $L$ whose $\Mstar$ (Equation \ref{eq:mstar-def}) is statistically invariant under $G$.  Under an arbitrary choice of base point (arbitrary from Lemma \ref{lem:base_point}), we associate with $\Mstar$ a kernel $C : G \rightarrow \mathbb{R}$ as in Lemma \ref{lem:kernel}.  From Lemma \ref{lem:hermitian}, $\widehat{C}(\rho)$ admits an orthonormal eigenbasis $\{u_{\rho}^s  \}_{s=1}^{d_{\rho}} \subset \mathbb{C}^d$ with real eigenvalues $\lambda_{\rho}^{(s)}$.  Let $\widehat{G}$ denote a complete set of irreducible unitary representations of $G$ as usual.
For each $\rho \in \Ghat$, each eigenvector index
$s \in \{1,\dots,d_\rho\}$ and each row index $a \in \{1,\dots,d_\rho\}$, we define
\begin{equation}
\label{eq:eigenfunctions}
\psi^{\rho,a,s}(g) \;:=\; \sqrt{\tfrac{d_\rho}{n}}\;
\big[\rho(g)\, u^{(s)}_{\rho}\big]_a .
\end{equation}
Then:
\begin{enumerate}[label=(\roman*)]
\item Each $\psi^{\rho,a,s}$ is an eigenvector of $\Mstar$ with eigenvalue
$\lambda^{(s)}_{\rho}$:
\(
\Mstar\,\psi^{\rho,a,s} = \lambda^{(s)}_{\rho}\, \psi^{\rho,a,s}.
\)  Furthermore, there exists a choice of equivalent irreducible unitary representations, $\rho^{\prime}$, such that we have eigenvectors $\sqrt{\frac{d_{\rho}}{n}}\rho'(g)_{as}$.
% \item (Corpus-adapted irreps.) Define $U_\rho := \big[u^{(1)}_\rho \,\big|\,
% \cdots \,\big|\, u^{(d_\rho)}_\rho\big] \in U(d_\rho)$ and the unitarily
% equivalent copies $\rho' := U_\rho^{\dagger}\, \rho\, U_\rho$. Then
% $\Chat(\rho') = U_\rho^{\dagger}\,\Chat(\rho)\,U_\rho =
% \mathrm{diag}\big(\lambda^{(1)}_\rho, \dots, \lambda^{(d_\rho)}_\rho\big)$,
% so in these copies the eigenvectors are the standard basis vectors and
% \eqref{eq:eigenfunctions} reduces to the bare matrix coefficients,
% \[
% \psi^{\rho,a,s}(g) \;=\; \sqrt{\tfrac{d_\rho}{n}}\;\rho'(g)_{as}.
% \]
% The two descriptions differ by a unitary mixing of the degeneracy index
% $a$ within each eigenspace, i.e., by the global rotation quotiented
% throughout.

\item The family
$\{\psi^{\rho,a,s} : \rho \in \Ghat,\ 1 \le a,s \le d_\rho\}$
is an orthonormal basis of $\ell^2(G) \cong \mathbb{C}^n$ ($\sum_\rho d_\rho^2 = n$ vectors).
\item Hence, the spectrum of $\Mstar$ is
$\{\lambda^{(s)}_\rho : \rho \in \Ghat, s \in [d_{\rho}]\}$, each occurring with multiplicity (at least)
$d_\rho$ and
\[
P \Mstar P \;=\; \sum_{\rho \neq \rho_{\mathrm{triv}}}\ \sum_{s=1}^{d_\rho}
\lambda^{(s)}_{\rho} \sum_{a=1}^{d_\rho}\,
\psi^{\rho,a,s}\,\big(\psi^{\rho,a,s}\big)^{\dagger}.
\]
where $P = I - \frac{1}{n}\bm{1}\bm{1}^T$.

\end{enumerate}
\end{theorem}

\begin{proof}
(i) Substituting $h = gk$,
\begin{align*}
(\Mstar\psi^{\rho,a,s})(g)
&= \sum_{h} C(g^{-1}h)\,\psi^{\rho,a,s}(h)
 = \sqrt{\tfrac{d_\rho}{n}} \sum_{k} C(k)\,\big[\rho(g)\rho(k)\,
   u^{(s)}_\rho\big]_a \\
&= \sqrt{\tfrac{d_\rho}{n}}\,
   \Big[\rho(g)\,\Chat(\rho)\, u^{(s)}_\rho\Big]_a
 = \lambda^{(s)}_{\rho}\, \psi^{\rho,a,s}(g).
\end{align*}
This proves the first claim.  Now, for some irrep $\rho$, let $U_{\rho} = \left[u_{\rho}^{(1)}|\cdots| u_{\rho}^{(d_{\rho})}\right]$.  Note that $U_{\rho}$ is unitary since $\Chat(\rho)$ is Hermitian.  Let us define the equivalent unitary irrep $\rho' := U_{\rho}^{\dagger}\,\rho \,U_{\rho}$.  Furthermore,  $\widehat{C}(\rho)$ = $U^{\dagger}\Lambda U$ where $\Lambda = \mathrm{diag}\left(\lambda_\rho^{(1)},\ldots, \lambda_{\rho}^{(d_{\rho})}\right).$  Hence, \[
\widehat{C}(\rho') = \sum_{g\in G}C(g)U_{\rho}^{\dagger}\,\rho \,U_{\rho} = U_{\rho}^{\dagger}\,\widehat{C(\rho)}\,U_{\rho} = \Lambda. 
\]
We can choose the basis vectors $\{e_i  \}_{i=1}^{d_{\rho}}$ as the eigenbasis for $\Chat(\rho')$. Plugging into our formula for the eigenvectors above, we obtain
\[
\sqrt{\frac{d_{\rho}}{n}}[\rho(g) e_{s}]_a = \sqrt{\frac{d_{\rho}}{n}} \rho(g)_{as}
\]
(ii) Orthonormality: expanding
$\psi^{\rho,a,s}(g) = \sqrt{d_\rho/n}\sum_b \rho_{ab}(g)\,
(u^{(s)}_\rho)_b$ and applying \eqref{eq:schur},
\[
\inner{\psi^{\rho,a,s}}{\psi^{\sigma,c,t}}
= \tfrac{\sqrt{d_\rho d_\sigma}}{n} \sum_{b,e} (u^{(s)}_\rho)_b
\overline{(u^{(t)}_\sigma)_e} \sum_g \rho_{ab}(g)\overline{\sigma_{ce}(g)}
= \delta_{\rho\sigma}\,\delta_{ac}\,
\inner{u^{(t)}_\rho}{u^{(s)}_\rho}_{\C^{d_\rho}}
= \delta_{\rho\sigma}\delta_{ac}\delta_{st}.
\]
Completeness follows by counting: there are $\sum_\rho d_\rho^2 = n$
orthonormal vectors in an $n$-dimensional space. (iii) is immediate from
(i)--(ii) and Lemma~\ref{lem:centering}.
\end{proof}

Two freedoms are apparent in Theorem \ref{thm:finite-main}: the choice of equivalent irreps, and the choice of eigenbasis of $\Chat(\rho)$.  We show the third potential source of freedom, choosing the base point, produces no permissible $\psi^{\rho,a,s}$ outside of the previous two freedoms in Lemma \ref{lem:base_point}.  Both of these freedoms do not amount to ``structural'' changes.  We spare the small derivations, but one can see the first freedom amounts to a unitary rotation of our eigenbasis in the $a$ coordinate; i.e., the basis vectors $\{\psi^{\rho,a,s} \}_{a=1}^{d_{\rho}}$ are unitarily rotated within their subspace.  Additionally, the choice of eigenbasis amounts to a unitary rotation across $s$ in $\psi^{\rho,a,s}$ (for fixed $\rho,a$) for which all $\psi^{\rho,a,s}$ posses the same eigenvalue.

\begin{lemma}[Choice of base point]
\label{lem:base_point}
Consider a $G$-torsor word set $L$.  We can fix a base point $x_0 \in L$ and identify $L \cong G$ via $g \mapsto g \cdot x_0$.  Under the alternative choice of base point $x_0' = b\cdot x_0$, the group elements assigned to the word $x$, $g_x$, becomes $g_x \,b^{-1}$ .
Firstly, this relabeling has no effect on $\Mstar$ nor its eigendecomposition (what we denote as $\Mstar$'s indices has no effect on its numerical values).  Hence, we can freely choose any base point without any effect on the word embeddings. 

What may change is our eigenvectors in equation \ref{eq:eigenfunctions}.  Fortunately, the proposed change of base point produces the same set of permissible $\psi^{\rho,a,s}$'s as in Theorem \ref{thm:finite-main}, with the usual freedoms of choosing equivalent irreps and an eigenbasis of $\Chat(\rho)$.

% where $a \in G$, we consider its modified kernel $C^{a}(\rho) = C(a^{-1}ka)$.  The generalized Fourier transform of $C^{a}(g)$ with respect to $G$ obeys, $\hat{C}^{a}(\rho) = \rho(a)\hat{C}(\rho)\rho(a)^{-1}$.
\end{lemma}

\begin{proof}
$\Mstar_{x,y} = C(g_x^{-1}g_y)$ where $g_x,g_y$ are the elements of $G$ assigned to words $x,y$ before the base point change.  Since the matrix elements do not change, after selecting the new base point, $g \rightarrow gb^{-1}$, and we have that 
\[
\Mstar_{x,y} = C'(g_x'^{-1}g'_y ) = C'(bg_x^{-1}g_y b^{-1}). \]
Hence, $C'(bkb^{-1}) = C(k)\;\; \forall k \in G$, so $C'(k) = C(b^{-1}kb)$.
where $C'$ is the new kernel.
We can see that
\[
\Chat'(\rho) = \rho(b)\Chat(\rho)\rho (b)^{-1}.
\]
Therefore, we can choose our eigenbasis of $\Chat'(\rho)$ to be $\{\rho(b) u_{\rho}^{(s)} \}_{s=1}^{d_\rho}$.  The formula for the a modified $\psi$ then becomes
\[
\sqrt{\frac{d_{\rho}}{n}} [\rho(g')\rho(b)u_{\rho }^{(s)}]_a = \sqrt{\frac{d_{\rho}}{n}} [\rho(g)u_{\rho }^{(s)}]_a,
\]
unchanged from before.  Furthermore, if we used our freedom of choice of equivalent irreps or eigenbasis of $\widehat{C}'(\rho)$ when constructing the modified $\psi$, those same freedoms are present in the original set of permissible $\psi$'s.  Hence, the sets of permissible $\psi$ before and after the base change are the same.
\end{proof}
Similar results to Lemma \ref{lem:base_point} can be shown in the cases that $G$ is replaced with a compact group (Appendix \ref{app:compact}), or homogeneous space (Appendix \ref{app:homog}).
% \begin{remark}[Structural reading: the commutant]
% \label{rem:commutant}
% Theorem~\ref{thm:finite-main} is the concrete form of an abstract fact
% worth keeping in mind. Assumption~\ref{ass:invariance} says exactly that
% $H$ commutes with the left regular representation
% $(L_a f)(g) = f(a^{-1}g)$:
% \[
% (H L_a f)(g) = \sum_h C(g^{-1}h) f(a^{-1}h)
% = \sum_k C\big((a^{-1}g)^{-1}k\big) f(k) = (L_a H f)(g).
% \]
% By Peter--Weyl, $\ell^2(G) \cong \bigoplus_\rho V_\rho \otimes V_\rho^*$
% with $L$ acting as $\rho \otimes \id$, so the commutant of $L$ is
% $\bigoplus_\rho \id_{V_\rho} \otimes \mathrm{End}(V_\rho^*)$: any invariant
% operator is a direct sum of matrices $\Chat(\rho)^{\top}$ acting on the
% multiplicity factors, each repeated $d_\rho$ times. This one-line argument
% explains all the multiplicities below and generalizes verbatim to compact
% groups and, in modified form (Theorem~\ref{thm:GK}), to homogeneous
% spaces.
% \end{remark}

\subsection{Special case: kernel is class function}
\label{app:class}
The natural analogue of the \citet{karkada2026symmetry} assumption that statistics
depend only on \emph{distance} in some cyclic latent space is that our kernel $C$, as in Lemma \ref{lem:kernel}, is a \emph{class function}. I.e., it is constant on conjugacy classes of some respective group $G$.  An explanation as to why follows.

Let's assume $\Mstar_{g,h} = D(g,h)\; \forall g,h \in G$ for some ``distance function'' $D: G \times G \rightarrow \mathbb{R}$.  Intuitively, distances between points shouldn't change as we move them together through space.  This suggests the condition that $D$ is bi-invariant under $G$. In other words, 
\[D(ag,ah) = D(ga,ha) = D(g,h)\; \forall a,g,h \in G.\]
  Hence, 
\[
C(g) = D(e,g) = D(a,ag) = D(e,aga^{-1}) = C(aga^{-1})\quad \forall a,g \in G,
\]
so $C$ being a class function is indeed the natural analogue.

% as follows.  Consider a finite group $G$.  Suppose
% $\Mstar_{xy} = D(x,y)$ for a bi-invariant function $D: G \times G \rightarrow \mathbb{R}$, i.e., 
% $D(ag, ah) = D(ga, ha) = D(g,h) \; \forall a,g,h \in G$. Then the kernel $C$ corresponding to $\Mstar$ obeys
% \[
% C(h) = D(e, h) = D(s, sh) = D(e, shs^{-1}) = C(shs^{-1}),
% \]
% so $C$ is a \emph{class function}: constant on conjugacy classes.

\begin{proposition}[Fourier transform of class function $C$ is scaled identity]
\label{prop:class}
If $C$ is a class function, then for every $\rho \in \Ghat$,
\[
\Chat(\rho) \;=\; \hat c_\rho\, \id_{d_\rho},
\qquad
\hat c_\rho \;=\; \frac{1}{d_\rho} \sum_{g \in G} C(g)\, \chi_\rho(g)
\;\in\; \R,
\]
where $\chi_\rho = \tr \rho$. Consequently in
Theorem~\ref{thm:finite-main} the 
eigenvalue $\hat c_\rho$ has multiplicity $d_\rho^{\,2}$, one for each matrix coefficient of $\rho$, and we can take
$\psi^{\rho,a,b}(g) = \sqrt{d_\rho/n}\;\rho(g)_{ab}$.
\end{proposition}

\begin{proof}
For any $a \in G$,
\[
\rho(a)^{-1}\,\Chat(\rho)\,\rho(a)
= \sum_{g \in G} C(g)\,\rho(a^{-1} g a)
= \sum_{g \in G} C(a g a^{-1})\,\rho(g)
= \sum_{g \in G} C(g)\,\rho(g) = \Chat(\rho),
\]
so $\Chat(\rho)$ commutes with $\rho$.  Hence, by Schur's lemma $\Chat(\rho)$
must be a scalar (which we will denote $\hat{c}_\rho$) times the identity matrix. 
Taking traces of $\Chat(\rho) = \hat{c}_\rho \id_{d_\rho}$ gives
$d_\rho \hat c_\rho = \sum_g C(g)\chi_\rho(g)$, the desired relation. 

% Reality of $\hat c_\rho$:
% $\overline{\chi_\rho(g)} = \chi_\rho(g^{-1})$, and the substitution
% $g \mapsto g^{-1}$ together with $C(g^{-1}) = C(g)$ shows the sum equals
% its own conjugate. 

The consequences for Theorem~\ref{thm:finite-main} follow as every unit vector of $\mathbb{C}^{d_\rho}$
is now an eigenvector of $\Chat(\rho)$.  Additionally, note that our eigenvalues of $\widehat{C}(\rho)$, $\hat{c}_\rho$, are still real as before since
\[
\overline{\hat{c}_\rho} = \frac{1}{d_{\rho}} \sum_{g\in G} \overline{C(g)}\overline{\chi_{\rho}(g)} = \frac{1}{d_{\rho}} \sum_{g \in G} C(g) \chi_{\rho}(g^{-1}) = \frac{1}{d_\rho}\sum_{g \in G} C(g^{-1})\chi_{\rho}(g) = \frac{1}{d}\sum_{g\in G} C(g) \chi_{\rho}(g) = \hat{c}_{\rho}
\]
\end{proof}

% \begin{remark}
% The hierarchy is: \emph{left-invariance alone} gives eigenvalue blocks of
% multiplicity $d_\rho$ (one per eigenvector of the $d_\rho \times d_\rho$
% Hermitian matrix $\Chat(\rho)$); \emph{bi-invariance} (class function)
% fuses these into single blocks of multiplicity $d_\rho^2$. Our chord data
% realize the first case (Appendix~\ref{app:dihedral}); the sphere realizes a
% multiplicity-free analogue of the second (Appendix~\ref{app:gelfand}).
% \end{remark}

\subsection{Realification of Eigenfunctions}
\label{app:realification_finite}
Embeddings are real, while the eigenvectors \eqref{eq:eigenfunctions} are in general complex.  $\Mstar$ is a real symmetric matrix, so it always possible to find a real orthonormal basis for its eigenspace.  The only question is whether it is somehow possible to modify our theory's outputted irrep-based eigenvectors in a formulaic manner as to obtain real embeddings.

The answer depends on the ``type'' of irrep we are considering.  Namely, we consider its Frobenius-Schur indicator 
\[
\nu(\rho) \;=\; \frac{1}{n}\sum_{g \in G} \chi_\rho(g^2) \;\in\;
\{1, 0, -1\},
\]
with $\chi_{\rho}(g^2)$ denoting the trace of $\rho(g^2)$.  $\nu = 1,0,-1$ are denoted as \emph{real} type, \emph{complex} type, and \emph{quaternionic} type, respectively.  

\begin{proposition}[Realification Protocols]
\label{prop:FS}
Transforming the eigenvectors \eqref{eq:eigenfunctions} into real form depends on the type (Frobenius-Schur indicator) of the given irrep.  The different methodologies are as follows.
\begin{enumerate}[label=(\alph*)]
\item \textbf{Real type} ($\nu(\rho) = 1$). This is the simplest case.  Since $\rho$ is real type, it is guaranteed to be expressible as real matrices upon some appropriate change of basis.  Furthermore, after enforcing the reality of $\rho$, the eigenvectors of $\Chat(\rho)$ ,$\{u_{\rho}^{(s)}\}_{s=1}^d$, can be chosen real as $\widehat{C}$ is now a real symmetric matrix. Together, these enforce the reality of the eigenvectors in \eqref{eq:eigenfunctions}. 
% Choose the basis of
% $V_\rho$ in which $\rho$ is real orthogonal. Then $\Chat(\rho)$ is a real
% symmetric matrix, $u^{(s)}_\rho$ may be chosen real, and the
% $\psi^{\rho,a,s}$ of \eqref{eq:eigenfunctions} are already real. Real
% multiplicity: $d_\rho$.
\item \textbf{Complex type} ($\nu(\rho) = 0$). Since $\rho$ is of complex type, no change of basis can convert it to real matrices. Hence, a different approach is required.  Namely, for irreps $\rho$ and $\bar{\rho}$, and $a,s \in [d_{\rho}]$, we generate new real orthonormal eigenvectors through
\begin{align}
\psi^{\rho,a,s}, \psi^{\bar{\rho},a,s} \rightarrow \sqrt{2}\,\mathrm{Re}\,\psi^{\rho,a,s},\sqrt{2}\,\mathrm{Im}\,\psi^{\rho,a,s}
\end{align}
\item \textbf{Quaternionic type} ($\nu(\rho) = -1$). Since $\rho$ is of quaternionic type, it is impossible to express it as real matrices.  Furthermore, unlike the complex case, there is generally no ``simple'' way to arrive at real eigenvectors which are clean functions of our eigenvectors in equation \eqref{eq:eigenfunctions}.  An approach instead is to take the real and imaginary components of each $\psi^{\rho,a,s}$ and just apply Gram-Schmidt to arrive at real, orthogonal eigenvectors.   

\end{enumerate}
\end{proposition}
\begin{proof}
(a) Proof already discussed in proposition.  

(b) Since $\rho$ is of complex type, it is well known that $\rho\not\cong \overline{\rho}$.  $\rho$ and $\overline{\rho}$ are genuinely nonequivalent irreps which correspond to different eigenvectors in equation \eqref{eq:eigenfunctions}.  The reality of our proposed substitution is immediate.  All that is left is to demonstrate that the proposed vectors are indeed eigenvectors of $\Mstar$ and are orthonormal.

We begin by proving that $\mathrm{Re}\,\psi^{\rho, a ,s}$ and $\mathrm{Im}\,\psi^{\rho, a ,s}$ are  eigenvectors with the same eigenvalue $\lambda_{\rho}^{(s)}$.  Namely, 
\[
\mathrm{Re}\,\psi^{\rho, a ,s} = \frac{1}{2}(\psi^{\rho,a,s} + \overline{\psi^{\rho,a,s}}),\quad\mathrm{Im}\,\psi^{\rho, a ,s} = \frac{1}{2i}(\psi^{\rho,a,s} - \overline{\psi^{\rho,a,s}})
\]
Note that 
\[
\overline{\psi^{\rho,a,s}} = \sqrt{\frac{d_{\rho}}{n}} \left[\overline{\rho}(g)\overline{u_{\rho}^{(s)}}\right]_a
.\]
$\overline{u_{\rho}^{(s)}}$ is an eigenvector of $\overline{\Chat}(\rho)$, which in turn equals $\Chat(\overline{\rho})$ from the reality of $C$.  Hence, 
\[
\overline{\psi^{\rho,a,s}} = \sqrt{\frac{d_{\rho}}{n}} \left[\overline{\rho}(g)u_{\overline{\rho}}^{(s)}\right]_a = \psi^{\overline{\rho},a,s}
,\]
so $\overline{\psi^{\rho,a,s}} = \psi^{\overline{\rho},a,s}$ is another eigenvector of $\Mstar$.  Furthermore, $\psi^{\overline{\rho},a,s}$ and $\psi^{\rho,a,s}$ share the same eigenvalue since
\[
\Mstar \overline{\psi^{\rho,a,s}} = \overline{\Mstar \psi^{\rho,a,s}} = \overline{\lambda_{\rho}^{(s)}}\overline{\psi^{\rho,a,s}} = \lambda_{\rho}^{(s)}\overline{\psi^{\rho,a,s}},
\]
taking advantage of $\Mstar$ being real symmetric.  Therefore, $\mathrm{Re}\,\psi^{\rho, a ,s}$ and $\mathrm{Im}\,\psi^{\rho, a ,s}$ are indeed eigenvectors with eigenvalue $\lambda_{\rho}^{(s)}$.  

Finally, checking the orthonormality of  $\sqrt{2}\,\mathrm{Re}\,\psi^{\rho,a,s}$ and $\sqrt{2}\,\mathrm{Im}\,\psi^{\rho,a,s}$ is relatively simple as $\psi^{\rho,a,s}$ and $\psi^{\overline{\rho},a,s} = \overline{\psi^{\rho,a,s}}$ are orthonormal with respect to any of the original eigenvectors.

(c) Gram-Schmidt returns real vectors if real vectors are first provided.  Furthermore, define the eigen-subspace $W_{\lambda} = \{\phi: \Mstar \phi = \lambda \phi \}$.  $W_{\lambda}$ and $W_{\lambda'}$ are orthogonal for $\lambda \neq \lambda'$ since $\Mstar$ is real symmetric. Hence, if Gram-Schmidt is provided $k$ vectors in $W_{\lambda}$ it will return $k$ vectors in $W_{\lambda}$, i.e., it will return other eigenvectors.

All that is left to show is that the real and imaginary components of the $\psi$'s are all eigenvectors and span the space of all eigenvectors of $\Mstar$. Firstly, note that since $\Mstar$ is real with real eigenvalues, the real and imaginary components of any $\psi^{\rho,as}$ are automatically eigenvectors.  Namely,
\[
\Mstar\psi^{\rho,a,s}  = \lambda^{(s)}_{\rho} \psi^{\rho,a,s} \Rightarrow \Mstar\, \mathrm{Re\,\psi^{\rho,a,s}} + i\Mstar\,\mathrm{Im}\,\psi^{\rho ,a,s} = \lambda_{\rho}^{(s)}\,\mathrm{Re}\,\psi^{\rho,a,s} + i\lambda_{\rho}^{(s)}\,\mathrm{Im}\,\psi^{\rho,a,s}
\]
splits into two separate eigenvalue-eigenvector equations. Finally, these real and imaginary components of the $\psi$ span the space of all eigenvectors automatically since the $\psi$ span the space of all eigenvectors.  Hence, Gram-Schmidt will not be missing any eigenvectors.

% We define $K: \mathbb{C}^d \rightarrow \mathbb{C}^d $ by $Kv = \overline{v}$ where $v \in \mathbb{C}^d$.  $K$

\end{proof}

Only quaternionic type irreps would provide real eigenvectors whose structure do not fall neatly from our equation \ref{eq:eigenfunctions}.  Fortunately, these types of irreps are arguably ``rarer'' in more typical groups, and do not appear in any of our experiments.

\subsection{Special cases: cyclic and dihedral groups}
\label{app:special_finite}
We briefly state the structure Theorem~\ref{thm:finite-main} predicts
for the cyclic ($\Z_{12}$) and dihedral ($D_{12}$) cases relevant to
our experiments.

\paragraph{Cyclic: $G = \Z_{12}$ (months).} All irreps are
one-dimensional: characters $\chi_k(m) = e^{2\pi i k m/12}$,
indexed $k = 0, \dots, 11$.  The generalized Fourier transform is actually the discrete Fourier transform.  Namely,  
$\Chat(\rho)$ become scalars
\[
\hat c_k = \sum_{d=0}^{11} C(d)\, e^{-2\pi i k d/12}
= \sum_d C(d) \cos(2\pi k d /12),
\] which are real since $C(d) = C(-d)$. After
realification, the conjugate pairs $(\chi_k, \chi_{-k})$,
$k = 1, \dots, 5$, fuse into two-dimensional cosine/sine planes sharing
the eigenvalue $\hat c_k$. $k = 0$ is the constant mode (removed by
centering) and $k = 6$ the one-dimensional alternating mode, which empirically tends to be neglected in natural data. The
predicted embedding of month $m$ is therefore, up to one orthogonal transform per
plane,
\[
m \;\longmapsto\;
\Big(\sqrt{\hat c_k}\,\cos\big(\tfrac{2\pi k m}{12}\big),\
\sqrt{\hat c_k}\,\sin\big(\tfrac{2\pi k m}{12}\big)\Big)_{k \in \mathcal{K}},
\qquad
\mathcal{K} = \{k \ge 1 : \hat c_k > 0,\ \text{decreasing } \hat c_k\}:
\]
concentric circles traversed at $30k^\circ$ per month, radius
$\sqrt{\hat c_k}$.

\paragraph{Dihedral: $G = D_{12}$ (major/minor triads).} Let $T_m$ and $I_m$ denote the $m^{th}$ major and minor triad respectively for $m \in \{0,\ldots, 11\}$ (with base chord C).  $D_{12} =
\Z_{12} \rtimes \Z_2$ has four one-dimensional irreps and five
two-dimensional irreps $E_h$, $h = 1, \dots, 5$, with
$E_h(T_m) = R(2\pi h m/12)$ and $E_h(I_m) = R(2\pi h m /12)\,F$.
$R(\theta)$ denotes typical 2D rotation matrix by $\theta$ and $F = \mathrm{diag}(1, -1)$. Note that all irreps are
of real type.  We will neglect the one-dimensional irreps as they do not tend to prevalent, as least for our experiments. For each $E_h$, $\Chat(E_h)$ is a real
symmetric $2 \times 2$ matrix with eigenpairs
$(\lambda_h^{(s)}, u_h^{(s)})$, $s = 1, 2$. Writing
$u^{(s)}_h = (\cos\varphi, \sin\varphi)$, each realized mode
$(h, s)$ with $\lambda^{(s)}_h > 0$ contributes a plane in which (up to one orthogonal transform per plane)
\[
T_m \rightarrow  \sqrt{\lambda_h^{(s)}}
\big(\cos\left(\tfrac{2\pi h m}{12} + \varphi\right),\,
\sin\left(\tfrac{2\pi h m}{12} + \varphi\right)\big),
\quad
I_m \rightarrow \sqrt{\lambda_h^{(s)}}
\big(\cos\left(\tfrac{2\pi h m}{12} - \varphi\right),\,
\sin\left(\tfrac{2\pi h m}{12} - \varphi\right)\big),
\]
two interleaved twelve-point circles of \emph{equal radius} (forced by
orthogonality of $E_h$), stepping $30h^\circ$ per semitone, with relative offset $2\varphi$.

\section{Compact groups}
\label{app:compact}

The finite theory extends to compact groups with only minor adjustments: sums become (Haar) integrals, matrices become integral
operators, and the spectral theorem for compact self-adjoint operators
supplies the discrete spectrum that makes ``top eigenvalues'' (and hence
low-rank embeddings) meaningful.

One practical detail: in this section, we are intrinsically assuming that we have a bijection between a continuous group $G$ and some set of words.  In reality, one has a finite set of words which sample the continuous group.  Hence, for theorem \ref{thm:compact-main}, one forms predicted word embeddings by taking the eigenfunctions and sampling them at the group elements corresponding to the words, forming vectors. 

\subsection{Harmonic Analysis Preliminaries}

We let $G$ be a compact Hausdorff topological group with normalized Haar
measure $dg$ ($\int_G dg = 1$), and work in
the Hilbert space $L^2(G)$ with $\inner{f_1}{f_2} = \int_G f_1 \overline{f_2}\, dg$. Let
$\Ghat$ be the (now countably infinite, in general) set of equivalence
classes of unitary irreps $\rho: G \to U(d_\rho)$. The Peter--Weyl theorem
states that the matrix coefficients
$\{\sqrt{d_\rho}\,\rho_{ab} : \rho \in \Ghat,\ 1 \le a,b \le d_\rho\}$
form a complete orthonormal basis of $L^2(G)$, with orthogonality
\begin{equation}
\label{eq:schur-cont}
\int_G \rho_{ab}(g)\,\overline{\sigma_{ce}(g)}\, dg
= \frac{1}{d_\rho}\,\delta_{\rho\sigma}\,\delta_{ac}\,\delta_{be}.
\end{equation}

The generalized Fourier transform of some $C \in L^2 (G)$ now takes the form
\[
\Chat(\rho) \;=\; \int_G C(g)\,\rho(g)\, dg \;\in\;
\C^{d_\rho \times d_\rho}.
\]

\subsection{Integral operator and the discrete spectrum}
\label{app:integ_op}

We naturally extend the notion of the previous real symmetric matrix  $\Mstar$ to an integral operator $\mathcal{M}$.  Namely, we first define the integral operator $\mathcal{M}$ on some $f \in L^2(G)$ by
\[
(\mathcal{M}f)(g) \;:=\; \int_G k(g,h)\, f(h)\, dh\
\]
where $k \in L^2(G \times G)$

The reality and symmetry of $\Mstar$ translate to requiring that $k$ is both real and symmetric in its inputs.  Finally, the statistical invariance of $\Mstar$ with respect to some finite group $G$ becomes the property that $\mathcal{M}$ commutes with left translation $L_b$ $\forall b \in G$, where $(L_b f)(g) = f(b^{-1}g)$.  I.e., $\mathcal{M}L_b = L_b\mathcal{M}$ $\forall b \in G$.

From these properties, $\mathcal{M}$ takes the form of
\begin{align}
\label{eq:M_compact}
(\mathcal{M}f)(g) \;:=\; \int_G C(g^{-1}h)\, f(h)\, dh
\end{align}
for some real kernel $C \in L^2(G)$ with
$C(g^{-1}) = C(g)$.

\begin{lemma}[Hilbert--Schmidt]
\label{lem:HS}
For $C \in L^2(G)$, $\mathcal{M}$ is a Hilbert--Schmidt (hence compact)
operator, with $\norm{\mathcal{M}}_{\mathrm{HS}} = \norm{C}_{L^2}$.
\end{lemma}

\begin{proof}
By left invariance of
Haar measure,
\[
\norm{\mathcal{M}}_{\mathrm{HS}}^2
= \int_G \int_G |C(g^{-1}h)|^2\, dh\, dg
= \int_G \Big(\int_G |C(u)|^2 du\Big) dg = \norm{C}_{L^2}^2 < \infty.
\]
\end{proof}

\begin{lemma}[Self-adjointness]
\label{lem:SA}
If $C$ is real with $C(g^{-1}) = C(g)$, then
$\mathcal{M} = \mathcal{M}^{*}$.
\end{lemma}

\begin{proof}
$\overline{k(g,h)} = \overline{C(h^{-1}g)}
= \overline{C\big((g^{-1}h)^{-1}\big)} = C(g^{-1}h) = k(g,h)$.
\end{proof}
The properties of $\mathcal{M}$ being Hilbert-Schmidt and self-adjoint allow us to sensibly discuss its top eigenvalues as the well known result in Corollary \ref{cor:spectral} describes.
\begin{corollary}[Discrete real spectrum]
\label{cor:spectral}
Let $G$ be a compact group with normalized Haar measure, and let
$C \in L^2(G)$ be real with $C(g^{-1})=C(g)$. Then the
operator $(\mathcal{M}f)(x) = \int_G C(x^{-1}y)\,f(y)\,dy$ on $L^2(G)$
is Hilbert--Schmidt (Lemma \ref{lem:HS}) and self-adjoint (Lemma \ref{lem:SA}). Consequently: $L^2(G)$ admits an
orthonormal basis of eigenfunctions of $\mathcal{M}$. The eigenvalues
$\{\mu_k\}$ are real with $\sum_k \mu_k^2 = \norm{C}_{L^2}^2 < \infty$ and
every nonzero eigenvalue has finite multiplicity and $\mu_k \to 0$. 
% In
% particular the eigenvalues may be ordered by decreasing $|\mu_k|$, the
% largest is attained, and low-rank truncations converge to $\mathcal{M}$
% in Hilbert--Schmidt norm.
\end{corollary}

\begin{lemma}[Centering]
\label{lem:centering-cont}
The constant function $\bm 1$ (already unit-norm) is an eigenfunction:
$\mathcal{M}\bm1 = \mu_0 \bm 1$ with $\mu_0 = \int_G C$. With
$P = \id - \Pi_0$, where $(\Pi_0 f)(g) = \int_G f\,dh$ is the projection
onto constants, $P\mathcal{M}P$ retains every eigenpair of $\mathcal{M}$
except that $(\mu_0, \bm1)$ becomes $(0, \bm1)$.
\end{lemma}

\begin{proof}
Identical to Lemma~\ref{lem:centering}, with sums replaced by integrals.
\end{proof}

\subsection{Embedding Theorem for Compact Groups}

Lemma~\ref{lem:hermitian} itself holds verbatim.  Only the proof changes slightly as the sums become Haar integrals and we utilize the inversion invariance of the Haar measure.  Hence, $\Chat(\rho)$ is Hermitian, giving
orthonormal eigenpairs
$\Chat(\rho)u^{(s)}_\rho = \lambda^{(s)}_\rho u^{(s)}_\rho$,
$\lambda^{(s)}_\rho \in \R$.

Our setup allows us to state a theorem that is largely equivalent to the finite case.

\begin{theorem}[Diagonalization; compact case]
\label{thm:compact-main}
Let $G$ be a compact group with normalized Haar measure $dg$.  Let integral operator $\mathcal{M}$ take the form as in Equation \ref{eq:M_compact}, so $\mathcal{M}$ has a discrete real spectrum from Corollary \ref{cor:spectral}.  Let the kernel of $\mathcal{M}$ be $C : G \rightarrow \mathbb{R}$.  Then $\widehat{C}(\rho)$ admits an orthonormal eigenbasis $\{u_{\rho}^{(s)}  \}_{s=1}^{d_{\rho}} \subset \mathbb{C}^d$ with real eigenvalues $\lambda_{\rho}^{(s)}$ as mentioned above.

Furthermore, define functions
\[
\psi^{\rho,a,s}(g) \;=\; \sqrt{d_\rho}\;\big[\rho(g)\,
u^{(s)}_\rho\big]_a,
\qquad \rho \in \Ghat,\quad 1 \le a, s \le d_\rho,
\]
Then: 
\begin{enumerate}[label=(\roman*)]

\item Each $\psi^{\rho,a,s}$ is an eigenfunction of $\mathcal{M}$ with eigenvalue $\lambda_{\rho}^{(s)}$.
\item The family
$\{\psi^{\rho,a,s} : \rho \in \Ghat,\ 1 \le a,s \le d_\rho\}$
is an orthonormal basis of $L^2(G)$.
\item Hence, the spectrum of $\mathcal{M}$ is
$\{\lambda^{(s)}_\rho: \rho \in \Ghat, s \in [d_{\rho}]\} \cup \{0\}$, each nonzero eigenvalue occurring
with finite multiplicity, at least $d_\rho$, and
\[
P \mathcal{M} P \;=\; \sum_{\rho \neq \rho_{\mathrm{triv}}}\ \sum_{s=1}^{d_\rho}
\lambda^{(s)}_{\rho} \sum_{a=1}^{d_\rho}\,
\psi^{\rho,a,s}\,\inner{\psi^{\rho,a,s}}{\cdot\,},
\]
where $P$ is defined in \ref{lem:centering-cont}.

% The family $\{\psi^{\rho,as,}:\rho \in \hat{G},  1\ \leq a,d\leq d_{\rho} \}$
% form a complete orthonormal system of eigenfunctions of $\mathcal{M}$,
% with $\mathcal{M}\psi^{\rho,a,s} = \lambda^{(s)}_\rho\, \psi^{\rho,a,s}$;
% each eigenvalue $\lambda^{(s)}_\rho$ carries multiplicity $d_\rho$ in the
% index $a$. If $C$ is a class function, then
% $\Chat(\rho) = \hat c_\rho \id$ with
% $\hat c_\rho = \tfrac{1}{d_\rho}\int_G C\,\chi_\rho\,dg \in \R$, and the
% multiplicity becomes $d_\rho^2$, carried by all matrix coefficients of
% $\rho$.

\end{enumerate}
\end{theorem}

\begin{proof}
We spare writing every detail as the proof is largely a copy of the finite case.
(i) The eigenfunction computation is the integral version of
Theorem~\ref{thm:finite-main}(i). 
(ii) Orthonormality comes from \eqref{eq:schur-cont} while 
completeness results from Peter--Weyl.  (iii) follows immediately from (i) and (ii).
\end{proof}

\begin{remark}[Kernel is class function, compact case]
\label{rem:class-compact}
The discussion in Appendix \ref{app:class} is largely the same in the compact case.  Eigenvalues still take on multiplicities of $d_{\rho}^2$, and we can express our eigenfunctions as $\psi^{\rho,a,s} = \sqrt{d_{\rho}}\,\rho(g)_{a,b}$.
\end{remark}

\begin{remark}[Realification, compact case]
\label{rem:FS-compact}
The discussion in Appendix~\ref{app:realification_finite} is largely analogous. The Frobenius-Schur indicator becomes
$\nu(\rho) = \int_G \chi_\rho(g^2)\, dg \in \{1, 0, -1\}$, and
Proposition~\ref{prop:FS} holds nearly verbatim.  (Note that in the quaternionic case, Gram-Schmidt can be applied to functions as it works on inner product spaces in general, not just vectors)
\end{remark}

% \begin{remark}[Smoothness and spectral decay]
% \label{rem:decay}
% Square-summability (Corollary~\ref{cor:spectral}) is the minimum; if $C$
% is smooth, the $\hat c_\rho$ decay rapidly (the compact-group analogue of
% the decay of Fourier coefficients), so low-rank truncations of
% $\mathcal{M}$ are themselves smooth kernels. Conversely, truncating the
% spectral expansion at the top blocks is a \emph{low-pass filter} on $C$.
% This is the theoretical face of an empirical observation in the sky
% experiments: a rank-limited representation cannot carry high-degree kernel
% content, so comparing kernels after matching bandwidths is the correct
% like-for-like comparison.
% \end{remark}

\section{Homogeneous spaces, Gelfand pairs, and the sphere}
\label{app:homog}

Our final generalization considers words in bijection with a homogeneous space of $G$.  As mentioned at the start of Appendix \ref{app:compact}, assuming we have a word set in bijection with a continuous set is not realistic.  In practice, we have a finite word set which samples a homogeneous space. We in turn sample our predicted eigenfunctions in Theorem \ref{thm:main_homog_space} at these points when making actual predictions.  

\subsection{Setup}

Let $K \le G$ be a closed subgroup of the compact group $G$, and
consider $X = G/K$, a homogeneous space  with $G$-invariant 
probability measure $dx$. Words are
identified with points of $X$. 

As in Appendix~\ref{app:finite}, we consider an operator $\mathcal{M}$ of the form
\[
(\mathcal{M}f)(g) \;:=\; \int_G C(g^{-1}h)\, f(h)\, dh\
\]
where $C$ is real, $C(g) = C(g^{-1})$, and $f \in L^2(G)$.  However, we now require that $C$ is bi-K-invariant: $C(k_1gk_2) = C(g)$ for $k_1,k_2 \in K$ (so its value is independent of the particular representative of some coset $gK$).  $\mathcal{M}$ descends to operating on $f \in L^2(X)$ through
\begin{align}
\label{eq:M_homog_space}
(\mathcal{M}f)(y):= \int_X \kappa(y,x)f(x)dx
\end{align}
where $\kappa \in L^2(X \times X)$ is real symmetric and $G$-invariant ($\kappa(gx,gy) = \kappa (x,y)\; \forall g \in G$).  We could equivalently replace $\kappa$ with a kernel $C$ defined by $C(g) = \kappa(eK, gK) \; \forall g \in G$.    Note that $C$ is immediately bi-K-invariant, and has the usual properties of being real and obeying $C(g) = C(g^{-1})\; \forall g \in G$.  This gives us the alternative form of $\mathcal{M}$
\begin{align}
\label{eq:M_homog_lift}
(\mathcal{M}f)(gK) \;=\; \int_G C(g^{-1}h)\, f(hK)\, dh,
\end{align}
expressed instead as an integral over the group.

% and
% Assumption~\ref{ass:invariance} becomes
% \[
% \Mstar(g\cdot x,\; g\cdot y) = \Mstar(x, y)
% \qquad \forall g \in G,\ x, y \in X.
% \]
% Writing $x = g_1 K$, $y = g_2 K$, invariance is equivalent to
% $\Mstar(x,y) = C(g_1^{-1} g_2)$ for a function $C$ on $G$ that is
% \emph{bi-$K$-invariant}: $C(k_1 g k_2) = C(g)$ for $k_1, k_2 \in K$
% (so that the value is independent of coset representatives).  Symmetry 
% again forces $C(g^{-1}) = C(g)$.  The operator is
% \[
% (\mathcal{M}f)(x) = \int_X \Mstar(x,y)\, f(y)\, dy,
% \qquad f \in L^2(X).
% \]

Let $\Ghat$ be the set of equivalence classes of unitary irreps of $G$, $\rho: G \rightarrow U(V_{\rho})$.  Recall that
\begin{equation}
\label{eq:LGK}
L^2(G/K) \;\cong\; \bigoplus_{\rho \in \Ghat} V_\rho^{\oplus m_\rho},
\qquad
m_\rho \;=\; \dim V_\rho^{K},
\end{equation}
where $V_\rho^K = \{v \in V_\rho : \rho(k)v = v\ \forall k \in K\}$ is the
subspace of $K$-fixed vectors. An irrep contributes exactly as many copies of $V_{\rho}$
as it has $K$-fixed directions.

\subsection{Embedding Theorem for Homogeneous Spaces}

\begin{theorem}[Diagonalization; homogeneous space]
\label{thm:main_homog_space}
Let $K \leq G$ be a closed subgroup of compact group $G$ and let $X = G/K$.
Assume we have an integral operator $\mathcal{M}$ as in Equation \ref{eq:M_homog_lift} with kernel $C: G \rightarrow \mathbb{R}$.  Let $\Ghat$ be the set of equivalence classes of unitary irreps of $G$, $\rho: G \rightarrow U(V_{\rho})$.  We define
\[
B_{\rho}:= P_K \hat{C}(\rho)P_K
\]
where $P_K = \int_K \rho(k)dk$ is the orthogonal projection of $V_{\rho}$ onto $V_{\rho}^K$.  It can be seen that $B_{\rho}$ is Hermitian and preserves $V_{\rho}^K$, so its restriction to $V_{\rho}^K$ admits an orthonormal eigenbasis $\{u_{\rho}^s  \}_{s=1}^{m_{\rho}} \subset V_{\rho}^k$ with real eigenvalues $\lambda_{\rho}^{(s)}$.  
Furthermore, define functions
\[
\psi^{\rho,a,s}(gK) \;=\; \sqrt{d_\rho}\;\big[\rho(g)\,
u^{(s)}_\rho\big]_a,
\qquad \rho \in \Ghat,\quad 1 \le a\le d_\rho,\ 1 \leq s \leq m_\rho
\]
where $m_\rho = \textrm{dim}\,V_{\rho}^K$.
Then: 
\begin{enumerate}[label=(\roman*)]
\item $\psi^{\rho,a,s}$ are well-defined on $X$.
\item Each $\psi^{\rho,a,s}$ is an eigenfunction of $\mathcal{M}$ with eigenvalue $\lambda_{\rho}^{(s)}$.
\item The family
$\{\psi^{\rho,a,s} : \rho \in \Ghat,\ 1 \le a \leq d_\rho,\, 1 \leq s\leq m_{\rho}\}$
is an orthonormal basis of $L^2(X)$.
\item Hence, the spectrum of $\mathcal{M}$ on $L^2(X)$ is
$\{\lambda^{(s)}_\rho : \rho \in \Ghat,\ s \in [m_\rho]\} \cup \{0\}$; each nonzero
eigenvalue occurs with finite multiplicity, at least $d_\rho$, and
\[
P \mathcal{M} P \;=\; \sum_{\substack{\rho \neq \rho_{\mathrm{triv}} \\ m_{\rho} \neq 0}}\ \sum_{s=1}^{m_\rho}
\lambda^{(s)}_{\rho} \sum_{a=1}^{d_\rho}\,
\psi^{\rho,a,s}\,\inner{\psi^{\rho,a,s}}{\cdot\,},
\]
where $P$ is defined in \ref{lem:centering-cont}.
\end{enumerate}

\end{theorem}

\begin{proof}

(i) For
any $k \in K$,
\[
\psi^{\rho,a,s}(gkK)
= \sqrt{d_\rho}\,\big[\rho(g)\,\rho(k)\, u^{(s)}_{\rho}\big]_a
= \sqrt{d_\rho}\,\big[\rho(g)\, u^{(s)}_{\rho}\big]_a
= \psi^{\rho,a,s}(gK),
\]
using $u^{(s)}_{\rho} \in V_{\rho}^{K}$. 

(ii) The computation of
Theorem~\ref{thm:finite-main} applies in a similar fashion: substituting $h = gk$,
\[
(\mathcal{M}\psi^{\rho,a,s})(gK)
= \int_G C(g^{-1}h)\,\psi^{\rho,a,s}(hK)\, dh
= \sqrt{d_\rho}\,\Big[\rho(g)\,\Chat(\rho)\,u^{(s)}_{\rho}\Big]_a .
\]
Since $u^{(s)}_{\rho} \in V_{\rho}^{K}$ and
$\Chat(\rho)$ agrees with $B_{\rho}$ on $V_{\rho}^{K}$
(as $P_K u^{(s)}_\rho = u^{(s)}_\rho$ and
$P_K \Chat(\rho) = \Chat(\rho)$ by bi-$K$-invariance),
\[
\Chat(\rho)\, u^{(s)}_{\rho} = B_{\rho}\, u^{(s)}_{\rho}
= \lambda^{(s)}_{\rho}\, u^{(s)}_{\rho},
\]
hence $\mathcal{M}\psi^{\rho,a,s} = \lambda^{(s)}_{\rho}\,
\psi^{\rho,a,s}$.

(iii) Orthonormality: under the identification of $L^2(X)$ with the
right-$K$-invariant functions in $L^2(G)$, the inner
products may be computed on $G$, where Schur orthogonality gives
\[
\inner{\psi^{\rho,a,s}}{\psi^{\sigma,c,t}}_{L^2(X)}
= \inner{\psi^{\rho,a,s}}{\psi^{\sigma,c,t}}_{L^2(G)}
= \delta_{\rho\sigma}\,\delta_{ac}\,
\inner{u^{(t)}_{\rho}}{u^{(s)}_{\rho}}_{V_\rho}
= \delta_{\rho\sigma}\,\delta_{ac}\,\delta_{st}.
\]
Completeness: by \eqref{eq:LGK}, the $\rho$-isotypic component of
$L^2(X)$ has dimension $d_\rho\, m_\rho$. The functions
$\{\psi^{\rho,a,s} : 1 \le a \le d_\rho,\ 1 \le s \le m_\rho\}$ are
$d_\rho m_\rho$ orthonormal elements of this component and therefore span it.
Taking the union over $\rho \in \Ghat$ gives completeness.

(iv) This result follows directly from (i)-(iii).  We note that the trivial irrep  is dropped due to centering as usual.
\end{proof}

Note that our discussion of realification as in Appendix~\ref{app:realification_finite} applies largely unchanged for homogeneous spaces, just as for the compact group case.

\subsection{Gelfand pairs}
\label{app:gelfand}

\begin{definition}
$(G, K)$ for compact $G$ is a \emph{Gelfand pair} if $m_\rho = \dim V_\rho^K \le 1$ for every
$\rho \in \Ghat$, i.e.\ the decomposition \eqref{eq:LGK} is
\emph{multiplicity-free}.
\end{definition}

% A convenient sufficient criterion (Gelfand's trick): if there exists an
% automorphism $\theta$ of $G$ with $\theta(g) \in K g^{-1} K$ for all $g$,
% then $(G,K)$ is Gelfand. For $(SO(3), SO(2))$ one may take conjugation by
% a rotation through $\pi$ about an axis orthogonal to the $SO(2)$-axis,
% which realizes $g \mapsto g^{-1}$ on double cosets; more simply, the
% double cosets are parameterized by the polar angle
% $\theta \in [0, \pi]$ (bi-invariant functions are functions of angular
% distance), and convolution of such ``zonal'' kernels is manifestly
% commutative.

\begin{corollary}[Diagonalization; Gelfand pair]
\label{cor:gelfand}
Assume $(G,K)$ is a Gelfand pair with $m_p = \dim V_\rho^K \le 1$ for every
$\rho \in \Ghat$.  Theorem \ref{thm:main_homog_space} can be simplified. For each $\rho$ with $m_\rho = 1$, fix a unit
$u_\rho \in V_\rho^K$. Then $B_\rho$ becomes the $1 \times 1$ scalar
\[
\lambda_\rho \;=\; \int_G C(g)\,\varphi_\rho(g)\, dg,
\qquad
\varphi_\rho(g) := \inner{\rho(g)\,u_\rho}{u_\rho}.
\]
The spectrum of $\mathcal{M}$ is the scalar sequence
$\{\lambda_\rho\}$, each with eigenspace of dimension at leaset $d_\rho$,
spanned by $\psi^{\rho,a}(gK) = \sqrt{d_\rho}\,[\rho(g)u_\rho]_a$.  The functions $\varphi_{\rho}$ are often denoted as spherical functions.
\end{corollary}

\begin{proof}
Since $\dim V_\rho^K = 1$, the restriction of $B_\rho$ to
$V_\rho^K = \C\,u_\rho$ is multiplication by the scalar
$\lambda_\rho = \inner{B_\rho u_\rho}{u_\rho}$. As $P_K u_\rho = u_\rho$
and $P_K$ is self-adjoint,
$\inner{B_\rho u_\rho}{u_\rho} = \inner{\Chat(\rho)\, u_\rho}{u_\rho}$.  Hence, 
\[
\lambda_{\rho} = \int_G C(g) \langle \rho (g)u_{\rho},u_{\rho}\rangle dg = \int_G C(g) \varphi_{\rho}(g)dg
\]
% Noting that $\varphi_{\rho}(g^{-1}) = \overline{\varphi_{\rho}(g)}$ gives the desired form.

% Inserting $\Chat(\rho) = \int_G C(g)\,\rho(g)\,dg$ and moving the
% integral outside the inner product,
% \[
% \lambda_\rho
% = \int_G C(g)\, \inner{\rho(g)\,u_\rho}{u_\rho}\, dg
% = \int_G C(g)\, \varphi_\rho(g)\, dg .
% \]
% Finally, $\varphi_\rho(g^{-1}) = \overline{\varphi_\rho(g)}$ by
% unitarity of $\rho$, so the substitution $g \mapsto g^{-1}$ together
% with $C(g^{-1}) = C(g)$ and inversion-invariance of Haar measure gives
% $\int_G C\,\varphi_\rho = \int_G C\,\overline{\varphi_\rho}$; in
% particular $\lambda_\rho \in \R$, consistent with the Hermiticity of
% $B_\rho$.
\end{proof}

% \begin{corollary}[Diagonalization; Gelfand pair]
% \label{cor:gelfand}
% If $(G,K)$ is a Gelfand pair, every block $B_\rho$ of
% Theorem~\ref{thm:GK} is a $1 \times 1$ matrix: a single real scalar
% $\lambda_\rho$ per participating irrep, with eigenspace of dimension
% exactly $d_\rho$, spanned by
% $\psi^{\rho, a}(gK) = \sqrt{d_\rho}[\rho(g) u_\rho]_a$ where $u_\rho$
% spans $V_\rho^K$. The eigenvalue is the spherical transform of $C$,
% \[
% \lambda_\rho \;=\; \int_G C(g)\,\varphi_\rho(g)\, dg
% \quad\text{(suitably normalized)},
% \qquad
% \varphi_\rho(g) := \inner{\rho(g)u_\rho}{u_\rho},
% \]
% with $\varphi_\rho$ the \emph{zonal spherical function} of $\rho$: the
% unique bi-$K$-invariant matrix coefficient with $\varphi_\rho(e) = 1$.
% In particular the entire spectrum of the statistics is a single scalar
% sequence indexed by $\Ghat_K := \{\rho : m_\rho = 1\}$, the eigenspaces
% are canonical (no $s$-index, no block-internal eigenvector fitting), and
% the only gauge left is the internal $O(d_\rho)$ (unitary/orthogonal)
% rotation of each $d_\rho$-dimensional eigenspace. This is the precise
% sense in which Gelfand pairs are the ``maximally rigid'' setting of the
% theory: the analogue of the class-function refinement is automatic, being
% forced by the geometry of $(G,K)$ rather than assumed of the statistics.
% \end{corollary}

\subsection{Special Case: The sphere}
\label{app:sphere}
We consider the particular case of the sphere which is relevant to our experiments.
Take $G = SO(3)$, $K = SO(2)$, $X = G/K = S^2$.  We state 3 well known facts about the sphere \citep{dai2013approximation}.

(1) The irreps of $SO(3)$: $\rho_{\ell}$ $\ell \geq0$, are real and have dimension 2$\ell +1$, each with $\dim\, V_{\rho_{\ell}}^K = 1$.  Hence, $(SO(3),SO(2))$ is a Gelfand pair.  

(2)  $L^2$ functions on the sphere have the isotypic decomposition $L^2(S^2) = \bigoplus_{\ell} \mathcal{H}_\ell$, where $\mathcal{H}_\ell$ are the degree-$\ell$
spherical harmonics.  Alternatively, for each $\ell \ge 0$, let
\[
\Psi_\ell \;:=\; \operatorname{span}\big\{\psi^{\rho_\ell, a} \;:\; 1 \le a \le 2\ell + 1\big\}
\;\subset\; L^2(S^2)
\]
span of the modes of Corollary~\ref{cor:gelfand} for the irrep
$\rho_\ell$.  We must have that $\mathcal{H}_{\ell} = \Psi_\ell$ as they correspond to the same isotypic component.  Hence, instead of using $\psi^{\rho,a}(gK)$ as our orthonormal basis, we can instead use the real degree $\ell$ spherical harmonics $\{Y_{\ell}^m\}_{\ell \geq 0, |m| \leq \ell }$.

(3) The spherical functions in Corollary~\ref{cor:gelfand} become Legendre polynomials, $\varphi_{\rho_{\ell}}(g) = P_{\ell}(\cos \theta_{g})$ where $\cos \theta_g = \langle g x_0 , x_0\rangle$, $x_0 = eK \in S^2$.  The inner product is done with respect to the points on the sphere realized in $R^3$.

Finally, note that the kernel $\kappa(x,y)$ on $S^2$ is invariant under $G = SO(3)$ iff $\kappa(x,y) = c(\langle x, y\rangle )$, i.e., the kernel is only some function $c$ of the inner product between two points.  Therefore, recall that our alternative form of the kernel $C$ obeys $C(g) = \kappa(eK, gK)$.  Hence $C(g) = c(\langle x_0, gK \rangle) = c(\cos \theta _g)$.  Altogether, applying Corollary~\ref{cor:gelfand} gives us that \[
\mathcal{M}Y = \lambda_{\ell}Y\quad \forall Y \in \mathcal{H}_{\ell}
\]
where $\lambda_{\ell}$ is given by
\begin{align*}
&\lambda_{\ell} = \int_G C(g) \varphi_{\rho}(g)dg = \int_G c(\cos\theta_g)P_{\ell}(\cos\theta_g)dg = \\
&\frac{1}{4\pi}\int_{0}^{2\pi} \int_{0}^{\pi} c(\cos \theta) P_{\ell}(\cos \theta) \sin \theta\, d\theta\,d\phi = \frac{1}{2}\int_{-1}^{1} c(t)P_{\ell}(t) dt
\end{align*}
and each $\lambda_{\ell}$ has multiplicity as least $2\ell +1$.  (Centering, i.e. considering $P\mathcal{M}P$, will set $\lambda_{0} = 0$)

\section{Months: experimental details}
\label{app:months-details}

\paragraph{Corpus.} \texttt{text8} \citep{mahoney2011text8}:
the first $10^8$ bytes of cleaned English Wikipedia ($17{,}005{,}207$
whitespace tokens; lowercase, punctuation stripped, digits spelled out).
Counts of $2{,}771$--$3{,}650$ per month after filtering ``may'' (see below).  Using this corpus, we form the normalized co-occurrence matrix $\Mstar \in \mathbb{R}^{12 \times 12}$ as in equation \ref{eq:mstar-def} (settings are $B = 16$, $f(\delta) = B{+}1{-}\delta$).  We
factorize  $\Mstar$ to form our empirical word embeddings (centering removes the constant mode).  Visually, The leading PCs are circular: PCs $1$--$2$ are the $k{=}1$ circle ($\lambda = 2.17, 1.59$), while PCs $3$--$4$ are $k{=}2$ circle
($0.88, 0.76$). 

\paragraph{Sense disambiguation for ``may''.} Raw counts:
$19{,}115$ for ``may'', ${\sim}6\times$ the other months, dominated by
the verb, which inflates the unigram $P_x$ in
Eq.~\ref{eq:mstar}. Fortunately, we can roughly select cases where ``may'' is used as a month.  Namely, we keep an occurrence of ``may'' iff its
two following tokens are both digits, or its preceding token is a
digit, or an adjacent token is another month. This retains
$3{,}129$ occurrences ($16.4\%$), in line with the other months. This
improves our per-month cosine prediction from $0.90$ to $0.96$.
% (per-month cosine $0.91 \to 0.96$; $k{=}2$ median angular step
% $51^\circ \to 66^\circ$ vs.\ theory $60^\circ$) while leaving the
% $k{=}1$ step at $31^\circ$; before filtering, ``may'' appears with
% correct angle but inflated radius, as predicted for unigram
% contamination. ``march''/``april'' polysemy is mild at these frequencies
% and left unfiltered.

\paragraph{Wavenumber assignment/purity (the coloring of
Figure~\ref{fig:months} A).}
Define the vectors
$c_k := \big(\cos(2\pi k t/12)\big)_{t=0}^{11}$ and
$s_k := \big(\sin(2\pi k t/12)\big)_{t=0}^{11}$ in $\R^{12}$, and let
$F_k := \operatorname{span}\{c_k, s_k\}$ be the ``Fourier plane'' of
wavenumber $k = 1, \dots, 5$ (for $k = 6$, the one-dimensional span of
the alternating vector $c_6$). Denote $P_k$ as the orthogonal projector onto
$F_k$. These subspaces orthogonally decompose the centered space, so for any unit eigenvector $v$ of
$P\Mstar P$ the ``energies'' $\norm{P_k v}^2$ sum to $1$ over
$k = 1, \dots, 6$. We assign $v$ the wavenumber capturing the most
energy, $k^{*}(v) = \arg\max_k \norm{P_k v}^2$, and call the winning
energy its \emph{purity}. $\norm{P_{k^*}v}^2 \in [0,1]$: $1$ is perfectly in one of these subspaces, while a uniformly random unit direction scores
$2/12 \approx 0.17$ in expectation ($1/12$ for $k = 6$). Leading-mode
purities: $0.99, 0.86$ ($k{=}1$ pair) and $0.92, 0.88$ ($k{=}2$ pair).

% \paragraph{Wavenumber assignment/purity 
% (the coloring
% of Figure~\ref{fig:months} A)} For $k = 1, \dots, 5$ let
% $F_k = \operatorname{span}\{\cos(2\pi k t/12),\,
% \sin(2\pi k t/12)\}_{t=0..11} \subset \R^{12}$ be the
% two-dimensional Fourier plane of wavenumber $k$ (one-dimensional for
% $k = 6$, the alternating vector), and $P_k$ the orthogonal projector
% onto it. For a unit eigenvector $v$ the energies
% $\norm{P_k v}^2$ sum to $1$ over $k = 1..6$ (the Fourier planes are an
% orthogonal decomposition of the centered space); $v$ is assigned the
% wavenumber capturing the most energy,
% $k^{*}(v) = \arg\max_k \norm{P_k v}^2$, and its \emph{purity} is that
% winning energy $\norm{P_{k^{*}} v}^2 \in [0, 1]$ ($1$ = a pure
% harmonic; chance level for a random direction is $2/11 \approx 0.18$).
% Leading-mode purities: $0.99, 0.86$ ($k{=}1$ pair) and $0.92, 0.88$
% ($k{=}2$ pair). 

\paragraph{Corpus Bootstrap (errors bars of Figure \ref{fig:months} A). }
The corpus is divided into $40$ contiguous segments (${\sim}425$k tokens each).  A resample draws 40 of these segments \emph{with replacement}, forming a new $\Mstar$ from which new eigenvalues can be computed.
$300$ resamples give percentile ($2.5\%$--$97.5\%$) intervals. 

% each segment's sufficient statistics---distance-weighted pair counts,
% month counts, token totals---are accumulated once; a resample draws
% $40$ segments \emph{with replacement}, sums their statistics, and
% recomputes the full pipeline ($\Mstar$, kernel, centered
% eigendecomposition) 

% Contiguous segments are the resampling unit because nearby
% co-occurrences are dependent (one passage generates several correlated
% events); segments far longer than the window keep that dependence
% intact \emph{within} units while the units themselves are approximately
% independent---an event-level bootstrap would understate every interval.
% Results: $95\%$ half-widths of $0.03$--$0.10$ on $c(d)$, the eigenvalue
% error bars in panel (b), and a significant half-year uptick,
% $c(6) - c(4) = 0.090$ with $95\%$ CI $[0.012, 0.162]$ (``January and
% July'' constructions are real). Within each predicted multiplicity-$2$
% pair the eigenvalue intervals overlap while the $k{=}1$ and $k{=}2$
% blocks remain separated: the degeneracy pattern survives its own error
% bars.

\paragraph{Theory vs. Empirical (Figure \ref{fig:months} C).} Our theoretical predictions take the form as in Appendix $\ref{app:special_finite}$, using $\Mstar$ to estimate the kernel $C$ (by averaging over appropriate elements).  We align our predictions to each empirical plane with by solving an orthogonal Procrustes problem.  The first PCs align with our theoretical predictions at a per month cosine of $0.96$ (median $0.97$).
A month label-shuffle null has mean per-month cosine of $0.35$ ($95\%$ $0.53$) across 500 shuffles.

% Comparison uses strict
% block-for-block slices (theory block $k$ vs.\ empirical modes of the same
% rank), with one $O(2)$ gauge per plane fitted by orthogonal Procrustes;
% per-month cosine $0.96$ (median $0.97$); the label-shuffle null
% (permute the month-to-theory-point assignment, refit and scale,
% $500$ permutations) has mean $0.35$ and $95$th percentile $0.53$. Angular
% steps: adjacent-month angle in the $k{=}1$ plane, median $30.9^\circ$
% (theory $30^\circ$); in the $k{=}2$ plane, $66^\circ$ (theory
% $60^\circ$). 

% Worst-predicted months after
% filtering: October ($0.89$), July ($0.93$).

\section{Music: experimental details}
\label{app:music-details}

\paragraph{Corpus and chord extraction.} All $371$ Bach chorales from
the \texttt{music21} corpus \citep{cuthbert2010music21}. Scores are
``chordified'' into beat-level sonorities. Each sonority is reduced to its
pitch-class set and matched against the $24$ major/minor triads
(inversions and octave doublings identified; sevenths and other
sonorities retained as their own types). This yields
$24{,}374$ chord tokens over a vocabulary of $92$ types, of which the
$24$ triads ($78\%$ of tokens) are the subsystem under consideration.
We form our $\Mstar \in \mathbb{R}^{24 \times 24}$ from this corpus as in equation \ref{eq:mstar-def} (settings are $B = 3$, $f(\delta) = 1$, $\alpha = 0.5)$. 

\paragraph{Accessing quality of statistical hypotheses.}  Suppose we have a group $G'$ (which acts on the 24 triads) under which we suspect that $\Mstar$ could be approximately statistically invariant.  We can score how well this hypothesis holds by first constructing the projection of $\Mstar$ onto matrices that are exactly statistically invariant,  
$\Pi_{G'}\Mstar_{x,y} = \tfrac1{|G'|}\sum_{g \in G'} \Mstar_{gx,\,gy}$, setting matrix elements which would be equal to their average.  In general, a hypothesis need not be a group; it must only specify which elements of $\Mstar$ would be equal in an ideal situation.  
We then compute our score
\[
R^2 
\;=\;
\frac{\sum_{c} n_c\,\big(\overline{\Mstar}_{c} - \overline{\Mstar}\big)^2}
     {\sum_{x,y}\big(\Mstar_{xy} - \overline{\Mstar}\big)^2},
\]
where $c$ runs over the equivalence classes of pairs under the candidate
invariance, $n_c$ is the number of entries in class $c$,
$\overline{\Mstar}_c$ is the class mean of equivalence class $c$, and $\overline{\Mstar}$ is the
grand mean over all entries.  In other words, this is the fraction of the entry-wise variance of
$\Mstar$ explained by the class-wide averaging.

The $T/I$ group obtains an $R^2$ of $0.822$.  An illustration of its performance is shown in Figure \ref{fig:main_music} A.  None of $2000$ random relabeling of the chords are able to surpass this value.  The $Z_{12}$ hypothesis, which is the same as $T/I$ but with only transposition-invariance, scores barely higher at $0.830$.  It is impossible for $T/I$ to surpass $\mathbb{Z}_{12}$ as it has additional restrictions, yet it performs essentially the same.  This is strong evidence that the $T/I$ symmetry is valid.  Other hypotheses, as defined in Table $\ref{tab:models}$ score notably worse ($\mathbb{Z}_{24}$: $0.699$, $PLR$: $0.442$, $\mathbb{Z}_{12} \times \mathbb{Z}_{12}: 0.440$, latent $\mathbb{Z}_{12}: 0.632$, $T/I$ conjugacy classes: $0.435$).  Notably the $T/I$ conjugacy class hypothesis, which adds even more restrictions to the $T/I$ hypothesis, scores much worse, suggesting that $T/I$ is the sharpest symmetry that is reliably present in the data.

% Furthermore, it seems like we cannot obtain a sharper symmetry that the $T/I$ group.  If we consider conjugacy classes of the $T/I$ group, which would enforce even more elements of $\Mstar$ to be set equal, the $R^2$ plummets to ? Finally, an alternative $D_{12}$ group which acts on the 24 triads in music theory is the classic $PLR$ group.  While $T/I$ acts on pitches in a chord independently, $PLR$ does not share this property.  The $PLR$ group scores poorly with an $R^2$ of $?$.

% Scores: $\Z_{12}$
% (transposition only; invariant = function of interval and ordered
% quality pair; $48$ parameters) $0.811$; $D_{12}$ ($24$ parameters)
% $0.804$; $D_{12}$ class functions ($9$ conjugacy classes) $0.366$;
% right-$D_{12}$ (PLR) $0.375$. Permutation test: the $D_{12}$ $R^2$ recomputed
% under $2{,}000$ random relabelings of the $24$ triads; observed rank
% gives $p = 5\times10^{-4}$.

% \paragraph{Embeddings.} Top $6$ eigenpairs of the centered
% $\Mstar$.

\begin{figure}[t]
\centering
\includegraphics[width=0.8\textwidth]{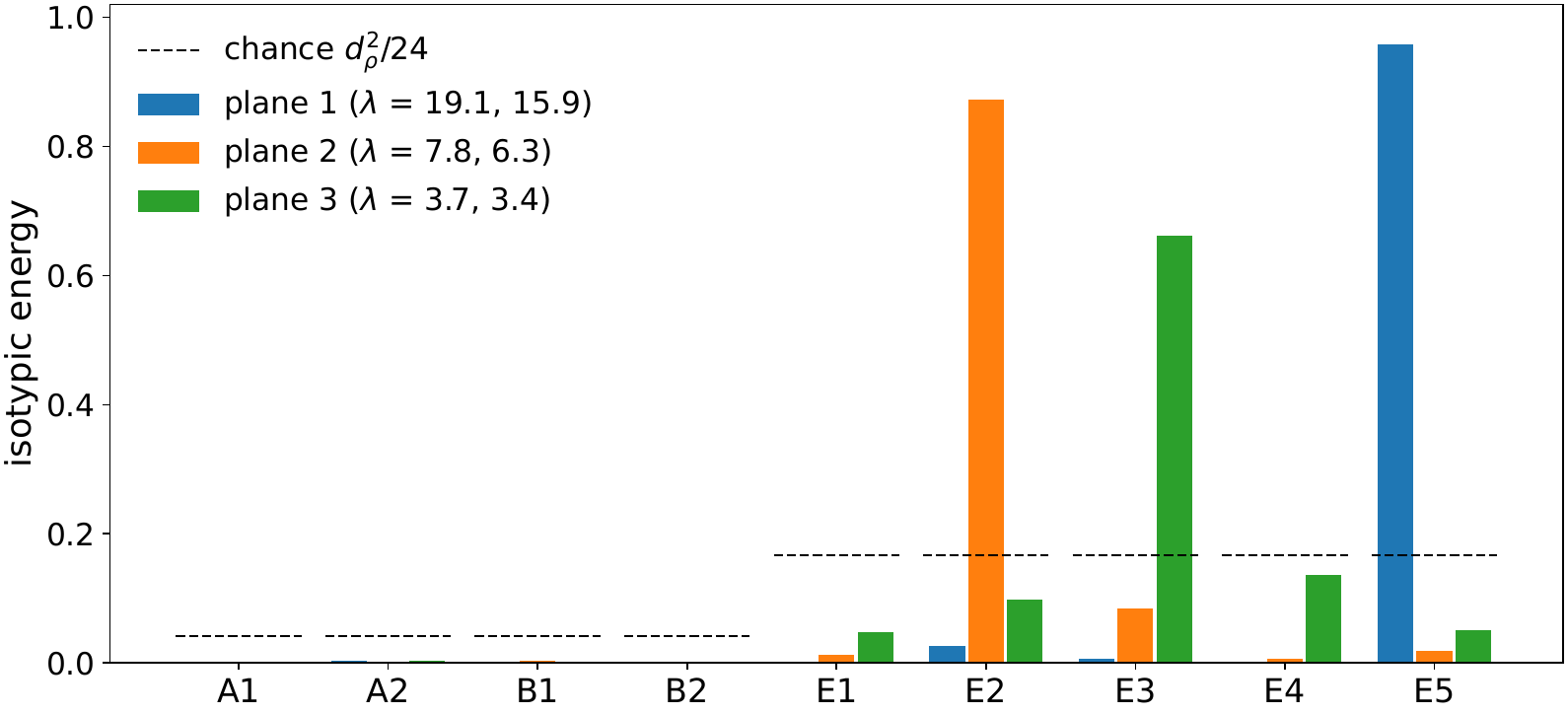}
\caption{Irrep content of the top three embedding planes: we consider the first six modes of the centered $\Mstar$.  One-dimensional irreps are $A_1, A_2, B_1, B_2$ and two-dimensional irreps are $E_1,\ldots E_5$.  $A_1$ corresponds to the trivial irrep which centering automatically removes.  We compute the isotypic energy of each plane for each irrep.  One irrep
dominates per plane ($E_5, E_2, E_3$; energies $0.958, 0.873, 0.662$ vs.\
chance $1/6$).  Eigenvalues of centered $\Mstar$ are listed in legend; notice they approximately pair.}
\label{fig:irrep-content}
\end{figure}

\begin{figure}[t]
\centering
\includegraphics[width=0.8\textwidth]{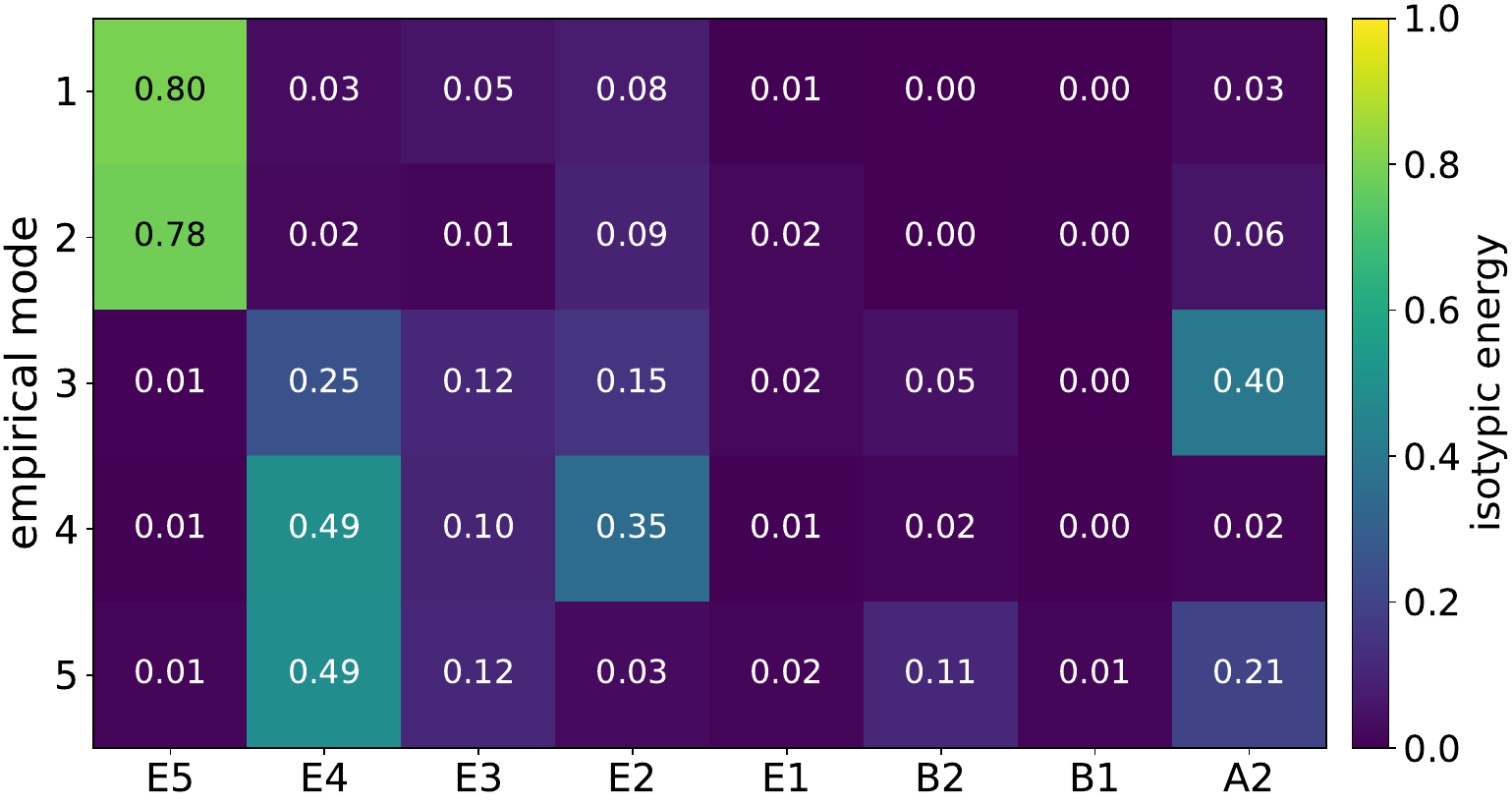}
\caption{Irrep content of ChordBERT layer 3 PCs: one-dimensional irreps are $B_1, B_2, A_2$, and two-dimensional irreps are $E_1, E_2, E_3, E_4, E_5$.  We do not show the remaining trivial irrep $A_1$ as it is removed automatically from centering.  The first 2 PCs demonstrate high isotypic energy in $E_5$.} 
\label{fig:irrep-content_BERT}
\end{figure}

\paragraph{Isotypic energy.} The $T/I$ action on the $24$ triads is a permutation
representation $L(g) \in \R^{24\times 24}$ (the regular representation acting on $\mathbb{C}^{24}$).  By the canonical Peter--Weyl decomposition of the regular representation,
$\C^{24} \cong \bigoplus_{\rho} V_\rho^{\oplus d_\rho}$: every irrep appears, with multiplicity equal to its dimension.  In other words, the space $\C^{24}$ can be canonically decomposed into ``isotypic'' subspaces corresponding to each given irrep of $T/I$.  The orthogonal
projector onto the $\rho$-component is given by 
\[
P_\rho \;=\; \frac{d_\rho}{24} \sum_{g \in T/I} \overline{\chi_\rho(g)}\, L(g).
\]
See \cite{serre1977linear} for more details on orthogonal projectors onto isotypic components.
For $T/I$ acting on triads, the two-dimensional irreps $E_h$, $h\in [5]$, 
each have isotypic dimension $d_\rho^2 = 4$.  The \emph{isotypic energy}
of an empirical plane (defined by orthonormal $v_1$, $v_2$) in
irrep $\rho$ is defined as $\tfrac12(\norm{P_\rho v_1}^2 + \norm{P_\rho v_2}^2) \in [0,1]$
with chance level for
a random plane $d_\rho^2/24$ (1/24 and 1/6 for one and two-dimensional irreps respectively). Note that the isotypic energy is invariant to the choice of $v_1$, $v_2$.  For computing the isotypic energy of a single vector $v$, we compute $\norm{P_{\rho} v}^2 \in [0,1]$ instead. See Figure~\ref{fig:irrep-content} and Figure~\ref{fig:irrep-content_BERT} for the isotypic energies corresponding to Bach chorales and ChordBERT layer 3 respectively.

\paragraph{ChordBERT activations}
We input the 371 Bach Chorales separately into ChordBERT (\cite{chordbert}) and in layer 3's residual stream average the activations corresponding to each of the 24 triads over every single one of its occurrences.  We focus on layer 3 in particular as its first 2 modes have very high isotypic energy in irrep $E_5$ (Figure \ref{fig:irrep-content_BERT}), suggesting the potential presence of a circle of fifths (which does indeed emerge, Figure \ref{fig:main_music} B).

\paragraph{Theory vs. Empirical (Figure \ref{fig:main_sky} B)} For the left subplot we use the $\Mstar$ formed from the 371 Bach chorales to form empirical embeddings.  On the right, 
we instead consider the averaged embeddings per triad in 
layer 3 of ChordBERT, as discussed above. In place of $\Mstar$ we consider the Gram matrix across the centered embeddings ($\in \mathbb{R}^{24 \times 24})$, normalized by the root mean square of its diagonal.  Our theoretical predictions take the form as in Appendix $\ref{app:special_finite}$ in both subplots, with the kernel $C$ being estimated from $\Mstar$ or the Gram matrix respectively (by averaging over appropriate elements).  A fitting orthogonal transformation aligns the theoretical predictions with the empirical embeddings in both cases.  The per-chord cosine is $0.98$ (median $0.99)$ for the chorales and $0.91$ (median $0.95$) for ChordBERT.  A chord relabeling null with 500 permutations gives $0.40$ ($95\%$ $0.56$) and $0.29$ ($95\%$ $0.48$) respectively in each case.

Furthermore, recall from Figure \ref{fig:irrep-content} that the first three planes of the Bach chorale embeddings all have high isotypic energy in their respective 2D irreps.  This suggests comparing our theoretical predictions against the first 6 dimensions of the empirical embeddings from $\Mstar$, using one fitting orthogonal transformation per plane.  We obtain a per-chord cosine similarity of $0.95$ (median $0.96)$ overall with per-plane values of  $0.98$, $0.94$, and $0.80$, respectively.  Our theory certainly holds predictive power beyond the first 2 dimensions, but weakens as we look as less prevalent modes.

\begin{figure}[t]
\centering
\includegraphics[width=\textwidth]{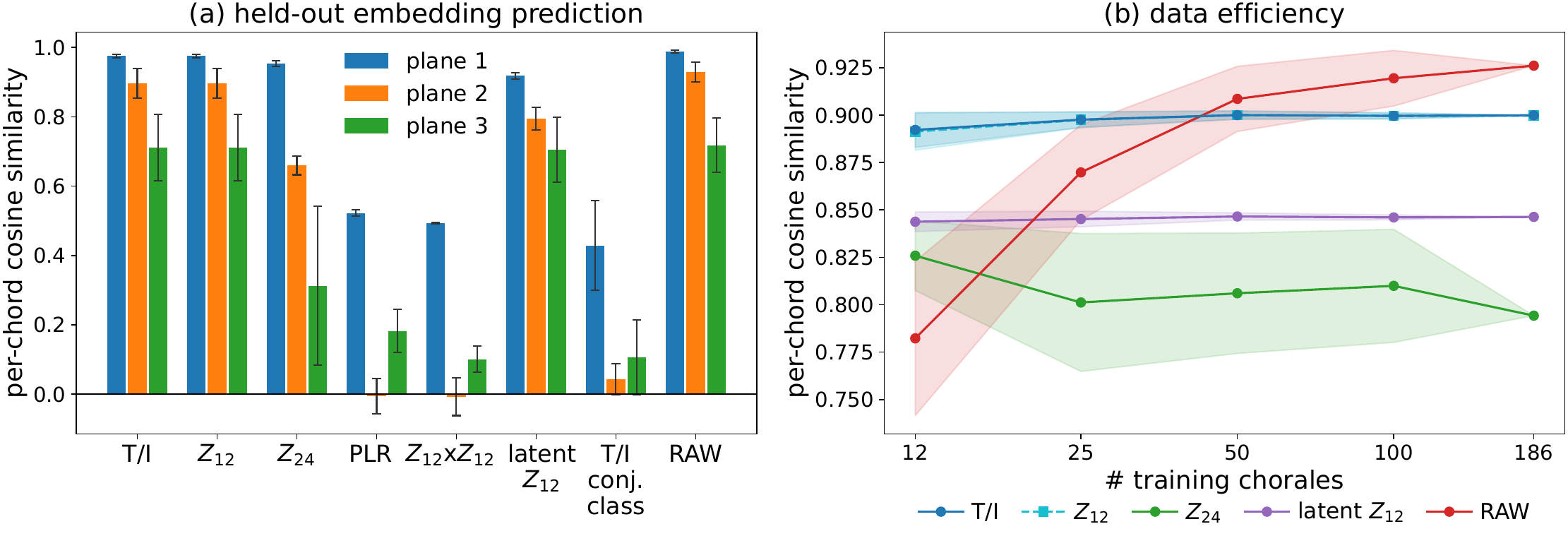}
\caption{(a)~Held-out embedding prediction per plane for the eight
hypotheses of Table~\ref{tab:models}: mean $\pm$ sample standard
deviation over $10$ random splits of the chorales into disjoint
training/test halves. (b)~Data efficiency: Training sets are nested subsets drawn without replacement
from one fixed half of the $371$ chorales, the other half serving as a
fixed test set.  Each point averages $15$ redraws (shaded: $\pm 1$
sample standard deviation), except the rightmost, where the training
set is the entire pool and no redraw is possible. The $19$-parameter $T/I$
hypothesis converges by ${\sim}12$ chorales.  The $300$-parameter
unconstrained model (RAW) needs $\approx 4\times$ more data
to match it.}
\label{fig:model-comparison}
\end{figure}

\paragraph{Model comparison protocol.} Chorales are split into disjoint
halves ($10$ random splits). Each hypothesis is fit on the training half.  Namely, the empirical $\Mstar$ is projected onto a matrix which satisfies the hypothesis $G'$, i.e.,  $\Pi_G^{'} \Mstar$ as defined above
(the RAW model uses the
training $\Mstar$ itself without projection).  $\Pi_G^{'} \Mstar$ then predicts the \emph{held-out} half's
embeddings.  Namely, the first 6 modes of the centered $\Mstar$ (of the held out data) are aligned with the modes of $\Pi_{G^{'}}\Mstar$ (from the training data) using one orthogonal transform per plane.  Per-chord cosines are then computed.  Results for various hypotheses (Table \ref{tab:models}) are shown in Figure \ref{fig:model-comparison} A.  The data itself without any imposed invariances,``RAW'', performs the best by a small margin.  However, $\mathbb{Z}_{12}$ and $T/I$ don't do much worse.  As mentioned before, the fact that $T/I$ performs about as well as $\mathbb{Z}_{12}$, when it has more restrictions and less independent parameters, is a great indication that $T/I$ is a natural symmetry of the data.  Figure~\ref{fig:model-comparison} B performs a similar analysis as in A, but instead
refits on nested subsets with a fixed held out test set. $T/I$ attains a cosine similarity with the held out data of $0.892$ at $12$ chorales, the best performance of all models at this point; RAW requires around $50$ training chorales to match.

% per-chord cosine on the top 6 modes configuration after a global
% rotation, exactly as in the overlay. Results
% (Figure~\ref{fig:model-comparison}a; mean per-chord cosine for plane 1): $D_{12}$
% $0.942$; RAW $0.945$; $\Z_{12}$ $0.942$; latent-$\Z_{12}$ $0.890$;
% $\Z_{24}$ $0.842$; $\Z_{12}\times\Z_{12}$ $0.714$; radial
% $0.712$; PLR $0.687$. Plane~2 separates the hypotheses more sharply.  The fact there is a $\Z_{12}$ tie between $\Z_{12}$ and $D_{12}$in performance is very informative. $D_{12}$-invariant matrices are a strict subclass of $\Z_{12}$-invariant ones.  $\Z_{12}$ has 26 independent parameters while $D_{12}$ only has 19 so $\Z_{12}$ can never fit worse by construction---that
% the constrained $D_{12}$ model nonetheless \emph{matches} it on
% held-out data means the freedom $\Z_{12}$ retains and $D_{12}$ forbids
% carries no predictive signal: the inversion symmetry is really present
% in the statistics, not merely unpunished. The
% learning curve (

\begin{table}[t]
\centering\small
\begin{tabular}{@{}l r r p{7.6cm}@{}}
\toprule
Model & Params & Indep. & Invariance assumed on $\Mstar$ over the
$24$ triads\\
\midrule
$T/I$ & $24$ & $19$ & Statistical invariance under $T/I$. \\
$\Z_{12}$ & $48$ & $26$ & Statistical invariance under only transposition of $T/I$. \\
$\Z_{24}$ & $24$ & $13$ & Invariance under a ``transposition'' that moves along both majors and minors as in the circular order seen in Figure \ref{fig:main_music} B.  This hypothesis predicts embeddings where the major and minor circles are perfectly inter-spaced.\\
$PLR$  & $24$ & $19$ & Statistical invariance under the classic $PLR$ group of music theory\\
$\Z_{12}\times\Z_{12}$ & $26$ & $15$ & Statistical invariance under transposition, but separately for both majors and minors. (let $m,m+1,m+2$ be sequential major chord indices while $n,n+1,n+2$ correspond to minors) E.g., $\Mstar_{m,m+1} = \Mstar_{m+1, m+2}$ and $\Mstar_{n, n+1} = \Mstar_{n+1,n+2}$ but in general we have $\Mstar_{m,m+1} \neq \Mstar_{n+1,n+2}$)  Major and minor cross terms of $\Mstar$ are set to one constant.  This hypothesis essentially assumes independent circular structures in majors and minors.
\\
latent $\Z_{12}$ & $12$ & $7$ & Majors are identified with their relative minors (music theory term, this identification looked reasonable from Figure \ref{fig:main_music} B), and invariance under transposition is assumed.  This hypothesis corresponds to the embeddings forming a single circle with majors on top of minors. \\
$T/I$ conjugacy class & $9$ & $9$ & We begin with statistical invariance under $T/I$.  Each equivalence class corresponds to a group element.  We add the restriction that if two or more of these group elements correspond to the same conjugacy class of $T/I$ (different type of class, group theoretic concept), then the equivalence classes are combined.\\
RAW & $576$ & $300$ & No invariance, the raw matrix\\
\bottomrule
\end{tabular}
\caption{The eight statistical hypotheses of the music model comparison.
Each hypothesis is applied to $\Mstar$ by setting elements of $\Mstar$ which would be equal under the hypothesis to their average
(Except RAW which is $\Mstar$ itself). \emph{Params} counts the model's number of equivalence classes without considering the symmetry of $\Mstar$. \emph{Indep.}\
counts the number of free parameters once this symmetry
merges equivalence classes together.
% The diagonal ---
% a self co-occurrence statistic, forced constant under each transitive
% invariance --- contributes one class to most models (two, one per
% quality, to $\Z_{12}$ and $\Z_{12}\times\Z_{12}$; latent $\Z_{12}$
% ties it into its distance-$0$ class), and $24$ free entries to RAW.
}
\label{tab:models}
\end{table}

\begin{figure}[t]
\centering
\includegraphics[width=\textwidth]{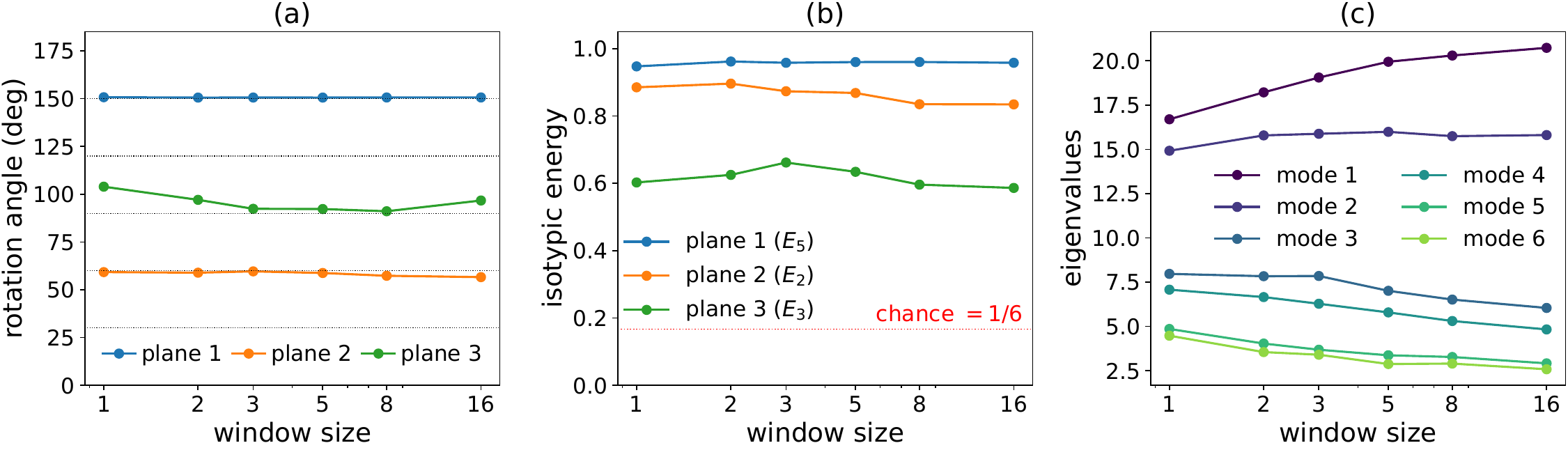}
\caption{We use various window sizes for constructing the centered $\Mstar$ (the original size was 3) which we then use to form 6-dimensional chord embeddings.   
(a) Rotation angles are computed by fitting a rotation matrix to each pair of dimensions (plane) of the embeddings; we consider the angle between chords in chromatic order separately for both the majors and minors.
(b) Isotypic energy of each plane in its respective irrep.  Chance level is always $d_{\rho}^2/24 = 1/6$ for the considered irreps. (c) eigenvalues of the centered $\Mstar$.}
\label{fig:window-sweep}
\end{figure}

\paragraph{Changing window size for constructing $\Mstar$.} As seen in Figure \ref{fig:window-sweep}, we construct $\Mstar$ and consider the corresponding embeddings for a variety of window sizes.
We see that the fitted rotation angle and isotypic energy remain largely consistent. Plane 3's values change somewhat, but this kind of behavior is understandable in weaker modes of the data.
Finally, we also consider the eigenvalues of the centered $\Mstar$ in Figure \ref{fig:window-sweep} C.  While rotation angle and isotypic energy are largely functions of the statistical symmetries of the data, the eigenvalues are very dependent on the corpus; corpora that have the same symmetries can still have wildly different eigenvalues.  Hence, it is not too surprising to see that the eigenvalues have noticeably more drift than the other considered quantities.

\section{Sky: experimental details}
\label{app:sky-details}

\paragraph{LLM activations} Released data of \citet{berdnikov2026sky}:
$188$ named sky objects (stars, constellations, etc.) $\times$ $25$ proximity-style prompts,
residual-stream activations at all layers exported as the top-$16$ PCA
coordinates per layer. We use
mean activations across the prompts for each layer. The layer from which the true celestial positions can be linearly decoded the best is selected as the working layer (layer $72$ of $89$ for Mistral Large 2).  In place of $\Mstar$, we consider Gram matrix ($\in \mathbb{R}^{188 \times 188}$) of the sky objects' activations, normalized by the root mean square of its diagonal.  We will denote this matrix as $G_{\mathrm{Gram}}$ for convenience.  We will also use 188 and $n$ interchangeably in this section. 

% \paragraph{Weighted estimators.} Objects sample the sphere non-uniformly
% (the catalogued sky favors the northern celebrated regions), so all
% estimators are density-weighted: weights $w_i \propto 1/\hat\rho(x_i)$
% with $\hat\rho$ an on-sphere Gaussian KDE of bandwidth $0.35$ rad;
% centering, eigenproblems, and regressions are computed in the
% $\sqrt{w}$ frame (equivalently, with respect to the weighted inner
% product), so that they estimate the uniform-measure objects of
% Theorem~\ref{thm:funk-hecke}. Sampled real spherical harmonics are
% orthonormalized degree-by-degree in the weighted inner product (lower
% degrees kept pure).

\paragraph{Need for weighted estimators.} 
Our theory in Appendix $\ref{app:homog}$ intrinsically assumes a continuous setting in which an infinite set of words are in bijection with the sphere.  Of course, this is not practical.  In reality, we are discretizing the theory by sampling points on $S^2$ which correspond to sky objects on the celestial sphere.  If all celestial objects were sampled uniformly on the sphere, no weighted estimators would be required.  However, the celestial objects we consider sample the sphere in a ``lightly'' non-uniform fashion.  Acceptable estimators could still be obtained by ignoring this small bias, but we decide to be proactive in counteracting it.

\paragraph{Inverse density weighting}
% which we are discretizing assumes we have a uniform measure on a given homogeneous space, in this case the sphere.  Hence, to counter balance any bias from our non-uniform sampling, we utilize inverse density weighted estimators through this section.  
Essentially all estimators in this section incorporate inverse density-weighted.  We fit a density $\hat{\rho}$ to our $188$ objects using an on-sphere Gaussian kernel density estimator (KDE).  Namely, let \(x_1,\ldots,x_n\in S^2\) be the sky-object positions and
\(\theta_{ij}=\arccos(\langle x_i,x_j\rangle)\) the great-circle angle.
With \(h=0.35\)\,rad, the on-sphere Gaussian KDE is
\[
\hat\rho(x_i)
\;=\;
\sum_{j=1}^{n}
\exp\!\left(-\frac{\theta_{ij}^{2}}{2h^{2}}\right).
\]
A weight $w_i$ is now associated with each object $i$.  These weights inversely related to $\hat{\rho}$, and are
renormalized to sum to 1:
\[
\tilde w_i=\frac{1}{\hat\rho(x_i)},
\qquad
w_i=\frac{\tilde{w}_i}{\sum_{k=1}^{n}\tilde w_k}.
\]
Hence, areas of the sphere which were previously over or under-sampled are counterbalanced.  

\paragraph{Empirical Word Embeddings}
The integral operator in Appendix \ref{app:homog} on some $f \in L^2(S)$  discretizes as $(\mathcal{M}f) \approx \sum_j \Mstar_{ij} f(x_j)$, where the typical normalized co-occurence matrix $\Mstar$ emerges.  However, with reweighting, this is lightly modified to $\sum_j \Mstar_{ij} w_j f(x_j)$.  Something else affected by reweighting is centering.  The previous $P = I - \bm{1}\bm{1}^T/n $  is modified to $P_w = I - \bm{1}w^T$. 

Altogether, for computing empirical embeddings, using say the $\Mstar$ from our book corpus, we solve generalized eigenproblem 
\begin{align}
\label{eqn:gen_eigen}
P_w\Mstar WP_wv = \lambda v
\end{align}
with diagonal $W \in R^{188 \times 188}$ and $W_{ii} = w_i$.

\paragraph{Regressions}
All regressions in this section incorporate reweighting.

\emph{(i) Kernel estimation} (LLMs would use $G_{\mathrm{Gram}}$ in place of the book corpus's $\Mstar$) The kernel $C$ (as in Theorem \ref{thm:main-GK}) is the
weighted estimate of $\mathbb{E}[\Mstar \mid t]$,
\[
\tilde C(t) \;=\;
\frac{\sum_{i<j} w_i w_j\, K_h(t - t_{ij})\, \Mstar_{ij}}
     {\sum_{i<j} w_i w_j\, K_h(t - t_{ij})},
\qquad t_{ij} = \inner{x_i}{x_j},
\]
where $t_{i,j}$ is the true cosine between two objects $x_i$, $x_j$ on the celestial sphere.  $K_h$ is a 1D Gaussian with standard deviation $h=0.08$.  Note, the kernel as described here is a function of cosine similarity while in the main paper it is a function of angle.  The two forms are interchangeable.

The local standard deviation of this fit is computed as follows.  Let
\[
\kappa_{ij}(t)
\;=\;
w_i w_j\, K_h(t-t_{ij})
\]
For convenience, write
\[
\hat m_1(t)=\tilde C(t),
\qquad
\hat m_2(t)
\;=\;
\frac{\sum_{i<j} \kappa_{ij}(t)\, (\Mstar_{ij})^2}
     {\sum_{i<j} \kappa_{ij}(t)}.
\]
The local standard deviation of the kernel fit is then
\[
\hat\sigma(t)
\;=\;
\sqrt{\max\bigl(\hat m_2(t)-\hat m_1(t)^2,\,0\bigr)},
\]
i.e.\ the estimate of
\(\sqrt{\mathrm{Var}(\Mstar_{ij}\mid t_{ij}=t)}\).

\emph{(ii) Decoding}: Consider
the unit vectors $x_i \in \R^3$ representing the $n$ celestial objects.
Suppose we are trying to linearly decode these positions from some $U \in \mathbb{R}^{n \times k}$, e.g., the residual stream of any of the LLMs.  Let
$\hat B = \arg\min_B \sum_i w_i
\norm{x_i - B^{\top} U_i}^2$ where $B \in \mathbb{R}^{ k\times 3}$ and $U_i \in \mathbb{R}^k$.  We report the quality of the fit of $\hat{B}$ as
\[
R^2_{\mathrm{dec}} \;=\; 1 -
\frac{\sum_{i=1}^n w_i \norm{x_i - \hat B^{\top} U_i}^2}
     {\sum_{i=1}^n w_i \norm{x_i - \bar x_w}^2},
\]
where $\overline{x}_w = \frac{\sum_{i=1}^n w_i x_i}{\sum_{i=1}^nw_i}$.  This metric chooses the working layer of each model where $U$ become the centered activations of each given LLM (Table \ref{tab:models}).

\emph{(iii) Gram zonality:}
(books would use $\Mstar$ in place of $G_{\mathrm{Gram}}$ in this section).
Write \(t_{ij}=\langle x_i,x_j\rangle\) and
\(\theta_{ij}=\arccos(t_{ij})\).
Using only upper-triangle pairs \(i<j\) with weights
\(\omega_{ij}=w_i w_j\), partition
\(\theta\in[0,\pi]\) into \(B=18\) equal bins.
In bin \(b\) set
\[
\hat\mu_b
\;=\;
\frac{\sum_{(i,j)\in b}\omega_{ij}\, M_{ij}}
     {\sum_{(i,j)\in b}\omega_{ij}}
\]
With \(\hat M_{ij}=\hat\mu_{b(ij)}\) and
\(\bar M=\sum_{i<j}\omega_{ij}M_{ij}\big/\sum_{i<j}\omega_{ij}\),
\[
R^2_{\mathrm{zonal}}
\;=\;
1-
\frac{\sum_{i<j}\omega_{ij}(M_{ij}-\hat M_{ij})^2}
     {\sum_{i<j}\omega_{ij}(M_{ij}-\bar M)^2}.
\]
A permutation null (shuffle object labels, keep \(\{M_{ij}\}\),
refit the bin means) supplies a chance baseline for
\(R^2_{\mathrm{zonal}}\).

\paragraph{Sampling Spherical Harmonics}
Sampled real spherical
harmonics $(Y_\ell^m(x_i))_{i=1}^n$ are only approximately orthonormal at
finite $N$.  We orthonormalize them with Gram-Schmidt in the weighted inner product $\inner{\cdot}{\cdot}_w$ ($\inner{u}{v}_w = \sum_{i=1}^n w_i u_iv_j$ where $u,v \in \mathbb{R}^n$.)
We conduct this degree by degree, beginning with $\ell = 1$ and moving upwards (ending for whatever degree is high enough for our purposes). This yields a matrix
$Q_\ell \in \R^{n \times (2\ell+1)}$ for each $\ell$, whose columns are an orthonormal
basis of the sampled degree-$\ell$ subspace,
with $Q_\ell^{\top} Q_{\ell'} = 0$ for $\ell \neq \ell'$ by
construction.  These $Q_{\ell}$ are the orthonormalized spherical harmonics we show as our theoretical predictions in Figure \ref{fig:main_sky} C.
% Gram--Schmidt is order dependent; starting with low $degrees$ low degrees exactly pure and pushes all finite-sample mixing
% into higher degrees.  We make this choice, since the paper's
% claims prioritize the low degrees. 

\emph{Degree energies (Figure \ref{fig:main_sky} B):} Consider a unit
eigenvector $v$ of the weighted eigenproblem in equation \ref{eqn:gen_eigen}. The ``degree-$\ell$'' energy of $v$, $E_{\ell}(v)$, is the squared orthogonal projection onto that subspace,
\[
E_\ell(v) \;=\; \norm{Q_\ell^{\top}  v}^2
\;=\; \sum_{m=-\ell}^{\ell} \inner{q_{\ell m}}{v}^2
\;\in\; [0, 1],
\]
computed for $\ell = 0, \dots, 4$. The vector's assigned
degree is $\ell^{*}(\tilde u) = \arg\max_\ell
E_\ell(\tilde u)$, and its \emph{purity} in that degree is
$E_{\ell^{*}}(\tilde u)$.  This is the color bar of \ref{fig:main_sky} B. Note that the considered degrees in this figure consider
$\sum_{\ell \le 4}(2\ell{+}1) = 25$ of the $n = 188$ dimensions, so energies
need not sum to $1$.  The chance energy level for a random direction is
$(2\ell{+}1)/n$ ($\approx 0.016$/$0.027$ for $\ell =1,2$), so
purities of $0.5$--$0.9$ sit two orders of magnitude above chance.  Finally, note that
$E_0(\tilde u) = 0$ exactly due to centering.

\begin{table}[t]
\centering\small
\begin{tabular}{@{}l r r r r l@{}}
\toprule
Model & layer & $R^2_{\mathrm{zonal}}$ & $R^2_{\mathrm{dec}}$ & canon$^2$ (top-3) &
degree sequence\\
\midrule
Qwen3-235B & 90 & 0.581 & 0.928 & 0.94 & $1\,1\,1\,1\,2\,2\,2\,2\,2$\\
Mistral-Large-123B & 72 & 0.490 & 0.905 & 0.91 &
$1\,1\,1\,1\,2\,2\,2\,2\,3$\\
GLM-4.5-Air & 46 & 0.484 & 0.904 & 0.91 & $1\,1\,1\,1\,2\,2\,2\,2\,2$\\
Qwen3-32B & 62 & 0.440 & 0.865 & 0.87 & $1\,1\,1\,2\,1\,2\,2\,2\,3$\\
Llama-3.3-70B & 63 & 0.395 & 0.898 & 0.91 &
$1\,1\,1\,2\,1\,2\,2\,3\,3$\\
Mixtral-8x22B & 56 & 0.238 & 0.836 & 0.84 & 1 1 1 2 1 2 2 4\\
GPT-OSS-120B & 31 & 0.072 & 0.634 & 0.64 & scrambled\\
\bottomrule
\end{tabular}
\caption{Mistral-Large-123B is the same as Mistral Large 2.  Each model is evaluated at the layer
where celestial positions on the sphere ($\in \mathbb{R}^3$) can be linearly decoded best from the activations ($R^2_{\mathrm{dec}}$).  $R^2_{\mathrm{zonal}}$: the
fraction of the activation Gram matrix structure explained by true
angular separation alone. Canon$^2$ (top-3): average squared canonical cosines between the
three leading activation modes and the dipole harmonics (subspace
overlap; $1$ = identical). Degree sequence: the harmonic degree
($\ell$) of each successive eigenmode---a clean harmonic embedding
reads $1\,1\,1\,2\,2\,2\,2\,2\,3\ldots$ Five of seven models show strongly harmonic structure (relatively high zonality $R^2$ with 
degree-ordered modes).  Mixtral-8x22B is borderline, and GPT-OSS-120B
is the true outlier, with weak spherical harmonic structure by every measure.}
\label{tab:sky-models}
\end{table}

\paragraph{Selection of Star-Guide Corpus} Candidate pool: $24$ public-domain astronomy books (Project Gutenberg, ${\sim}1.6$M words). Not every astronomy book carries the
sky's geometry in its statistics.  Many are organized by mythology,
best telescope practices, or by a handful of famous constellations. A
minority are written as a general \emph{tour of the sky}, directing the reader from object to nearby object with co-mentions largely determined by geometrical positions. The metric below selects for this property. 

Tokenization maps every object
alias to a canonical token.  Each book's $\Mstar$ (equation \ref{eq:mstar-def}, $B = 30, f(\delta) = 1, \alpha = 0.25$) is formed using objects with
$\ge 5$ mentions (fallback $3$, marked $\dagger$ in Table \ref{tab:astrobooks}). 
Consider the normalized co-occurence rate
$R_{ij}  = \frac{P_{ij}}{P_i P_j}$ as in equation \ref{eq:R}; this counts how much more or less often a pair of objects $i$ and $j$ co-occur relative to how much their individual prevalences would predict.
We would ideally select books where pairs of objects have high $R_{ij}$ if they are geometrically close and vis-versa.  Furthermore, recall that $\Mstar_{ij}$ is a strictly increasing function of $R_{ij}$  while also being bounded, which prevents any issues with $R_{ij}$'s heavy tail.
% Let $W_{ij}$ count within-window co-mentions and $n_i$ mentions; under
% independence the pair's expected number of co-occurrences is
% $\mathbb{E}_0[W_{ij}] \approx n_i n_j \cdot 2F/N$. The natural measure
% of association is then the ratio
% $R_{ij} = W_{ij}/\mathbb{E}_0[W_{ij}] =  P_{ij}/( P_i
%  P_j)$: how much more (or less) often the pair co-occurs than
% its members' popularities alone predict. Our $\Mstar$ is a strictly
% increasing function of exactly this quantity,
% $\Mstar_{ij} = 2(R_{ij}-1)/(R_{ij}+1)$---a bounded quantity that tames the $R_{ij}$'s heavy tail while
% preserving its order. 

This line of reasoning leads us to select our books by 
\begin{equation*}
D \;=\; \operatorname{corr}\big(\Mstar_{ij},\, \cos\theta_{ij}\big)
\quad \text{over unordered pairs, } \; p < 0.05
\end{equation*}
with $p$ coming from a one-sided location-permutation null (500 trials).  $\theta_{i,j}$ is the true angle between two astronomical objects $i$ and $j$ on the celestial sphere. Passing
books are \emph{A Field Book of the Stars} (Olcott 1907; $D = +0.34$,
$p < .004$), \emph{Round the Year with the Stars} (Olcott 1912;
$+0.45$, $p < .004$), \emph{Astronomy with an Opera-Glass} (Serviss
1888; $+0.22$, $p = .01$), \emph{Astronomy for Young Australians}
(Bonwick 1866; $+0.26$, $p = .01$).  In total we obtain $131$k tokens of star guide prose with $39$ sufficiently counted objects.  Notably, the selected 4-book corpus kernel correlates with the model's activation kernel at $0.95$, so they have similar structure.

Simpler statistics fail
in instructive ways: narrative
\emph{coherence} (median angular hop between successive distinct
mentions vs.\ a shuffled null, Table \ref{tab:astrobooks}) completely ignores whether successive mentions actually fall
within our $\Mstar$'s window. Alternatively, a metric completely based on co-occurrences within the window, something like the mean $\cos\theta$ of all co-mentions in the same window, also typically fails.  These metrics are often confounded by jointly famous pairs (e.g. Castor--Pollux) that are often mentioned together, but ignores all the geometrically close pairs that are never mentioned.  $D$ avoids these problems by working with $\Mstar$ and considering all pairs.
% The
% windowed event statistic (mean $\cos\theta$ over within-window events)
% incorporates the window but, computed on raw counts, is confounded by jointly
% famous pairs (e.g. Castor--Pollux) that tend to be mentioned together and is
% blind to the pairs that never co-occur despite many
% opportunities. $D$ incorporates both the window and 
% ``opportunity normalization'', and each pair co-occurrence votes in this metric.

A caught negative shows the metric working: \emph{Astronomy for Young
Folks} (monthly chapters revisiting the same famous celestial objects) is
strongly coherent ($z = 8.1$) yet fails on our selection metric ($D = -0.08$,
$p = 0.83$). Similarly, a large generic corpus will generally not perform well at this metric.  For example, \texttt{text8} ($17$M tokens of
cleaned Wikipedia, $130\times$ the size of the selected guides) demonstrably does
not.  Its $D = +0.01$ ($p = 0.48$) and coherence $z = 4.6$ and are both far behind the selected corpus (Table \ref{tab:astrobooks}).  \texttt{text8} is dominated by celestial terms with alternative meanings (algol, peacock, mimosa, etc.).  Any signal it does possess seems to arise from jointly famous pairs of celestial objects, a confounding that $D$ removes as discussed before.

\begin{table}[t]
\centering
\small
\begin{tabular}{@{}l l r r r r r@{}}
\toprule
Book & Author, year & ment. & $D$ ($p$) & coher.\ ($z$) & $\lambda_1$ ($p$) & kernel corr \\
\midrule
\textbf{Round the Year with the Stars} & Olcott, 1912 & 361 & \textbf{+0.45 (0.00)} & 30 (8.8) & +1.98 (0.00) & +0.86 \\
\textbf{A Field Book of the Stars} & Olcott, 1907 & 345 & \textbf{+0.34 (0.00)} & 46 (14.3) & +1.69 (0.00) & +0.85 \\
\textbf{Astronomy for Young Australians} & Bonwick, 1866 & 116 & \textbf{+0.26 (0.01)$^\dagger$} & 46 (9.6) & +1.16 (0.01) & +0.74$^\dagger$ \\
\textbf{Astronomy with an Opera-Glass} & Serviss, 1888 & 364 & \textbf{+0.22 (0.00)} & 33 (7.4) & +0.86 (0.03) & +0.75 \\
Half-Hours with the Telescope & Proctor, 1868 & 101 & +0.14 (0.15)$^\dagger$ & 25 (3.5) & +0.34 (0.08) & +0.50$^\dagger$ \\
Recreations in Astronomy & Warren, 1879 & 158 & +0.11 (0.12)$^\dagger$ & 18 (3.8) & +0.19 (0.17) & +0.61$^\dagger$ \\
Astronomy for Amateurs & Flammarion, 1904 & 136 & +0.02 (0.38)$^\dagger$ & 25 (4.2) & $-$0.03 (0.50) & +0.11$^\dagger$ \\
Pleasures of the Telescope & Serviss, 1901 & 156 & +0.01 (0.45)$^\dagger$ & 21 (2.6) & +0.01 (0.51) & $-$0.12$^\dagger$ \\
Astronomy for Young Folks & Lewis, 1922 & 301 & $-$0.08 (0.83) & 26 (8.2) & $-$0.21 (0.72) & $-$0.21 \\
\midrule
\emph{text8} (general Wikipedia, 17M tok) & --,\;--  & 792 & $+0.01$ (.48) &
79 (4.6) & $+0.00$ (.53) & $+0.18$\\
\bottomrule
\end{tabular}
\caption{
Corpus selection (leading 9 candidates of $24$).
\textbf{Selection column} $D = \operatorname{corr}(\Mstar, \cos\theta)$
with one-sided object celestial sphere location-permutation $p$; boldface: selected at
$p < 0.05$ with 500 shuffles.  Coherence is the median angular
hop between successive distinct mentions in degrees.  
$z \;=\; (\bar m_{\mathrm{null}} - m_{\mathrm{obs}})/\sigma_{\mathrm{null}}$ is
the hop's deficit below the object textual position order-shuffled null (500 shuffles), with
$\bar m_{\mathrm{null}}$ and $\sigma_{\mathrm{null}}$, the mean and
standard deviation of the median hop across all shuffles.
$\lambda_1$ is the empirical $c_1$ from Theorem \ref{thm:main-GK} using the
book's kernel; 500 shuffles with object celestial sphere location-permutation are used to compute $p$. Kernel corr is the correlation
with the LLM activation kernel. All listed books contain $\ge 15$ well-counted ($\geq 5$ mentions, $^{\dagger}$: fall back of 3 mentions) objects. }
\label{tab:astrobooks}
\end{table}

% \paragraph{Kernel agreement and bandwidth matching.} The corpus kernel
% correlates with the model's activation kernel at $0.92$; after projecting both to $\ell \le 2$ (Legendre low-pass), $0.97$. A scale sweep as we add additional books 
% shows agreement \emph{peaks} at the small pure-genre corpus and declines
% with topical dilution: genre purity beats token count.
 
% \paragraph{Negative control: a general corpus.} A large general corpus
% need not carry sky statistics, and \texttt{text8} ($17$M tokens of
% cleaned Wikipedia, $130\times$ the selected guides) demonstrably does
% not. After matching the book pipeline's conventions: $792$ mentions of the $188$ objects, dominated by celestial terms with alternative meanings (``algol'' the
% programming language, $148$; ``peacock'' the bird, $46$; ``mimosa'' the
% plant or drink, $23$); coherence $z = 4.6$, below the guides' $8$--$14$;
% $D = +0.01$ ($p = 0.48$) and $\lambda_1 = +0.000$ ($p = 0.53$). Kernel correlation with the model is $+0.18$ far below
% the guides' $+0.74$--$+0.86$.  Much of the signal that does
% exist concentrates on jointly famous, intrinsically close pairs
% (Castor--Pollux, etc.)---the confound $D$'s opportunity
% normalization removes. Genre, not volume, appears to carry the sky.
 
\paragraph{Theory vs. Empirical (Figure~\ref{fig:main_sky} C).}
We compare the centered Mistral Large 2 activations and the modes of the book corpus's centered $\Mstar$ with the dipole (degree 1) spherical harmonics evaluated at the object's true positions.  Note that spherical harmonics as orthonormalized as in the Sampling Spherical Harmonics paragraph above.  We will utilize the same $Q_{\ell}$.

\emph{Subspace alignment:}  There are $2\ell +1$ spherical harmonics of degree $\ell$, but the number of modes of the LLM activations with high degree $\ell$ purity often exceed $2\ell +1$ (Figure \ref{fig:main_sky} B).  Hence, for comparing our theoretical predictions with the empirical LLM activations, we first find the ``best aligned subspace'' (within some span of PC modes) in the activations with our theoretical predictions.  For the dipole ($\ell$ =1) alignment, we use PCs 1-4, while for quadrupole ($\ell = 2$) alignment, we instead use PCs 5-12.

The actual alignment for degree $\ell$ is conducted with canonical angles.  Let $U \in \mathbb{R}^{n \times k}$ have columns of PCA modes under consideration, where $k \geq 2\ell +1$.  Let the singular value decomposition (SVD) of $Q_{\ell}^T U$ take the usual form of $Q_{\ell}^T U = \Phi \Sigma \Psi^T$, where $\Phi \in \mathbb{R}^{(2\ell+1) \times (2\ell+1)}$, $\Sigma \in \mathbb{R}^{(2\ell+1)\times k}$, and $\Psi \in \mathbb{R}^{k \times k}$.  Our best aligned subspace is then $U\Psi_{:,1:2\ell+1}$; this becomes our empirical data.

Note that this procedure is not conducted for the book corpus.  We directly consider its $\Mstar$ modes 1-3 for dipole and 4-8 for quadrupole.  

\emph{Point by point alignment:}
After any potential subspace alignment, we solve an orthogonal Procrustes problem to align our theoretical predictions point wise with the empirical data for either the LLMs or the books.  A global scaling is also fit for the LLMs, but not for the books.  For the books, we utilize our theoretically predicted scaling as in Theorem \ref{thm:main-GK}.

\emph{Objection permutation null:} Our null in Figure \ref{fig:main_sky} (and for the quadrupole case) consists of randomly assigning objects to different positions and performing the exact same analysis.  We utilize 300 permutations.

\paragraph{Size effect on books worse performance
(Figure~\ref{fig:size-effect}).}
We take the fitted zonal kernel to the books ($\tilde{C}$ in Regressions (i) section above) as ground
truth and generate synthetic corpora that differ from the real one only
in being exactly zonal. Concretely, for each pair the kernel value is
converted back to count units by the following
(cf.\ App.~\ref{app:premise} for all definitions): 
\[
R^{\mathrm{fit}}_{ij}
= \frac{2 + \tilde C(t_{ij})}{2 - \tilde C(t_{ij})},
\qquad
\mathrm{rate}_{ij}
= n_i n_j\,\frac{2F}{N}
\times\; R^{\mathrm{fit}}_{ij}.
\]
Off-diagonal counts are drawn as
$W_{ij} \sim \mathrm{Poisson}(m \cdot \mathrm{rate}_{ij})$ at
multipliers $m \in \{0.25, 0.5, 1,2, 4,8, 16,32, 128\}$ (token count scaled from $N$ to $mN$) and are used to generate a new $\Mstar$. Each draw is then scored by the same pipeline as the observed corpus:
per-object cosine is found for the top 3 modes of the centered $\Mstar$ after an aligning orthogonal transform. $1000$ draws per multiplier are used.  Notably, one asymmetry is forced: a simulated draw has no measured diagonal (pair counts are drawn off-diagonal only), so its
diagonal is set to the refit $\tilde C(1)$.  

% However, 

% the zonal kernel is refit on the draw's own counts and the theory
% rebuilt from that refit. One asymmetry is forced: a simulated draw has
% no measured diagonal (pair counts are drawn off-diagonal only), so its
% diagonal is set to the refit's $\hat c_{\mathrm{sim}}(1)$ and the strict top-$3$ dipole per-object cosine is computed after the
% declared one rotation --- $60$ draws per multiplier are used. 

The observed value
sits at the ${\sim}93$rd percentile of the matched simulations at
$m = 1$ (star in Figure~\ref{fig:size-effect}; band at $1\times$:
$[0.27, 0.56]$, observed $0.55$), and the curve rises monotonically toward $1$, passing the level of
the model's per object cosine sim. between $32\times$ and $64\times$
the corpus's counts.  Hence, we can say that the ``effective corpus size'' of the LLM is roughly in this range.  It is plausible that the model could have easily consumed this amount of stargazing material during its training.

% \paragraph{Gram matrices and eigenvalue signs.} The activation Gram is
% positive semidefinite by construction and therefore estimates the kernel
% of $|\Mstar_{\mathrm{eff}}|$, blind to eigenvalue signs
% (Remark~\ref{rem:absM}). Agreement between the signed book kernel and
% the sign-blind activation kernel at $0.92$--$0.97$ doubles as evidence
% that no substantial negative block was rectified away.

\section{Reproducibility}
\label{app:repro}
Code for all experiments is available at \url{https://github.com/liamstoran01/group_invariant_statistics.git}.

\section{AI Use Statement}
\label{app:AI_use}
In this work, we used generative AI tools for formulating mathematical claims, providing critical ingredients for proving mathematical claims, implementing methods in code, assisting the writing of proofs, refining hypotheses,  providing feedback on computational experiments, and interpreting results.  We have not used generative AI tools for generating synthetic data sets, helping develop theoretical models or conceptual frameworks, assisting with translation, cleaning and reformatting a dataset, or supporting qualitative and thematic data analysis.  Additionally, we used generative AI tools for creating code, drafting parts of the paper, analyzing existing literature, proposing keywords for the paper, and helping to find relevant sources.  We have reviewed all AI-assisted work.  Any generated code or mathematics were reviewed and edited for correctness.  An independent literature survey was conducted.  We take responsibility for the final content of this work, including text, claims or artifacts produced with the aid of generative AI.

\end{document}